\documentclass[english,10pt,letterpaper]{article}
\ifdefined\pdfminorversion\pdfminorversion=7\fi

\usepackage[margin=1in]{geometry}
\usepackage[T1]{fontenc}
\usepackage[utf8]{inputenc}
\usepackage{lmodern}
\usepackage{amsmath,amssymb,amsthm,bm,dsfont}
\usepackage{xcolor,colortbl}
\usepackage{graphicx,wrapfig,subcaption,float}
\usepackage[section]{placeins}
\usepackage{tabularx,multirow,makecell,longtable,booktabs,array}
\usepackage{algorithm,algpseudocode}
\usepackage{enumitem,comment,framed,xspace,pifont}
\usepackage{tikz}
\usepackage{arydshln}
\usepackage[sort,compress]{natbib}
\setcitestyle{authoryear,round,citesep={;},aysep={,},yysep={;}}
\usepackage{hyperref}
\hypersetup{colorlinks=true,citecolor=blue,filecolor=blue,linkcolor=blue,urlcolor=blue}
\hypersetup{pdftitle={RobustRLlib: A Unified Library and Benchmark for Robust Reinforcement Learning Algorithms},pdfauthor={Xi Chen, Fengqi Hu, Zeyu Shen, Jiaying Xu, Shangding Gu, Pan Xu, Anqi Liu, Laixi Shi}}

\allowdisplaybreaks


\usepackage{amsmath,amsfonts,bm}

\def\eqref#1{equation~\ref{#1}}

\def\1{\bm{1}}

\DeclareMathAlphabet{\mathsfit}{\encodingdefault}{\sfdefault}{m}{sl}
\SetMathAlphabet{\mathsfit}{bold}{\encodingdefault}{\sfdefault}{bx}{n}

\newlength{\uncertaintyrowheight}
\definecolor{lxscolor}{HTML}{007A70}

\theoremstyle{plain}

\theoremstyle{definition}

\newcommand{\ouralg}{\emph{RobustRLlib}\xspace}

\newcommand{\rgv}{\ouralg}

\title{RobustRLlib: A Unified Library and Benchmark for Robust Reinforcement Learning Algorithms}
\makeatletter
\newcommand{\authormark}[1]{\textsuperscript{\normalfont\@fnsymbol{#1}}}
\newcommand{\authornote}[2]{\protected@xdef\@thanks{\@thanks\protect\footnotetext[#1]{#2}}}
\makeatother

\author{
	Xi Chen\footnotemark[2]\\
	JHU
	\and
	Fengqi Hu\footnotemark[1] \footnotemark[2]\\
	JHU
	\and
	Zeyu Shen\footnotemark[1] \footnotemark[3]\\
	JHU
	\and
	Jiaying Xu\footnotemark[5]\\
	UCSD
	\and
	Shangding Gu\footnotemark[6]\\
	UC Berkeley
	\and
	Pan Xu\footnotemark[7]\\
	Duke
	\and
	Anqi Liu\footnotemark[2]\\
	JHU
	\and
	Laixi Shi\footnotemark[4]\\
	JHU
	\authornote{1}{The second and third authors contributed equally.}%
	\authornote{2}{Department of Computer Science,
		\authormark{3}Department of Applied Mathematics and Statistics,
		\authormark{4}Department of Electrical and Computer Engineering,
		Johns Hopkins University, Baltimore, MD 21218, USA.}%
	\authornote{5}{University of California San Diego, La Jolla, CA 92093, USA.}%
	\authornote{6}{Department of Electrical Engineering and Computer Sciences,
		University of California, Berkeley, Berkeley, CA 94720, USA.}%
	\authornote{7}{Department of Biostatistics and Bioinformatics,
		Duke University, Durham, NC 27705, USA.}%
}
\date{\today}

\begin{document}

\maketitle

\begin{abstract}

Robust reinforcement learning (RL) is critical for reliable real-world deployment, where policies may fail under diverse shifts and disruptions. Existing robust RL algorithms offer a rich set of mechanisms and principles, yet two major barriers limit their broader potential: limited algorithm reusability and an unclear empirical capability boundary. We introduce \textit{RobustRLlib}, an algorithm-centric library and benchmark that addresses both challenges. It integrates 17 representative robust online, offline, and safe RL algorithms under a unified interface for substantially improved reusability, together with an enhanced evaluation platform spanning diverse real-world shifts and advanced humanoid and vision-language-action tasks. Systematic evaluation over 273 perturbation configurations provides an up-to-date capability map of current robust RL algorithms, revealing where they remain effective and where substantial robustness gaps persist. As an open-source, user-friendly platform, \textit{RobustRLlib} offers a reproducible and extensible codebase for applying existing methods to new problems and developing new robust RL algorithms. The code and tutorial are available at \url{https://robustrl-lib.github.io}.
\end{abstract}


\begin{figure}[h]
    \centering
    \includegraphics[width=0.85\linewidth]{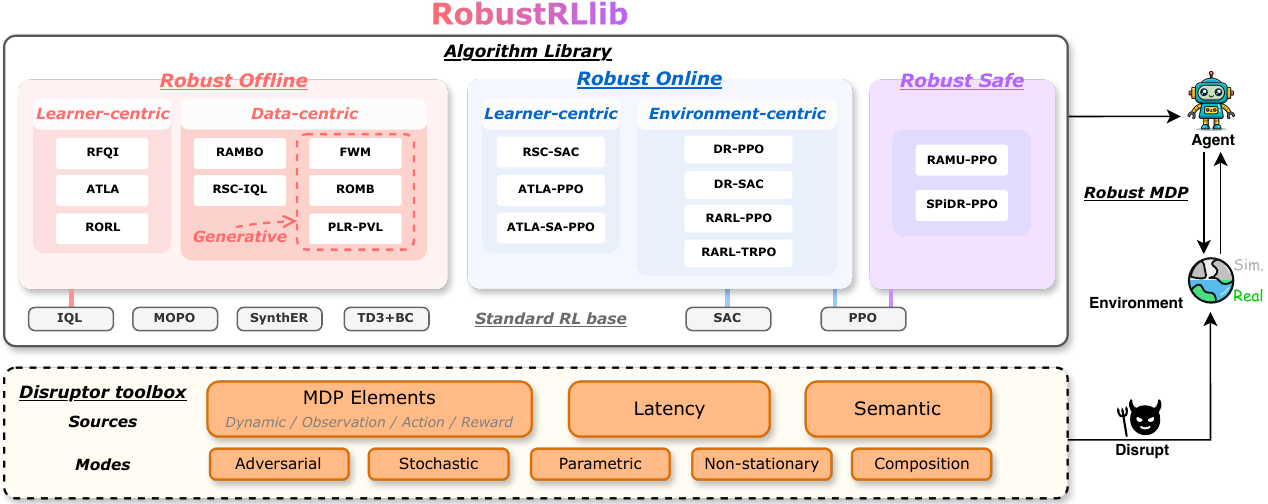}
    \caption{The overview of \ouralg.}
    \label{fig:overview}
\end{figure}

\setcounter{tocdepth}{2}
\tableofcontents

\section{Introduction}

\label{sec:introduction}

Despite remarkable advances in reinforcement learning (RL), its real-world deployment remains limited by robustness and safety concerns, particularly in high-stakes domains such as autonomous driving \citep{ding2023survey}, clinical trials \citep{liu2015robustness}, robotics \citep{li2021reinforcement}, and semiconductor manufacturing \citep{kozak2023stability}. Policies trained in idealized environments can easily fail under distribution shifts or disruptions affecting various stages of the interaction, such as observations \citep{zhang2020robust,zhang2021robust,han2022solution,sun2021exploring,xiong2022defending}, actions \citep{huang2017adversarial}, or environment dynamics \citep{iyengar2005robust,pinto2017robust}, induced by diverse factors such as sim-to-real gaps \citep{pinto2017robust}, model uncertainty \citep{bertsimas2019adaptive}, noise, or adversarial attacks \citep{zhang2020robust,klopp2017robust,mahmood2018benchmarking}.

Robust RL therefore aims to develop policies that remain reliable under diverse types of disruptions and shifts, or their combinations in real-world scenarios \citep{bertsimas2019adaptive,tversky1974judgment,arthur1991designing}. A broad range of robust RL algorithms has shown strong performance in their specific type of disrupted application scenarios. More importantly, existing methods provide a rich set of robustness mechanisms and principles, including adversarial training~\citep{pinto2017robust,zhang2021robust,rigter2022rambo}, dynamics randomization~\citep{peng2018sim}, robust value updates~\citep{yang2022rorl,panaganti2022robust}, and data augmentation~\citep{yu2020mopo,lu2023synthetic,ding2024diffusion}, offering substantial promise across a far broader range of applications and shift scenarios.

However, realizing this broader potential faces two fundamental barriers: \emph{algorithm reusability} and \emph{systematic characterization of the capability boundary}.
First, existing robust RL algorithms are fragmented across separate implementations and training pipelines, lacking a unified library or interface through which practitioners can readily try and compare a broad collection of robust RL methods on a new application. Second, the empirical capability boundary of current robust RL remains unclear: for a given application and distribution shift, we lack a systematic understanding of how well existing methods perform, when they remain effective, and where they fail. This is largely because the methods are typically evaluated in distinct environments and shift scenarios, making cross-method comparison difficult and performance beyond their original settings poorly understood.

To address these two gaps, we introduce \textit{RobustRLlib}, an algorithm-centric robust RL benchmark that brings together (1) a robust RL method library with reusable implementations to substantially lower the barrier to applying existing methods to new problems, and (2) a comprehensive evaluation pipeline across tasks and diverse real-world shift scenarios. Complementing existing robust RL benchmarks that primarily focus on developing evaluation tasks and shift scenarios~\citep{dulac2020empirical,zouitine2024rrls,gu2025robust}, \textit{RobustRLlib} focuses on making robust RL algorithms systematically accessible, comparable, and extensible, while providing a broader evaluation framework (Comparison in Appendix~\ref{app:benchmark_comparison}). Our main contributions are summarized as follows:

\begin{itemize}[leftmargin=*,itemsep=3pt,topsep=3pt]
    \item \textbf{A robust RL algorithm library.}
    We integrate existing 17 representative robust online, offline, and safe RL algorithms, understanding and systematically organizing them by their learning methods, intended  target shifts, and base algorithms for applicators' easy reference. It provides a reusable foundation for applying existing methods to new applications and
developing new robust RL algorithms.

    \item \textbf{An enhanced evaluation platform.}
Our extensible disruptor toolbox covers a broad range of distribution shifts, including MDP-element shifts in observations, actions, dynamics, and reward/cost, as well as control-loop latency and task-semantic shifts. It supports diverse shift modes, including stochastic, adversarial, and parametric perturbations with non-stationary schedules, enabling both isolated and compound-shift evaluation. The same disruptor interface further extends to massively parallel training in Isaac Lab~\citep{mittal2025isaaclab} and advanced vision--language--action (VLA) robotic tasks~\citep{intelligence2025pi05visionlanguageactionmodelopenworld,posttraining_robotics_2026}.

\item \textbf{An up-to-date capability map of robust RL algorithms.} Evaluation over 273 perturbation configurations reveals that
   robust methods outperform standard references on
average, but both groups deteriorate markedly as shifts intensify. A method's
claimed target shift does not reliably predict its empirical strengths, gains vary
across tasks and shift sources, with online methods showing larger incremental effects than offline ones. By connecting
diverse algorithm features to their measured strengths, this map provides
guidance for designing robust RL methods for deployment shifts.

\end{itemize}

\section{Related Work}
\label{sec:related_work}

\paragraph{Robust RL problems under diverse deployment shifts.}

Robust RL addresses deployment shifts affecting different components and stages of the
agent--environment interaction, and prior work can be categorized accordingly.
\citet{zhang2020robust,zhang2021robust,oikarinen2021robust}
address \emph{observation shift}, which perturbs policy inputs through
sensing errors or external attacks.
\citet{tessler2019action,tan2020robustifying}
address \emph{action shift}, which perturbs or replaces executed actions.
\emph{Dynamics shift}, involving uncertainty in or changes to the transition kernel,
is addressed either by optimizing worst-case performance over ambiguity sets
\citep{nilim2005robust,iyengar2005robust,xu2010distributionally,
wiesemann2013robust,goyal2023robust,shi2026curious,
wang2024sample,shi2024distributionally}
or by introducing random or adversarial perturbations to physical parameters
and external forces
\citep{rajeswaran2016epopt,peng2018sim,pinto2017robust,mankowitz2019robust}.
\citet{queeney2023risk,ma2019policy}
address \emph{reward and cost shifts} through constraint-aware learning, and reward-poisoning analysis.
Beyond these MDP components,
\citet{bouteiller2021reinforcement,peng2018sim,tan2018sim}
address \emph{latency shift}, involving delays or irregularities in observation
and action timing.
\citet{tobin2017domain,ding2023seeing}
address \emph{semantic shift}, involving changes in appearance or task-relevant
relationships among visual cues, objects, and goals.
In practice, real-world scenarios may involve multiple shifts, such as changes in dynamics
parameters, process noise, and observation noise
\citep{mandlekar2017adversarially}, or non-stationary shifts that evolve
during deployment \citep{lecarpentier2019non}.
Since robustness to one shift does not necessarily extend to others or their
combinations, this motivates developing algorithms for co-occurring shifts
and evaluating them under a unified framework.

\paragraph{Robust RL algorithm taxonomy.}
Robust RL algorithms promote robustness through two main approaches:
modifying the learning objective or algorithm (\emph{learner-centric})
and modifying the training data distribution (\emph{data-centric}).
For \emph{robust online RL}, learner-centric methods seek robust policies
through optimization against an adversary~\citep{zhang2020robust,zhang2021robust},
worst-case robust value objectives~\citep{liang2022efficient,li2024towards},
smoothing the policy over perturbed observations~\citep{sun2024breaking},
or causal-inspired data augmentation to handle spurious correlated shifted scenarios~\citep{ding2023seeing}. 
Data-centric methods can collect trajectories by interacting with shifted environments through physical parameter randomization~\citep{peng2018sim},
model ensembles~\citep{rajeswaran2016epopt,mankowitz2019robust,tanabe2022max},
or by applying adversarial disturbances~\citep{pinto2017robust}.
For \emph{robust offline RL}, learner-centric methods address
model uncertainty and limited data coverage through robust Bellman
updates over an uncertainty set of transitions~\citep{panaganti2022robust,blanchet2023double,panaganti2023bridging}, and mitigate value overestimation by smoothness regularization over perturbed states~\citep{yang2022rorl}.
Data-centric methods instead augment or reweight fixed datasets by generating samples from adversarial environments~\citep{rigter2022rambo,bhardwaj2023adversarial}, mixing diffusion-generated synthetic data with real data~\citep{lu2023synthetic}, or downweighting corrupted samples~\citep{yang2024uncertainty}.
In particular, generative-model methods, as a data-centric subfamily, learn dynamics models from offline data and generates plausible
future transitions or multi-step trajectories for robust policy learning
\citep{ding2024diffusion,berdica2024robust,foffano2026adversarial}.
Furthermore, \emph{robust safe RL} extends robust policy learning
to jointly maintain task performance and satisfy cost constraints
under uncertainty
\citep{liu2022robustness,wang2022robust,queeney2023risk}.
In this work, \rgv integrates 17 representative algorithms spanning
robust online, offline, and safe RL, systematically organizes them by
algorithm features, and evaluates them across diverse shift
sources and modes under a unified protocol.
It provides a reusable foundation for applying existing methods to
new applications and developing new robust RL algorithms.

\paragraph{Related RL benchmarks.}
Existing RL benchmarks can be broadly categorized into
\emph{data-oriented benchmarks}, which primarily provide fixed datasets for algorithm development
\citep{fu2020d4rl,gulcehre2020rl,qin2022neorl};
\emph{task-oriented benchmarks}, which offer standardized tasks and environments
for developing and evaluating RL algorithms, ranging from continuous control to advanced robotics
\citep{duan2016benchmarking,yu2020meta,sferrazza2024humanoidbench,
ji2023safety,pumacay2024colosseum}; and
\emph{algorithm-oriented libraries and benchmarks}, which provide reproducible implementations,
training infrastructure, and pipelines for existing RL algorithms,
serving as a foundation for downstream applications, fair comparisons, and further algorithm development
\citep{liang2018rllib,hu2023marllib,raffin2021stable,
huang2022cleanrl,weng2022tianshou}.
In particular, for robust RL, closely related task-oriented benchmarks include
RRLS, which focuses on continuous control \citep{zouitine2024rrls},
and Robust-Gymnasium, which covers a broader range of control and robotics tasks
\citep{gu2025robust}.
Complementing these efforts, this work introduces \ouralg,
the first unified algorithm-oriented library and benchmark for robust RL.
It provides a reusable foundation for applying existing methods to
new applications and developing new robust RL algorithms, while enabling
easy comparisons with existing methods.
In addition, \ouralg extends the considered deployment shifts to include
latency and semantic shifts, and expands evaluation to advanced high-dimensional robotic tasks.
There exist other benchmarks that address related settings, such as off-dynamics transfer,
real-world operating stressors, and adversarial robustness
\citep{lyu2024odrl,dulac2020empirical,mohan2026road}. \looseness = -1

\paragraph{Sim-to-real gap in robotics.}
The sim-to-real gap is a longstanding challenge in robotics, where mismatches between simulation and real-world hardware can substantially degrade the performance of policies trained in simulation. Prior work addresses this gap through various approaches, including dynamics randomization~\citep{peng2018sim,andrychowicz2020learning}, actuator and latency modeling~\citep{tan2018sim}, and online adaptation to changing dynamics~\citep{kumar2021rma}. These considerations have been explored in whole-body humanoid control~\citep{gu2024humanoid,he2024omnih2o}, safety-critical robotics~\citep{as2026spidr}, real-time VLA policies that overlap execution and inference~\citep{black2026real}, and demonstration-driven policies for contact-rich manipulation~\citep{sharma2026demonstrationobjectsgeneralizingmanipulation}. However, many existing approaches are task-specific and address particular sources of sim-to-real mismatch. In this work, we provide a unified robust RL algorithm library with 17 representative algorithms covering diverse tasks and sim-to-real gaps. These algorithms and their underlying design principles have the potential to support broader robotics applications while reducing manual iteration before hardware deployment.

\section{\ouralg: An Algorithm-Centric Robust RL Benchmark}
\label{sec:robust_gymnasium_v2}

\ouralg is organized around \emph{algorithm attribution}: it records how a
robust RL method changes learning, which shift the method claims to
address, and which base algorithm realizes the method.  The disruptor toolbox
then instantiates the claimed target shift independently of the base learner, while
the evaluation platform measures both common robustness and method-specific
specialization.  This separation makes each result traceable to an algorithmic
feature and an evaluation condition rather than treating robust RL as a
single undifferentiated class.

\paragraph{Markov decision processes (MDP).}
We consider a finite-horizon MDP
$\mathcal{M}_0=(\mathcal{S},\mathcal{A},T,P_0,r_0)$, where
$\mathcal{S}$ and $\mathcal{A}$ are the state and action spaces,
$T$ is the horizon, and $P_0$ and $r_0$ denote the nominal transition
kernel and reward function. At time $t$, the policy selects
$a_t\sim\pi_t(\cdot\mid s_t)$, receives reward $r_{0,t}(s_t,a_t)$,
and the environment transitions according to
$s_{t+1}\sim P_{0,t}(\cdot\mid s_t,a_t)$. Classical robust MDPs replace the nominal
transition model $P_0$ with a shift set $\mathcal{P}$ and seek a policy
with high worst-case return~\citep{iyengar2005robust,nilim2005robust}:
\begin{equation}
\label{eq:disruptor_robust_objective}
\max_{\pi}\inf_{P\in\mathcal{P}}J(\pi;P,r_0),
\qquad
J(\pi;P,r_0)=
\mathbb{E}_{\pi,P}\!\left[\sum_{t=1}^{T}r_{0,t}(s_t,a_t)\right].
\end{equation}

For safety-constrained tasks, we augment the MDP with a cost function
$c=\{c_t\}_{t=1}^{T}$ and a budget $d$, requiring
$\mathbb{E}_{\pi,P}[\sum_{t=1}^{T}c_t(s_t,a_t)]\leq d$ in addition to
maximizing task return.  This cost interface is retained when shift is
introduced, allowing robustness and safety to be evaluated jointly.

\subsection{Robust RL algorithm library construction}
\label{sec:library_objective}

\newcommand{\labeldash}{\xleaders\hbox to 4pt{\hss\rule[2.1pt]{2pt}{\arrayrulewidth}\hss}\hfill}
\newcommand{\dashedlabel}[1]{\smash{\raisebox{-5.6pt}{\makebox[\dimexpr0.20\textwidth+2\tabcolsep\relax]{\labeldash\ #1\ \labeldash}}}}

\begin{table*}[t]
\centering
\footnotesize
\setlength{\aboverulesep}{0pt}
\setlength{\belowrulesep}{0pt}
\setlength{\tabcolsep}{4pt}
\renewcommand{\arraystretch}{1.25}
\begin{tabular}{>{\centering\arraybackslash}m{0.125\textwidth}|>{\centering\arraybackslash}m{0.20\textwidth}|c|c|c|c|c|c}
\hline\toprule
\multirow{2}{*}{Problem} & \multirow{2}{*}{Algorithm feature} & \multirow{2}{*}{Method} & Target & Base & Shifted-env & Adversarial & Learned \tabularnewline
 & & & shift & algorithm & rollout & network & model \tabularnewline
\hline\toprule
\multirow{8}{=}{\centering\textbf{Robust\\Offline}}
 & \multirow{3}{*}{Learner-centric} & RFQI & Dynamic & FQI & & & \tabularnewline
 & & RORL & Observation & SAC & & & \tabularnewline
 & & ATLA & Observation & IQL & & \ding{51} & \tabularnewline
\cline{2-8}
 & & RSC-IQL & Semantics & IQL & & & \ding{51} \tabularnewline
 & \multicolumn{1}{@{}c@{}|}{\dashedlabel{Data-centric}} & RAMBO & Dynamic & SAC & & \ding{51} & \ding{51} \tabularnewline
 & \multirow{3}{*}{\textit{Generative model}} & ROMB & Dynamic & IQL & & \ding{51} & \ding{51} \tabularnewline
 & & FWM & Dynamic & IQL & & & \ding{51} \tabularnewline
 & & PLR-PVL & Dynamic & IQL & & & \ding{51} \tabularnewline
\hline\toprule
\multirow{5}{=}{\centering\textbf{Robust\\Online}}
 & \multirow{3}{*}{Learner-centric} & ATLA & Observation & PPO & & \ding{51} & \tabularnewline
 & & ATLA-SA & Observation & PPO & & \ding{51} & \tabularnewline
 & & RSC & Semantic & SAC & & & \ding{51} \tabularnewline
\cline{2-8}
 & \multirow{2}{*}{Environment-centric} & RARL & Dynamic & PPO / TRPO & \ding{51} & \ding{51} & \tabularnewline
 & & DR & Dynamic & SAC / PPO & \ding{51} & & \tabularnewline
\hline\toprule
\multirow{2}{=}{\centering\textbf{Robust\\Safe}}
 & \multirow{2}{*}{Robust safe} & RAMU & Dynamic & PPO & & & \tabularnewline
 & & SPiDR & Dynamic & PPO & \ding{51} & & \ding{51} \tabularnewline
\hline\toprule
\end{tabular}
\caption{\textbf{Structural comparison of the robust methods in the library.}
Online methods are grouped as learner-centric or environment-centric, offline
methods as learner-centric or data-centric; \emph{generative model} is a data-centric
subfamily, and robust safe methods form a separate family. \emph{Shifted-env rollout} marks
rollouts collected from a deliberately perturbed physical environment.
\emph{Adversarial network} marks a learned adversary
optimized against the policy. \emph{Learned model} marks a
trained next-state predictor used by the  method.
\emph{Target shift} records the intended shift source. A checkmark
denotes a present feature; a blank denotes its absence.}
\label{tab:method_checkbox}
\end{table*}

The library has three tracks: \emph{offline} RL learns from fixed data, \emph{online} RL learns through continued
environment interaction,  and
\emph{safe} RL adds cost constraints, including robust-safe methods implemented
here in the online setting. Within the offline and online tracks, robustness
primarily enters through two routes: \emph{learner-centric} methods change how
experience is interpreted or optimized, while \emph{data-centric} and \emph{environment-centric} methods change
the distribution available for learning. 
 Table~\ref{tab:method_checkbox} summarizes each method's
feature, target shift, and base algorithm.

\subsubsection{Robust offline RL algorithms}
\label{sec:offline_library}

Offline RL learns from a fixed dataset without environment interaction,
making limited data coverage and optimistic value estimates for unsupported
actions central challenges. Robust offline RL must also maintain performance
under distribution shift. We therefore group its methods by how they address
these challenges: \emph{learner-centric} methods modify the objective or update,
whereas \emph{data-centric} methods reshape the training distribution,
including through data augmentation or generative models.
As standard RL baselines, IQL~\citep{kostrikov2021offline} and
TD3+BC~\citep{fujimoto2021minimalist} learn directly from logged transitions;
MOPO~\citep{yu2020mopo} adds model rollouts, and
SynthER~\citep{lu2023synthetic} adds diffusion-generated datasets.

\paragraph{Learner-centric methods.}
These methods make policy learning less sensitive to uncertain transitions or
perturbed observations by changing the value target or policy update.
\begin{itemize}[leftmargin=*, nosep]
    \item RFQI: worst-case Bellman update.
    RFQI~\citep{panaganti2022robust} replaces the nominal fitted-$Q$ target
    with a worst-case backup over transition ambiguity.

    \item RORL: local policy and value smoothing.
    RORL~\citep{yang2022rorl} combines pessimistic bootstrapping with
    smoothing around states to stabilize predictions under
    observation perturbations.

    \item ATLA: observation adversary.
    The offline ATLA adaptation~\citep{zhang2021robust} trains an observation
    adversary alongside standard base algorithm, exposing policy updates to adverse inputs.
\end{itemize}

\paragraph{Data-centric methods.}
These methods expand the experience available to the base learner through
counterfactual transitions or model-generated rollouts.
\begin{itemize}[leftmargin=*, nosep]
    \item RSC: causal counterfactual rewrite.
    The offline RSC~\citep{ding2023seeing}  adaptation  rewrites samples using an inferred causal mask to weaken spurious semantic correlations.
    \item RAMBO: adversarial model rollouts.
    RAMBO~\citep{rigter2022rambo} branches short synthetic rollouts from
    dataset states and adjusts its dynamics model adversarially within a
    data-fit constraint.
\end{itemize}

\paragraph{\emph{Generative model methods.}}
These data-centric methods learn conditional models of transition dynamics
and use generated next states or multi-step futures in policy learning,
which potentially improve robustness under dynamics shifts. SynthER is excluded from this family because it models the distribution of logged transitions  to enlarge the dataset rather than modeling dynamics for policy learning.
\begin{itemize}[leftmargin=*, nosep]
    \item ROMB: maximin model ensemble.
    ROMB~\citep{chen2025policy} optimizes against adverse transitions
    represented by an ensemble of learned models.

    \item FWM: flow-generated multi-step targets.
    FWM uses flow-matching to replace diffusion world model~\citep{ding2024diffusion} and samples future trajectories to construct multi-step critic targets.

    \item PLR-PVL: worst-case world model selection.
    PLR-PVL~\citep{berdica2024robust} searches a flow-model ensemble for
    adverse plausible futures during policy learning.
\end{itemize}

\subsubsection{Robust online RL algorithms}
\label{sec:online_library}
Online robust RL methods preserve continued interaction with the environment
but differ in where robustness enters training.  We distinguish two primary
routes.  \emph{Learner-centric} methods modify the policy input and learning objective while retaining the nominal physical
transition law.  \emph{Environment-centric} methods instead interact with deliberately perturbed environments to collect
training rollouts. Whether a method updates on-policy or
off-policy remains an important implementation choice, so we record it through
the base algorithm. 
PPO~\citep{schulman2017proximal}, TRPO~\citep{schulman2015trust} and SAC~\citep{haarnoja2018soft} serve as the
standard base algorithms in our experiments.

\paragraph{Learner-centric methods.}
These methods train the policy to make stable decisions using adversarial
observations, local regularization, or causal counterfactual experience.
\begin{itemize}[leftmargin=*, nosep]
    \item ATLA: observation adversary.
    ATLA~\citep{zhang2021robust} alternates updates between the policy and a
    learned adversary that perturbs its observations.

    \item ATLA-SA: state-adversarial regularization.
    ATLA-SA~\citep{zhang2021robust} adds recurrent history and SA-MDP
    regularization to stabilize decisions under perturbed observations.

    \item RSC: causal counterfactual replay.
    RSC~\citep{ding2023seeing} infers causal dependencies and rewrites
    experience counterfactually to reduce reliance on spurious semantic features.
\end{itemize}

\paragraph{Environment-centric methods.}
These methods build robustness through training trajectories generated under
varied physical environments or external disturbances.

\begin{itemize}[leftmargin=*, nosep]
    \item RARL: external force adversary.
    RARL~\citep{pinto2017robust} perturbs the environment with bounded forces
    selected by an adaptive adversary. The protagonist and adversary train in
    a two-player game, concentrating experience on the policy's current
    vulnerabilities.

    \item DR: domain randomization.
    DR~\citep{peng2018sim} samples physical parameters across training
    episodes, exposing the policy to a prescribed envelope of transition dynamics.
\end{itemize}

\subsubsection{Robust safe RL algorithms}

Robust safe RL methods seek to maximize worst-case expected return while ensuring cost-constraint satisfaction across a prescribed dynamics uncertainty set.

\begin{itemize}[leftmargin=*, nosep]
    \item RAMU: risk-sensitive reward and cost targets.
    RAMU~\citep{queeney2023risk} applies risk measures over transition-model
    uncertainty to both reward and cost Bellman targets.

    \item SPiDR: robust safety via penalized costs.
    SPiDR~\citep{as2026spidr} measures disagreement among next-state
    predictions from domain-randomized models and penalizes
    uncertain safety outcomes.
\end{itemize}

\subsection{Expandable disruptor toolbox}
\label{sec:disruptor}

Robust RL preserves the underlying task objective of standard RL by maximizing cumulative return but extends the single nominal MDP into a family of robust MDP with different types of shifts.  These shifts intervene in different components of the agent–environment interaction, such as observations, actions, transition dynamics, rewards or costs, latency, and task semantics. The resulting robust RL problem therefore seeks a policy that performs reliably across these shifted interaction conditions.

\subsubsection{Shift sources}
\label{sec:shift_types}

\begin{figure}[t]
    \centering
    \begin{subfigure}[b]{0.310\linewidth}
        \centering
        \includegraphics[width=\linewidth]{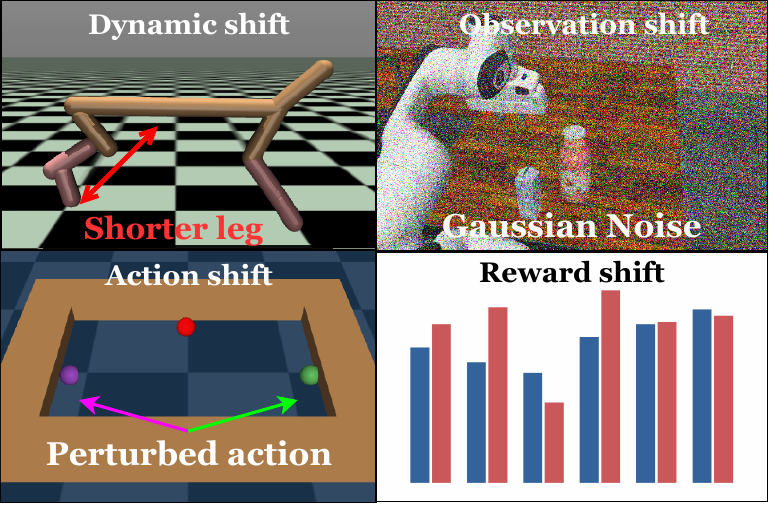}
        \caption{MDP-element shifts.}
        \label{fig:mdp}
    \end{subfigure}\hfill
    \begin{subfigure}[b]{0.341\linewidth}
        \centering
        \includegraphics[width=\linewidth]{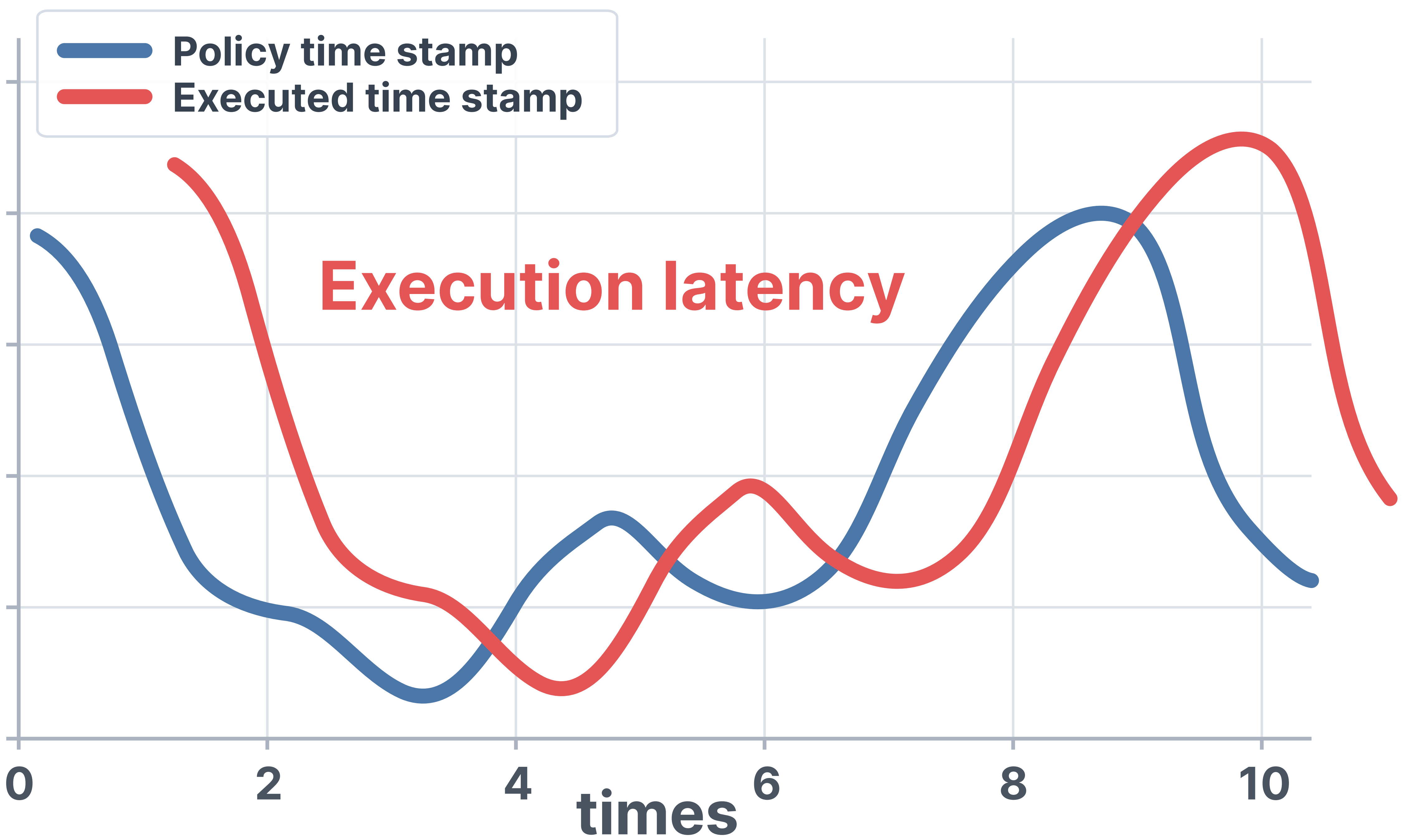}
        \caption{Latency shift.}
        \label{fig:latency_shift}
    \end{subfigure}\hfill
    \begin{subfigure}[b]{0.318\linewidth}
        \centering
        \includegraphics[width=\linewidth]{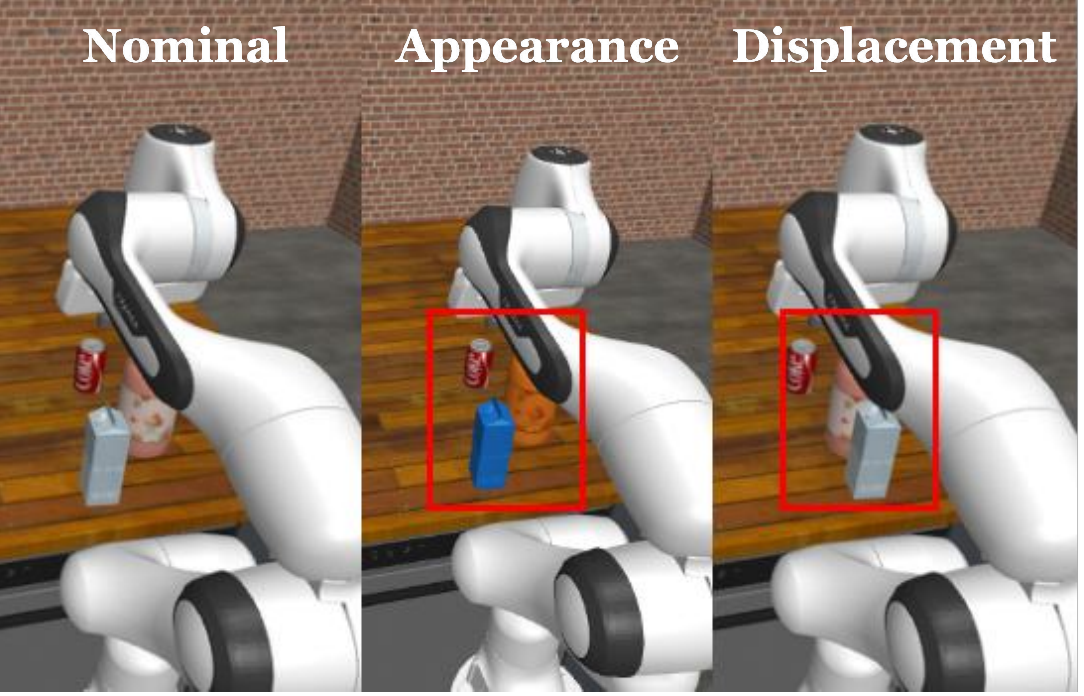}
        \caption{Semantic shift.}
        \label{fig:semantic_shift}
    \end{subfigure}
    \caption{\textbf{Shift sources covered in \rgv.} (a) Shifts of the MDP elements:
    dynamic, observation, action, and reward. (b) Latency between the policy's and the executed time stamp. (c) Semantic shift in appearance
    and goal displacement.}
    \label{fig:shift_sources}
\end{figure}

We classify a shift by where it enters the agent--environment interaction.
The first four sources act directly on components of the MDP interface
(Figure~\ref{fig:mdp}); latency (Figure~\ref{fig:latency_shift}) and semantics
(Figure~\ref{fig:semantic_shift}) then capture deployment factors that are
not represented cleanly by a single nominal MDP component.
Given $x_t=(s_t,\tilde a_t)$ and
$y_t=(s_t,\tilde a_t,s_{t+1})$, we expose the six shift sources as below:

\begin{itemize}[leftmargin=1.4em,itemsep=2pt,topsep=3pt]
    \item \textbf{Dynamic shift.}
    Changes the environment response under the same state and
    executed action,
    $s_{t+1}\sim P_{\theta^p}(\cdot\mid x_t)$,
    where $\theta^p$ specifies physical-parameter changes,
    external disturbances, and latent process noise.
    Its operators cover physical changes including
    \emph{internal dynamics} such as morphology, friction, gravity
    and \emph{external disturbance} such as wind, external pushes,
    and latent noise.
    These variations represent physical mismatch and motivate
    dynamics randomization for sim-to-real
    control~\citep{peng2018sim}.

    \item \textbf{Observation shift.}
    Changes the information available to the policy,
    $\tilde o_t\sim O_{\theta^o}(\cdot\mid s_t)$ and
    $a_t\sim\pi(\cdot\mid\tilde o_t)$,
    where $\theta^o$ specifies the perturbation applied to the
    policy observation.
    It represents sensor noise, bias, missing or imperfect feedback,
    and bounded observation attacks~\citep{dulac2020empirical}.

    \item \textbf{Action shift.}
    Changes the command that reaches the plant,
    $\tilde a_t\sim A_{\theta^a}(\cdot\mid s_t,a_t)$,
    where $\theta^a$ specifies how the intended action is altered
    before execution.
    It captures execution noise, calibration error, saturation,
    opposition, and other actuator-side
    distortions~\citep{tessler2019action}.

    \item \textbf{Reward/cost shift.}
    Changes the feedback used for task learning or safety evaluation,
    $\tilde r_t\sim R_{\theta^r}(\cdot\mid y_t)$ and
    $\tilde c_t\sim C_{\theta^c}(\cdot\mid y_t)$,
    where $\theta^{rc}=(\theta^r,\theta^c)$ specifies the reward
    and cost perturbations.
    It represents noisy or biased supervision while keeping
    the underlying transition explicit.

    \item \textbf{Latency shift.}
    Captures deployment-time mismatch in the timing of sensing,
    action execution, control, and VLA inference delay
    \citep{dulac2020empirical,bouteiller2021reinforcement,tan2018sim}.
    We treat latency as a composable deployment-shift source within
    this unified robust RL evaluation framework.
    Specifically, we expose observation delay, action delay,
    and control-period variation through
    $\tilde o_t=\mathcal{I}(\mathcal{H}_t,t-\delta_o),
    \tilde a_t=a_{t-\delta_a},
    s_{t+1}\sim P_0^{(\kappa_t)}(\cdot\mid s_t,\tilde a_t)$,
    capturing delayed sensing, delayed execution, and variable
    control rates, respectively.
    Implementation details are provided in
    Appendix~\ref{app:latency}.

    \item \textbf{Semantic shift.}
    Changes task-relevant scenes or structures rather than merely
    adding task-agnostic noise.
    Such shifts can be represented as
    $o_t\sim O_{\theta^z}(\cdot\mid s_t),
    s_{t+1}\sim P_{\theta^z}(\cdot\mid x_t),
    (r_t,c_t)\sim(R_{\theta^z},C_{\theta^z})(\cdot\mid y_t)$,
    where $\theta^z$ specifies appearance or task-semantic
    parameter changes.
    Camera, lighting, color, blur, and distractor shifts act
    primarily through $O_{\theta^z}$, whereas relocating a goal
    or object may also change the transition dynamics and task
    feedback.
    Such semantic shifts are increasingly important for visually
    grounded policies, where deployment may involve diverse
    task-specific scenes, including different objects, workspaces,
    viewpoints, layouts, or instructions
    \citep{brohan2023rt,pumacay2024colosseum,fei2025libero},
    potentially inducing semantic spurious correlations as an
    important deployment shift
    \citep{tobin2017domain,ding2023seeing}.
\end{itemize}

\subsubsection{Shift modes}
A mode says \emph{how} an intervention is built, and is one field of the
specification tuple (source, factor, mode, magnitude, resampling rule, optional
schedule, seed): the \emph{source} says \emph{where} the intervention enters the agent--environment
interaction (Section~\ref{sec:shift_types}), the \emph{factor} says \emph{what} it perturbs (e.g., gravity or
friction), the \emph{magnitude} says \emph{how far} it departs from nominal, the resampling rule and
optional schedule say \emph{how it evolves} within an episode, and the seed makes its random draws
reproducible. Each instantiated tuple defines one \emph{perturbation configuration}. We retain the stochastic interface of
Robust-Gymnasium~\citep{gu2025robust}, and extend it with several new modes as below:

\begin{itemize}[leftmargin=1.4em,itemsep=2pt,topsep=3pt]
  \item \textbf{Stochastic.} Samples signal errors or parameters from
  specified distributions.
  \item \textbf{Adversarial.} Computes bounded
  perturbations from the current states to
  counter the policy.
  \item \textbf{Parametric.} Applies a specified assignment, scale,
  offset, or translation at a severity.
  \item \textbf{Non-stationary.} Varies a mode's severity over time or
  triggers intermittent disturbances.
  \item \textbf{Composition.} Stacks compatible
  shifts into reproducible multi-shift scenarios.
\end{itemize}

Together, these controls support both individual shift and
compound shift scenarios. The appendix~\ref{app:benchmark_construction} provides the operators, magnitudes, resampling rules, and composition procedure.

\subsubsection{Expandable to advanced robotics tasks}
Because a shift is fully described by its specification tuple, a new backend
needs only one adapter to the simulator's own primitives. We validate this on two deployment-oriented platforms.  For Isaac Lab~\citep{mittal2025isaaclab}, the
adapter realizes transition, action, and latency shifts on the vectorized GPU
environment and exposes a joint-limit cost channel for every method's evaluation, so unconstrained and constrained learners are
compared on one safety quantity.  For a VLA
policy~\citep{intelligence2025pi05visionlanguageactionmodelopenworld} fine-tuned on the bimanual demonstrations~\citep{posttraining_robotics_2026}, the adapter injects appearance,
proprioception, action, latency, and physical shifts into a closed-loop
simulator. 

\section{Experimental Evaluation}
\label{sec:exp_evaluation}

\begin{figure}[h]
    \centering
    \includegraphics[width=\linewidth]{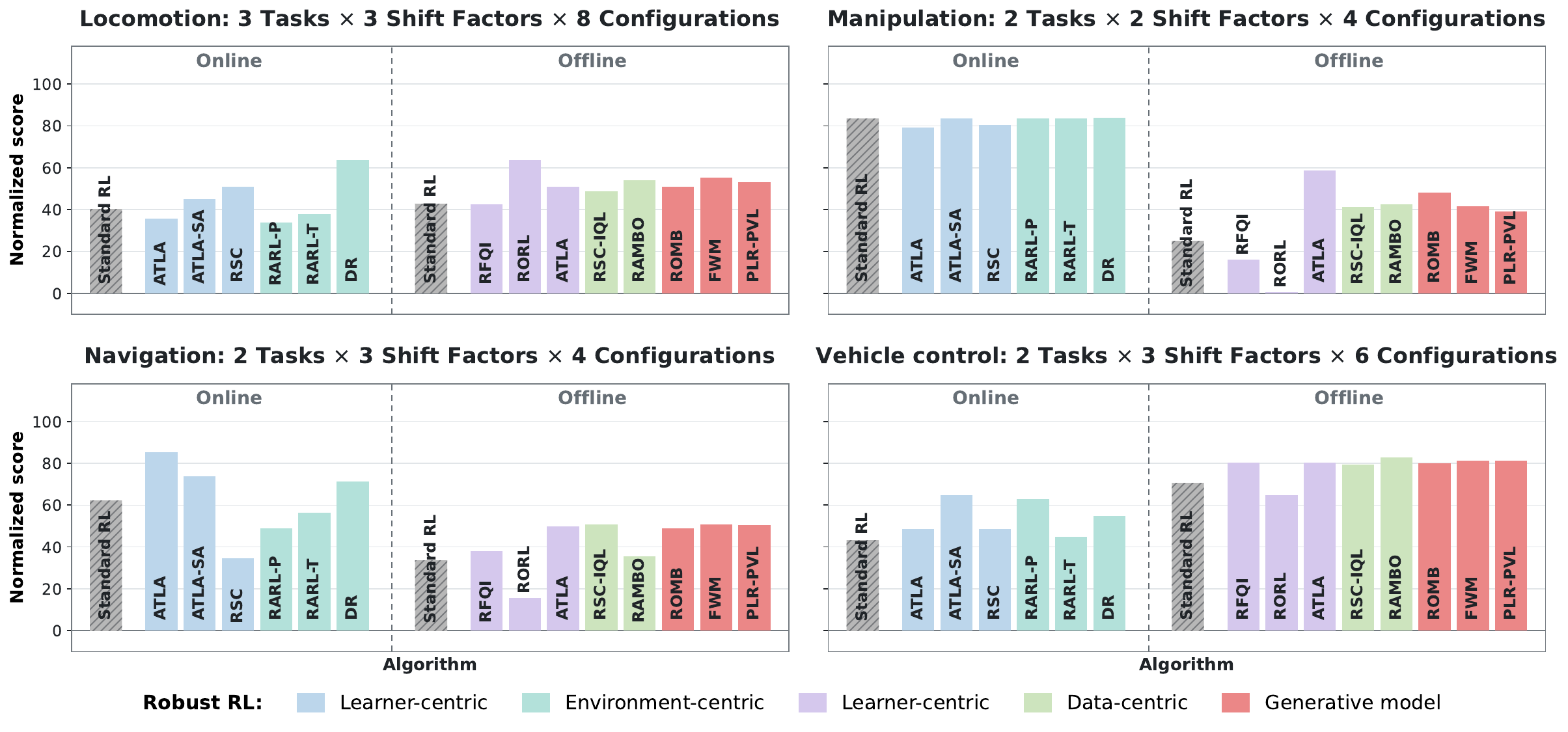}
    \caption{\textbf{Performance evaluation of 20 robust RL algorithms in the \ouralg library under unified perturbation configurations.} Every colored bar is one of the fourteen robust methods, keyed by algorithm feature and online/offline setting. Scores average the perturbation configurations within each shift factor, factors within each task, and then tasks with equal weight. The aggregation covers 20 algorithms across 220 distinct perturbation configurations.}
    \label{fig:part1_regime}
    \label{fig:main_res}
\end{figure}

The evaluation proceeds in three blocks over our library.
Section~\ref{sec:exp_part1} measures how much robust methods gain across the
whole library and how that gain behaves as the shifts intensify.
Section~\ref{sec:channels} isolates each shift source and composes
them, to ask which shift source a method actually defends.
Section~\ref{sec:advanced_tasks} carries the same interface to
deployment-oriented humanoid and vision--language--action (VLA) systems.

\paragraph{Experimental setup.}
The library implements 23 algorithms: 6 standard RL references, 15 robust RL  and 2 robust safe RL algorithms. The library-wide grid runs the 20 algorithms across
Locomotion, Manipulation, Navigation and Vehicle Control; one robust RL and two robust safe
algorithms enter on the advanced robotic tasks. Every algorithm starts from the same nominal task
specification, uses its native training procedure and prescribed budget, and is
then frozen and evaluated on the same shift grid. The results use five training seeds and 20 paired
evaluation episodes per condition.
Exact environments, perturbation configurations, and protocols are in Appendix~\ref{app:evaluation_grid}.

\paragraph{Performance metric.}
For a comparable return across different tasks, we report the unclipped normalized score $\mathrm{S}(\bar R)=100(\bar R-R_{\min})/(R_{\max}-R_{\min})$ as our evaluation metric, where $R_{\min}$ and $R_{\max}$ are fixed task-level reference returns shared by
every method. Appendix~\ref{app:metric} details the metrics.

\subsection{Evaluation of the algorithm library}
\label{sec:exp_part1}

\begin{figure}[h]
    \centering
    \includegraphics[width=0.85\linewidth]{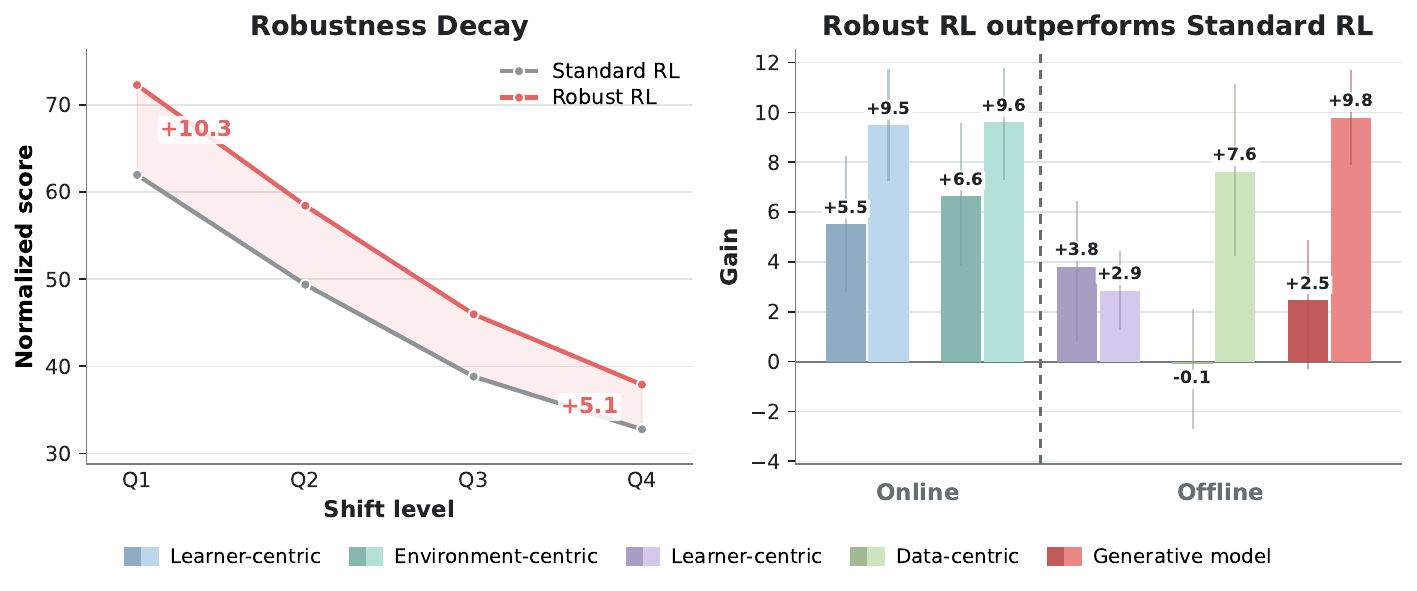}
    \caption{
\textbf{Robust RL gains across shift levels and algorithm features.}
\emph{Left}: Aggregated performance of standard and robust RL algorithms across diverse environment shift levels Q1--Q4 (Refer to Appendix~\ref{app:eval} for more details).
\emph{Right}: performance gains over respective base algorithms (dark) and the average of standard RL (light) across diverse algorithm designs. See Appendix~\ref{app:gain_columns} for details.}

    \label{fig:decay}
\end{figure}
Robust RL methods differ in their training assumptions and access to data and
environments, making results from their original settings difficult to compare
directly. We therefore evaluate their final policies on representative tasks to identify
their capability boundaries. Our first comparison covers 20 online
and offline algorithms, including 6 standard and 14 robust RL, over 220 perturbation configurations across four task
families. Figure~\ref{fig:part1_regime} shows each robust algorithm separately,
while Figure~\ref{fig:decay} aggregates the same evaluation by shift levels and
algorithm features to show both the robust--standard gap and feature-grouped algorithms' gain
over its base algorithms.

\paragraph{Robust RL helps at every shift level, but both degrade as shifts intensify.}
Across all four shift levels, robust methods achieve a higher pooled
normalized score than standard RL. Their advantage remains positive but
narrows from $+10.3$ points in Q1 to $+5.1$ in Q4, while both groups' scores
decline as shifts intensify.

\paragraph{Methods that explicitly model dynamics shifts exhibit greater robustness to such shifts.}
Two families explicitly target dynamics shifts: environment-centric online methods such as DR-SAC perturb the environment simulator during training, while generative offline methods such as FWM learn a generative dynamics model and train on its trajectories. These families achieve the strongest performance under the controlled dynamic shifts within their respective online and offline settings. Overall, robustness gains reflect both the robust methods and the strength of the base learner.

\subsection{Evaluation under diverse deployment shifts}
\label{sec:channels}

We evaluate method-specific robustness through individual shift source sweeps across 44 perturbation configurations and
reference-derived compound shifts. The former  cover multiple
perturbation modes. For the latter, Hopper adapts the coupled physical, actuation, force, and delay factors
of OmniH2O~\citep{he2024omnih2o}, while Pusher adapts the dynamics,
observation, and control-rate factors of \citet{peng2018sim}, with goal
displacement labeled separately. Appendix~\ref{app:eval} details the
individual and the compound shift
scenarios.

\begin{figure}[h]
    \centering
    \includegraphics[width=0.95\linewidth]{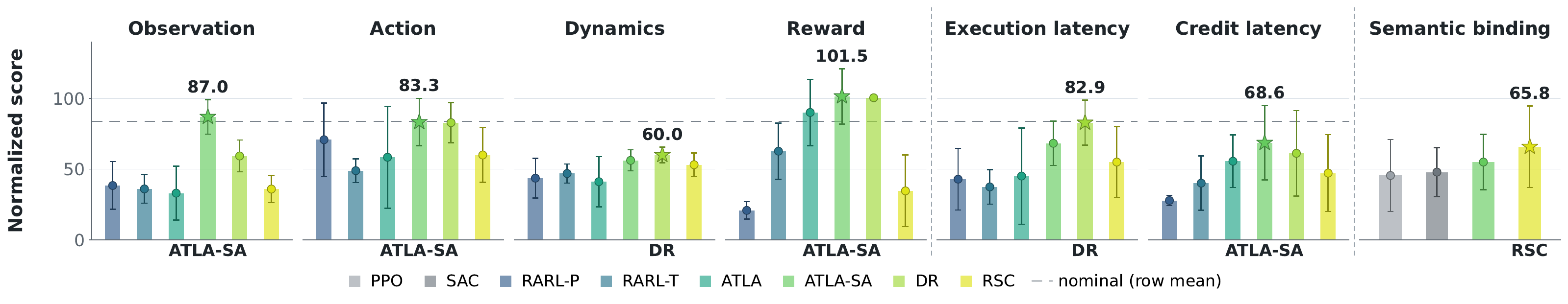}
    \includegraphics[width=0.95\linewidth]{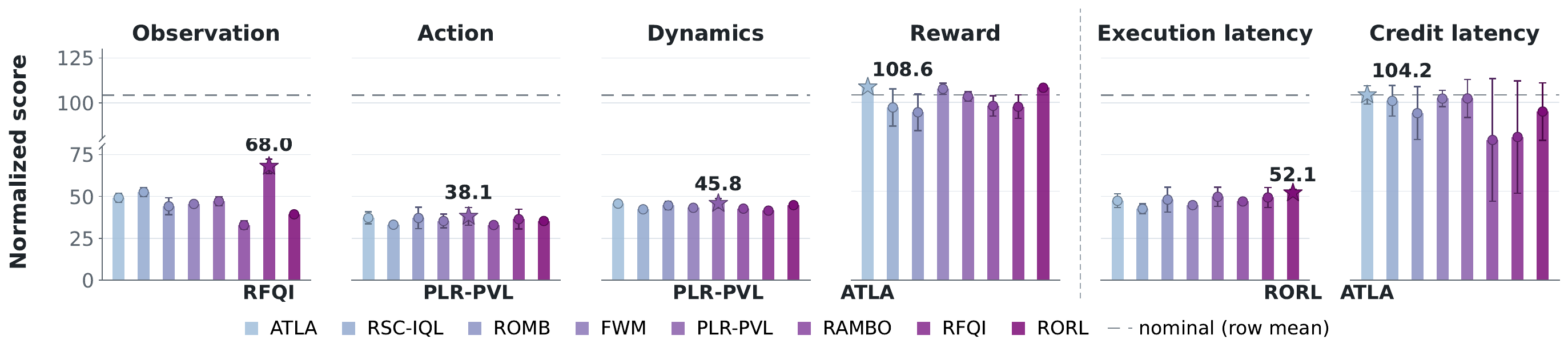}
  \caption{\textbf{Evaluation under diverse individual shift scenarios.}
The figure aggregates 14 robust and 2 standard RL methods across 44 distinct perturbation configurations. \emph{Semantic binding} is evaluated only online, on the
goal-conditioned causal door task. See Appendix~\ref{app:supplementary} for full results.}
    \label{fig:channel_sweeps}
\end{figure}

\paragraph{Claimed target shifts do not fully predict empirical strengths.}
Figure~\ref{fig:channel_sweeps} shows DR-SAC leads  dynamics and execution latency, while ATLA-SA-PPO leads observation
and action; offline leaders vary across shift sources. A method's intended target shift
therefore does not uniquely determine where it performs best. The measured
ranking is also conditional on the task, shift mode and specific perturbation used
to instantiate each shift.

\begin{figure}[t]
    \centering
    \begin{minipage}[t]{0.338\linewidth}
        \centering
        \includegraphics[width=\linewidth]{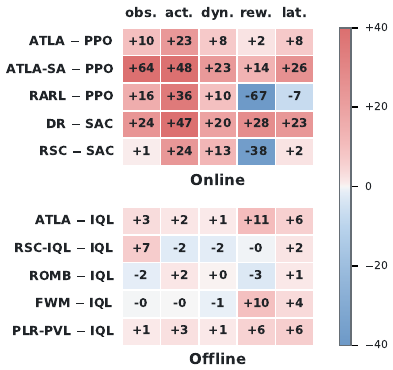}
        \caption{\textbf{Gains over base algorithms.} Each cell is the score difference between a robust method and its base algorithm, averaged over 44 perturbation configurations of five shift sources.\looseness=-1 }
        \label{fig:channel_paired_gain}
    \end{minipage}\hfill
    \begin{minipage}[t]{0.612\linewidth}
        \centering
        \includegraphics[width=\linewidth]{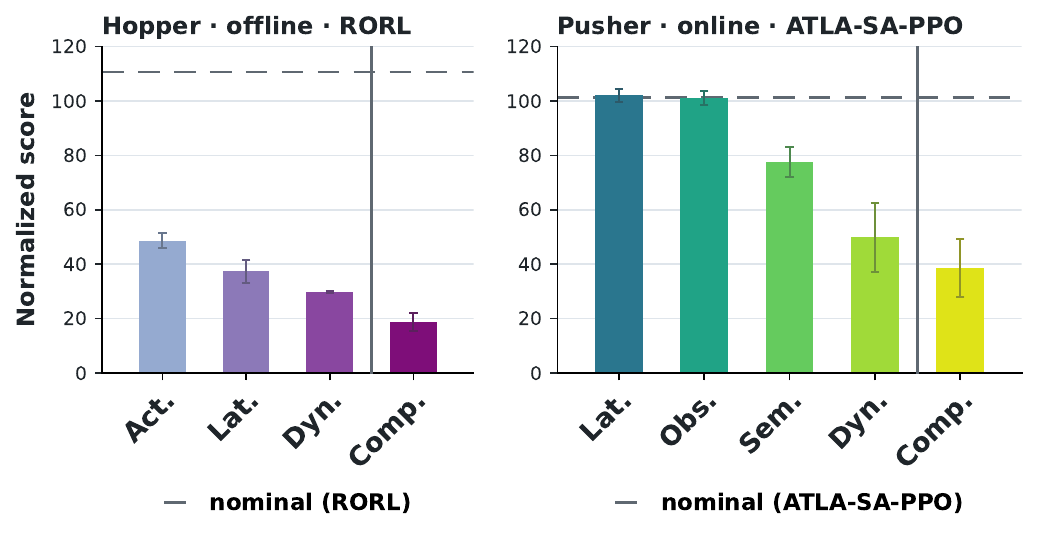}
        \caption{\textbf{Strongest algorithms performance under compound shifts from multiple sources.} Each panel shows the performance of the strongest offline (RORL) and online (ATLA-SA-PPO) robust algorithms under individual and compound shifts. The horizontal line marks each
algorithm's nominal score. See Appendix~\ref{app:supplementary} for full results.}
        \label{fig:compound_profiles}
    \end{minipage}
\end{figure}

\paragraph{Online robust methods generally achieve larger gains over their base algorithms than offline robust methods.}
Figure~\ref{fig:channel_paired_gain} shows that online methods can produce
large gains on most shift sources, whereas most offline differences remain close to their base algorithms. The smaller offline
increments may partly reflect limited coverage of the tested perturbation
modes in fixed datasets. Because offline RL relies on fixed data, methods often use conservative updates to limit value overestimation which potentially constrains robustness gains.

\paragraph{Dynamic shifts cause the largest performance drops, while the next most impactful shift varies by task.} We test our strongest online (ATLA-SA-PPO) and offline robust (RORL) algorithms under compound shifts.
Figure~\ref{fig:compound_profiles} shows that dynamic dominate both scenarios. Dynamic shift is the most damaging shift source in the two tasks, while the next most damaging differs by task. Locomotion loses about two-thirds of
its score to latency, because delayed action on a balancing gait is a fall
rather than a slower trajectory, while semantic shift on manipulation costs a quarter, because it moves the goal of the policy. Therefore, a sim-to-real design must identify which shift is core to
its task. Composition is worse than any of its parts, making both algorithms drop below half of nominal. This indicates the poor capability of algorithms in complex, multi-shift environments.

\subsection{Extension to advanced robotic tasks}
\label{sec:advanced_tasks}

We extend the benchmark from standard control to advanced robotics tasks, attaching the same disruptor interface through task-specific adapters, covering a high-dimensional humanoid, a contact-rich manipulator and a VLA control task.

\begin{figure}[h]
    \centering
    \includegraphics[width=0.95\linewidth]{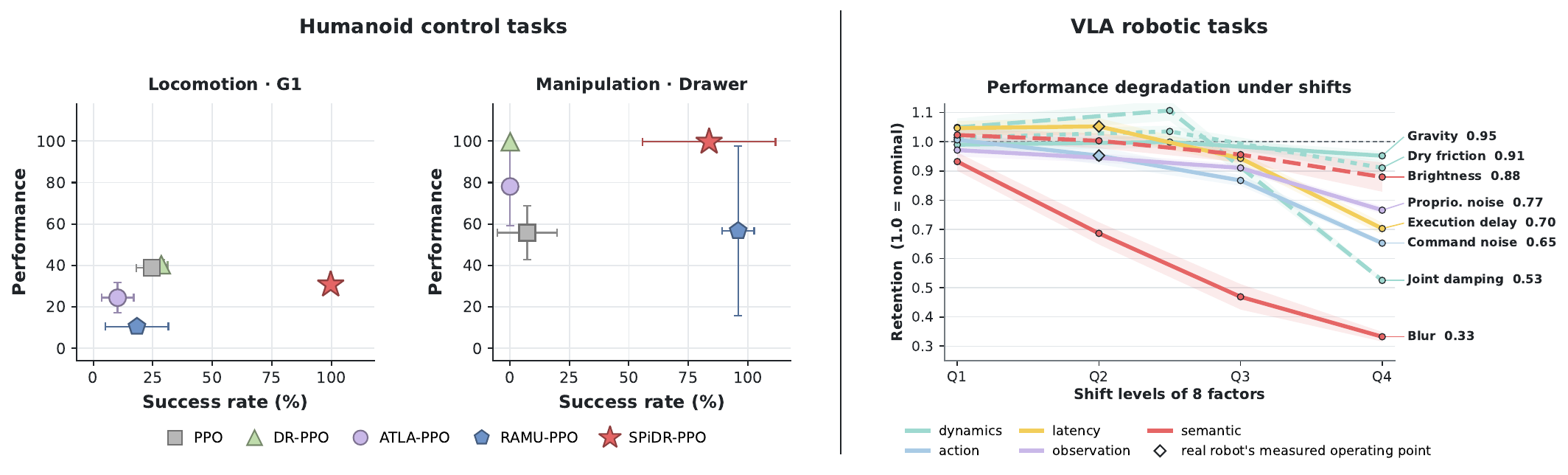}
    \caption{\textbf{Performance evaluation on Isaac Lab and VLA tasks.} \emph{Left:} each point is one algorithm,
plotting task performance against success rate on two tasks (G1: survival under action
delay; Drawer: episodes within the joint-limit cost budget). See Appendices~\ref{app:eval} and~\ref{sec:isaac_res} for the detailed setting and results. \emph{Right:} performance retention of a VLA policy as eight shift factors moves from Q1 to Q4. (Q1-Q4 definitions see Appendix~\ref{app:eval}). The diamond marks each factor's operating point measured on the real platform.}
    \label{fig:part4}
\end{figure}

\paragraph{Robust safe methods can ensure both performance and constraint satisfaction under shifts.}
We compare PPO, DR-PPO~\citep{schwarke2025rslrl} and ATLA-PPO against the robust safe methods RAMU-PPO and
SPiDR-PPO on Unitree G1 locomotion and Franka drawer manipulation in Isaac
Lab~\citep{mittal2025isaaclab}. Figure~\ref{fig:part4} shows  the two
highest-scoring policies on G1 survive about a quarter under action delay; while on the
drawer DR-PPO matches SPiDR-PPO's $99.6\%$ performance with $0\%$ of its episodes inside
the joint-limit cost budget against SPiDR-PPO's $84\%$. The results indicate that only robust safe methods especially SPiDR-PPO, keep both
metrics high on both tasks, so a higher performance is not the evidence that a
method is ready for sim-to-real deployment. 

\paragraph{Performance sensitivity differs sharply across shift sources, with semantic shifts more impactful in VLA tasks.}We fine-tune $\pi_{0.5}$
on the \emph{insert-mouse-battery} task (Figure~\ref{fig:vla_task})
and drive it under eight shift factors. At the largest measured level the policy
still remains $0.95$ of its nominal performance under gravity but only $0.33$ under image blur. Blur at its largest level moves only 13\% of the image but drops the policy to a third of nominal, while image brightness moves the same frames at seven times and costs almost nothing. Semantic shift is a core shift source which affects more than physical shifts since a clear scene is the base of the action chunk.  
\section{Conclusion}
\label{sec:conclusion}

In this work, we introduce \ouralg, an algorithm-centric library and evaluation
platform for robust RL under deployment shifts. It integrates 17 robust RL algorithms with 6 standard RL baselines, organized by
algorithm feature, claimed target shift, and base algorithm. A shared
interface evaluates MDP-element shifts alongside control-loop latency and task
semantics through individual and compound shift scenarios and extends to advanced robotics tasks. Under a unified evaluation over 273 perturbation configurations, robust methods as
a group outperform standard RL references. Individual and compound shift evaluations
further reveal each algorithmic feature's strengths and capability boundaries. Therefore, \ouralg provides a platform to support more comprehensive evaluation and guide the
design of future robust RL methods.

\bibliographystyle{apalike}
\bibliography{references}

@article{schulman2017proximal,
  title={Proximal policy optimization algorithms},
  author={Schulman, John and Wolski, Filip and Dhariwal, Prafulla and Radford, Alec and Klimov, Oleg},
  journal={arXiv preprint arXiv:1707.06347},
  year={2017}
}

@inproceedings{haarnoja2018soft,
  title={Soft actor-critic: Off-policy maximum entropy deep reinforcement learning with a stochastic actor},
  author={Haarnoja, Tuomas and Zhou, Aurick and Abbeel, Pieter and Levine, Sergey},
  booktitle={International conference on machine learning},
  pages={1861--1870},
  year={2018},
  organization={Pmlr}
}

@inproceedings{tobin2017domain,
  title={Domain randomization for transferring deep neural networks from simulation to the real world},
  author={Tobin, Josh and Fong, Rachel and Ray, Alex and Schneider, Jonas and Zaremba, Wojciech and Abbeel, Pieter},
  booktitle={2017 IEEE/RSJ international conference on intelligent robots and systems (IROS)},
  pages={23--30},
  year={2017},
  organization={IEEE}
}

@article{ding2023seeing,
  title={Seeing is not believing: Robust reinforcement learning against spurious correlation},
  author={Ding, Wenhao and Shi, Laixi and Chi, Yuejie and Zhao, Ding},
  journal={Advances in Neural Information Processing Systems},
  volume={36},
  pages={66328--66363},
  year={2023}
}

@inproceedings{pinto2017robust,
  title={Robust adversarial reinforcement learning},
  author={Pinto, Lerrel and Davidson, James and Sukthankar, Rahul and Gupta, Abhinav},
  booktitle={International conference on machine learning},
  pages={2817--2826},
  year={2017},
  organization={PMLR}
}

@article{kostrikov2021offline,
  title={Offline reinforcement learning with implicit q-learning},
  author={Kostrikov, Ilya and Nair, Ashvin and Levine, Sergey},
  journal={arXiv preprint arXiv:2110.06169},
  year={2021}
}

@article{yu2020mopo,
  title={Mopo: Model-based offline policy optimization},
  author={Yu, Tianhe and Thomas, Garrett and Yu, Lantao and Ermon, Stefano and Zou, James Y and Levine, Sergey and Finn, Chelsea and Ma, Tengyu},
  journal={Advances in neural information processing systems},
  volume={33},
  pages={14129--14142},
  year={2020}
}

@article{lu2023synthetic,
  title={Synthetic experience replay},
  author={Lu, Cong and Ball, Philip and Teh, Yee Whye and Parker-Holder, Jack},
  journal={Advances in Neural Information Processing Systems},
  volume={36},
  pages={46323--46344},
  year={2023}
}

@article{panaganti2022robust,
  title={Robust reinforcement learning using offline data},
  author={Panaganti, Kishan and Xu, Zaiyan and Kalathil, Dileep and Ghavamzadeh, Mohammad},
  journal={Advances in neural information processing systems},
  volume={35},
  pages={32211--32224},
  year={2022}
}

@article{rigter2022rambo,
  title={Rambo-rl: Robust adversarial model-based offline reinforcement learning},
  author={Rigter, Marc and Lacerda, Bruno and Hawes, Nick},
  journal={Advances in neural information processing systems},
  volume={35},
  pages={16082--16097},
  year={2022}
}

@article{yang2022rorl,
  title={Rorl: Robust offline reinforcement learning via conservative smoothing},
  author={Yang, Rui and Bai, Chenjia and Ma, Xiaoteng and Wang, Zhaoran and Zhang, Chongjie and Han, Lei},
  journal={Advances in neural information processing systems},
  volume={35},
  pages={23851--23866},
  year={2022}
}

@inproceedings{peng2018sim,
  title={Sim-to-real transfer of robotic control with dynamics randomization},
  author={Peng, Xue Bin and Andrychowicz, Marcin and Zaremba, Wojciech and Abbeel, Pieter},
  booktitle={2018 IEEE international conference on robotics and automation (ICRA)},
  pages={3803--3810},
  year={2018},
  organization={IEEE}
}

@article{gu2024humanoid,
  title={Humanoid-gym: Reinforcement learning for humanoid robot with zero-shot sim2real transfer},
  author={Gu, Xinyang and Wang, Yen-Jen and Chen, Jianyu},
  journal={arXiv preprint arXiv:2404.05695},
  year={2024}
}

@article{he2024omnih2o,
  title={Omnih2o: Universal and dexterous human-to-humanoid whole-body teleoperation and learning},
  author={He, Tairan and Luo, Zhengyi and He, Xialin and Xiao, Wenli and Zhang, Chong and Zhang, Weinan and Kitani, Kris and Liu, Changliu and Shi, Guanya},
  journal={arXiv preprint arXiv:2406.08858},
  year={2024}
}

@article{as2026spidr,
  title={Spidr: A simple approach for zero-shot safety in sim-to-real transfer},
  author={As, Yarden and Qu, Chengrui Ray and Unger, Benjamin and Kang, Dongho and van der Hart, Max and Shi, Laixi and Coros, Stelian and Wierman, Adam and Krause, Andreas},
  journal={Advances in Neural Information Processing Systems},
  volume={38},
  pages={90882--90924},
  year={2026}
}

@article{queeney2023risk,
  title={Risk-averse model uncertainty for distributionally robust safe reinforcement learning},
  author={Queeney, James and Benosman, Mouhacine},
  journal={Advances in Neural Information Processing Systems},
  volume={36},
  pages={1659--1680},
  year={2023}
}

@inproceedings{gu2025robust,
  title={Robust gymnasium: A unified modular benchmark for robust reinforcement learning},
  author={Gu, Shangding and Shi, Laixi and Wen, Muning and Jin, Ming and Mazumdar, Eric and Chi, Yuejie and Wierman, Adam and Spanos, Costas},
  booktitle={International Conference on Learning Representations},
  volume={2025},
  pages={102014--102041},
  year={2025}
}

@article{dulac2020empirical,
  title={An empirical investigation of the challenges of real-world reinforcement learning},
  author={Dulac-Arnold, Gabriel and Levine, Nir and Mankowitz, Daniel J and Li, Jerry and Paduraru, Cosmin and Gowal, Sven and Hester, Todd},
  journal={arXiv preprint arXiv:2003.11881},
  year={2020}
}

@article{fu2020d4rl,
  title={D4rl: Datasets for deep data-driven reinforcement learning},
  author={Fu, Justin and Kumar, Aviral and Nachum, Ofir and Tucker, George and Levine, Sergey},
  journal={arXiv preprint arXiv:2004.07219},
  year={2020}
}

@article{tan2018sim,
  title={Sim-to-real: Learning agile locomotion for quadruped robots},
  author={Tan, Jie and Zhang, Tingnan and Coumans, Erwin and Iscen, Atil and Bai, Yunfei and Hafner, Danijar and Bohez, Steven and Vanhoucke, Vincent},
  journal={arXiv preprint arXiv:1804.10332},
  year={2018}
}

@article{andrychowicz2020learning,
  title={Learning dexterous in-hand manipulation},
  author={Andrychowicz, OpenAI: Marcin and Baker, Bowen and Chociej, Maciek and Jozefowicz, Rafal and McGrew, Bob and Pachocki, Jakub and Petron, Arthur and Plappert, Matthias and Powell, Glenn and Ray, Alex and others},
  journal={The International Journal of Robotics Research},
  volume={39},
  number={1},
  pages={3--20},
  year={2020},
  publisher={SAGE Publications Sage UK: London, England}
}

@article{iyengar2005robust,
  title={Robust dynamic programming},
  author={Iyengar, Garud N},
  journal={Mathematics of Operations Research},
  volume={30},
  number={2},
  pages={257--280},
  year={2005},
  publisher={INFORMS}
}

@article{nilim2005robust,
  title={Robust control of Markov decision processes with uncertain transition matrices},
  author={Nilim, Arnab and El Ghaoui, Laurent},
  journal={Operations Research},
  volume={53},
  number={5},
  pages={780--798},
  year={2005},
  publisher={INFORMS}
}

@article{brohan2023rt,
  title={Rt-2: Vision-language-action models transfer web knowledge to robotic control},
  author={Brohan, Anthony and Brown, Noah and Carbajal, Justice and Chebotar, Yevgen and Chen, Xi and Choromanski, Krzysztof and Ding, Tianli and Driess, Danny and Dubey, Avinava and Finn, Chelsea and others},
  journal={arXiv preprint arXiv:2307.15818},
  year={2023}
}

@article{xu2010distributionally,
  title={Distributionally robust Markov decision processes},
  author={Xu, Huan and Mannor, Shie},
  journal={Advances in Neural Information Processing Systems},
  volume={23},
  year={2010}
}

@article{wiesemann2013robust,
  title={Robust Markov decision processes},
  author={Wiesemann, Wolfram and Kuhn, Daniel and Rustem, Ber{\c{c}}},
  journal={Mathematics of Operations Research},
  volume={38},
  number={1},
  pages={153--183},
  year={2013},
  publisher={INFORMS}
}

@inproceedings{tessler2019action,
  title={Action robust reinforcement learning and applications in continuous control},
  author={Tessler, Chen and Efroni, Yonathan and Mannor, Shie},
  booktitle={International Conference on Machine Learning},
  pages={6215--6224},
  year={2019},
  organization={PMLR}
}

@article{rajeswaran2016epopt,
  title={Epopt: Learning robust neural network policies using model ensembles},
  author={Rajeswaran, Aravind and Ghotra, Sarvjeet and Ravindran, Balaraman and Levine, Sergey},
  journal={arXiv preprint arXiv:1610.01283},
  year={2016}
}

@article{mankowitz2019robust,
  title={Robust reinforcement learning for continuous control with model misspecification},
  author={Mankowitz, Daniel J and Levine, Nir and Jeong, Rae and Shi, Yuanyuan and Kay, Jackie and Abdolmaleki, Abbas and Springenberg, Jost Tobias and Mann, Timothy and Hester, Todd and Riedmiller, Martin},
  journal={arXiv preprint arXiv:1906.07516},
  year={2019}
}

@article{zhang2020robust,
  title={Robust deep reinforcement learning against adversarial perturbations on state observations},
  author={Zhang, Huan and Chen, Hongge and Xiao, Chaowei and Li, Bo and Liu, Mingyan and Boning, Duane and Hsieh, Cho-Jui},
  journal={Advances in neural information processing systems},
  volume={33},
  pages={21024--21037},
  year={2020}
}

@article{oikarinen2021robust,
  title={Robust deep reinforcement learning through adversarial loss},
  author={Oikarinen, Tuomas and Zhang, Wang and Megretski, Alexandre and Daniel, Luca and Weng, Tsui-Wei},
  journal={Advances in Neural Information Processing Systems},
  volume={34},
  pages={26156--26167},
  year={2021}
}

@article{lecarpentier2019non,
  title={Non-stationary Markov decision processes, a worst-case approach using model-based reinforcement learning},
  author={Lecarpentier, Erwan and Rachelson, Emmanuel},
  journal={Advances in neural information processing systems},
  volume={32},
  year={2019}
}

@inproceedings{bouteiller2021reinforcement,
  title={Reinforcement Learning with Random Delays.},
  author={Bouteiller, Yann and Ramstedt, Simon and Beltrame, Giovanni and Pal, Christopher J and Binas, Jonathan},
  booktitle={ICLR},
  year={2021}
}

@article{bhardwaj2023adversarial,
  title={Adversarial model for offline reinforcement learning},
  author={Bhardwaj, Mohak and Xie, Tengyang and Boots, Byron and Jiang, Nan and Cheng, Ching-An},
  journal={Advances in Neural Information Processing Systems},
  volume={36},
  pages={1245--1269},
  year={2023}
}

@article{chen2025policy,
  title={Policy-driven world model adaptation for robust offline model-based reinforcement learning},
  author={Chen, Jiayu and Xu, Le and Venugopal, Aravind and Schneider, Jeff},
  journal={arXiv preprint arXiv:2505.13709},
  year={2025}
}

@article{ding2024diffusion,
  title={Diffusion world model: Future modeling beyond step-by-step rollout for offline reinforcement learning},
  author={Ding, Zihan and Zhang, Amy and Tian, Yuandong and Zheng, Qinqing},
  journal={arXiv preprint arXiv:2402.03570},
  year={2024}
}

@article{lipman2022flow,
  title={Flow matching for generative modeling},
  author={Lipman, Yaron and Chen, Ricky TQ and Ben-Hamu, Heli and Nickel, Maximilian and Le, Matt},
  journal={arXiv preprint arXiv:2210.02747},
  year={2022}
}

@inproceedings{berdica2024robust,
  title={Robust offline learning via adversarial world models},
  author={Berdica, Uljad and Li, Kelvin and Beukman, Michael and Goldie, Alexander David and Maiolino, Perla and Foerster, Jakob Nicolaus},
  booktitle={NeurIPS 2024 Workshop on Open-World Agents},
  year={2024}
}

@article{gulcehre2020rl,
  title={Rl unplugged: A suite of benchmarks for offline reinforcement learning},
  author={Gulcehre, Caglar and Wang, Ziyu and Novikov, Alexander and Paine, Thomas and G{\'o}mez, Sergio and Zolna, Konrad and Agarwal, Rishabh and Merel, Josh S and Mankowitz, Daniel J and Paduraru, Cosmin and others},
  journal={Advances in neural information processing systems},
  volume={33},
  pages={7248--7259},
  year={2020}
}

@article{kumar2021rma,
  title={Rma: Rapid motor adaptation for legged robots},
  author={Kumar, Ashish and Fu, Zipeng and Pathak, Deepak and Malik, Jitendra},
  journal={arXiv preprint arXiv:2107.04034},
  year={2021}
}

@article{fujimoto2021minimalist,
  title={A minimalist approach to offline reinforcement learning},
  author={Fujimoto, Scott and Gu, Shixiang Shane},
  journal={Advances in neural information processing systems},
  volume={34},
  pages={20132--20145},
  year={2021}
}

@article{black2026real,
  title={Real-time execution of action chunking flow policies},
  author={Black, Kevin and Galliker, Manuel and Levine, Sergey},
  journal={Advances in Neural Information Processing Systems},
  volume={38},
  pages={33383--33407},
  year={2026}
}

@article{lyu2024odrl,
  title={Odrl: A benchmark for off-dynamics reinforcement learning},
  author={Lyu, Jiafei and Xu, Kang and Xu, Jiacheng and Yan, Mengbei and Yang, Jingwen and Zhang, Zongzhang and Bai, Chenjia and Lu, Zongqing and Li, Xiu},
  journal={Advances in Neural Information Processing Systems},
  volume={37},
  pages={59859--59911},
  year={2024}
}

@article{zouitine2024rrls,
  title={Rrls: Robust reinforcement learning suite},
  author={Zouitine, Adil and Bertoin, David and Clavier, Pierre and Geist, Matthieu and Rachelson, Emmanuel},
  journal={arXiv preprint arXiv:2406.08406},
  year={2024}
}

@article{ji2023safety,
  title={Safety gymnasium: A unified safe reinforcement learning benchmark},
  author={Ji, Jiaming and Zhang, Borong and Zhou, Jiayi and Pan, Xuehai and Huang, Weidong and Sun, Ruiyang and Geng, Yiran and Zhong, Yifan and Dai, Josef and Yang, Yaodong},
  journal={Advances in Neural Information Processing Systems},
  volume={36},
  pages={18964--18993},
  year={2023}
}

@article{pumacay2024colosseum,
  title={The colosseum: A benchmark for evaluating generalization for robotic manipulation},
  author={Pumacay, Wilbert and Singh, Ishika and Duan, Jiafei and Krishna, Ranjay and Thomason, Jesse and Fox, Dieter},
  journal={arXiv preprint arXiv:2402.08191},
  year={2024}
}

@article{fei2025libero,
  title={Libero-plus: In-depth robustness analysis of vision-language-action models},
  author={Fei, Senyu and Wang, Siyin and Shi, Junhao and Dai, Zihao and Cai, Jikun and Qian, Pengfang and Ji, Li and He, Xinzhe and Zhang, Shiduo and Fei, Zhaoye and others},
  journal={arXiv preprint arXiv:2510.13626},
  year={2025}
}

@article{raffin2020generalized,
  title={Generalized state-dependent exploration for deep reinforcement learning in robotics},
  author={Raffin, Antonin and Stulp, Freek},
  journal={Arxiv},
  year={2020}
}

@inproceedings{schulman2015trust,
  title={Trust region policy optimization},
  author={Schulman, John and Levine, Sergey and Abbeel, Pieter and Jordan, Michael and Moritz, Philipp},
  booktitle={International conference on machine learning},
  pages={1889--1897},
  year={2015},
  organization={Pmlr}
}

@article{huang2017adversarial,
  title={Adversarial attacks on neural network policies},
  author={Huang, Sandy and Papernot, Nicolas and Goodfellow, Ian and Duan, Yan and Abbeel, Pieter},
  journal={arXiv preprint arXiv:1702.02284},
  year={2017}
}

@inproceedings{tan2020robustifying,
  title={Robustifying reinforcement learning agents via action space adversarial training},
  author={Tan, Kai Liang and Esfandiari, Yasaman and Lee, Xian Yeow and Sarkar, Soumik and others},
  booktitle={2020 American control conference (ACC)},
  pages={3959--3964},
  year={2020},
  organization={IEEE}
}

@article{goyal2023robust,
  title={Robust markov decision processes: Beyond rectangularity},
  author={Goyal, Vineet and Grand-Clement, Julien},
  journal={Mathematics of Operations Research},
  volume={48},
  number={1},
  pages={203--226},
  year={2023},
  publisher={INFORMS}
}

@article{ma2019policy,
  title={Policy poisoning in batch reinforcement learning and control},
  author={Ma, Yuzhe and Zhang, Xuezhou and Sun, Wen and Zhu, Jerry},
  journal={Advances in Neural Information Processing Systems},
  volume={32},
  year={2019}
}

@inproceedings{mandlekar2017adversarially,
  title={Adversarially robust policy learning: Active construction of physically-plausible perturbations},
  author={Mandlekar, Ajay and Zhu, Yuke and Garg, Animesh and Fei-Fei, Li and Savarese, Silvio},
  booktitle={2017 IEEE/RSJ International Conference on Intelligent Robots and Systems (IROS)},
  pages={3932--3939},
  year={2017},
  organization={IEEE}
}

@article{tanabe2022max,
  title={Max-min off-policy actor-critic method focusing on worst-case robustness to model misspecification},
  author={Tanabe, Takumi and Sato, Rei and Fukuchi, Kazuto and Sakuma, Jun and Akimoto, Youhei},
  journal={Advances in neural information processing systems},
  volume={35},
  pages={6967--6981},
  year={2022}
}

@article{liang2022efficient,
  title={Efficient adversarial training without attacking: Worst-case-aware robust reinforcement learning},
  author={Liang, Yongyuan and Sun, Yanchao and Zheng, Ruijie and Huang, Furong},
  journal={Advances in neural information processing systems},
  volume={35},
  pages={22547--22561},
  year={2022}
}

@article{sun2024breaking,
  title={Breaking the barrier: Enhanced utility and robustness in smoothed DRL agents},
  author={Sun, Chung-En and Gao, Sicun and Weng, Tsui-Wei},
  journal={arXiv preprint arXiv:2406.18062},
  year={2024}
}

@article{liu2022robustness,
  title={On the robustness of safe reinforcement learning under observational perturbations},
  author={Liu, Zuxin and Guo, Zijian and Cen, Zhepeng and Zhang, Huan and Tan, Jie and Li, Bo and Zhao, Ding},
  journal={arXiv preprint arXiv:2205.14691},
  year={2022}
}

@article{wang2022robust,
  title={Robust constrained reinforcement learning},
  author={Wang, Yue and Miao, Fei and Zou, Shaofeng},
  journal={arXiv preprint arXiv:2209.06866},
  year={2022}
}

@article{foffano2026adversarial,
  title={Adversarial diffusion for robust reinforcement learning},
  author={Foffano, Daniele and Russo, Alessio and Proutiere, Alexandre},
  journal={Advances in Neural Information Processing Systems},
  volume={38},
  pages={68627--68658},
  year={2026}
}

@article{yang2024uncertainty,
  title={Uncertainty-based offline variational bayesian reinforcement learning for robustness under diverse data corruptions},
  author={Yang, Rui and Wang, Jie and Wu, Guoping and Li, Bin},
  journal={Advances in Neural Information Processing Systems},
  volume={37},
  pages={39748--39783},
  year={2024}
}

@article{blanchet2023double,
  title={Double pessimism is provably efficient for distributionally robust offline reinforcement learning: Generic algorithm and robust partial coverage},
  author={Blanchet, Jose and Lu, Miao and Zhang, Tong and Zhong, Han},
  journal={Advances in Neural Information Processing Systems},
  volume={36},
  pages={66845--66859},
  year={2023}
}

@article{panaganti2023bridging,
  title={Bridging distributionally robust learning and offline rl: An approach to mitigate distribution shift and partial data coverage},
  author={Panaganti, Kishan and Xu, Zaiyan and Kalathil, Dileep and Ghavamzadeh, Mohammad},
  journal={arXiv preprint arXiv:2310.18434},
  year={2023}
}

@article{li2024towards,
  title={Towards optimal adversarial robust q-learning with bellman infinity-error},
  author={Li, Haoran and Zhang, Zicheng and Luo, Wang and Han, Congying and Hu, Yudong and Guo, Tiande and Liao, Shichen},
  journal={arXiv preprint arXiv:2402.02165},
  year={2024}
}

@inproceedings{liang2018rllib,
  title={RLlib: Abstractions for distributed reinforcement learning},
  author={Liang, Eric and Liaw, Richard and Nishihara, Robert and Moritz, Philipp and Fox, Roy and Goldberg, Ken and Gonzalez, Joseph and Jordan, Michael and Stoica, Ion},
  booktitle={International conference on machine learning},
  pages={3053--3062},
  year={2018},
  organization={PMLR}
}

@article{hu2023marllib,
  title={Marllib: A scalable and efficient multi-agent reinforcement learning library},
  author={Hu, Siyi and Zhong, Yifan and Gao, Minquan and Wang, Weixun and Dong, Hao and Liang, Xiaodan and Li, Zhihui and Chang, Xiaojun and Yang, Yaodong},
  journal={Journal of Machine Learning Research},
  volume={24},
  number={315},
  pages={1--23},
  year={2023}
}

@inproceedings{duan2016benchmarking,
  title={Benchmarking deep reinforcement learning for continuous control},
  author={Duan, Yan and Chen, Xi and Houthooft, Rein and Schulman, John and Abbeel, Pieter},
  booktitle={International conference on machine learning},
  pages={1329--1338},
  year={2016},
  organization={PMLR}
}

@inproceedings{yu2020meta,
  title={Meta-world: A benchmark and evaluation for multi-task and meta reinforcement learning},
  author={Yu, Tianhe and Quillen, Deirdre and He, Zhanpeng and Julian, Ryan and Hausman, Karol and Finn, Chelsea and Levine, Sergey},
  booktitle={Conference on robot learning},
  pages={1094--1100},
  year={2020},
  organization={PMLR}
}

@article{sferrazza2024humanoidbench,
  title={Humanoidbench: Simulated humanoid benchmark for whole-body locomotion and manipulation},
  author={Sferrazza, Carmelo and Huang, Dun-Ming and Lin, Xingyu and Lee, Youngwoon and Abbeel, Pieter},
  journal={arXiv preprint arXiv:2403.10506},
  year={2024}
}

@article{mohan2026road,
  title={RoAd-RL: A Unified Library and Benchmark for Robust Adversarial Reinforcement Learning},
  author={Mohan, Adithya and Kriegl, Daniel and Sch{\"o}n, Torsten},
  journal={arXiv preprint arXiv:2606.29867},
  year={2026}
}

@article{qin2022neorl,
  title={NeoRL: A near real-world benchmark for offline reinforcement learning},
  author={Qin, Rong-Jun and Zhang, Xingyuan and Gao, Songyi and Chen, Xiong-Hui and Li, Zewen and Zhang, Weinan and Yu, Yang},
  journal={Advances in Neural Information Processing Systems},
  volume={35},
  pages={24753--24765},
  year={2022}
}

@article{raffin2021stable,
  title={Stable-baselines3: Reliable reinforcement learning implementations},
  author={Raffin, Antonin and Hill, Ashley and Gleave, Adam and Kanervisto, Anssi and Ernestus, Maximilian and Dormann, Noah},
  journal={Journal of machine learning research},
  volume={22},
  number={268},
  pages={1--8},
  year={2021}
}

@article{huang2022cleanrl,
  title={Cleanrl: High-quality single-file implementations of deep reinforcement learning algorithms},
  author={Huang, Shengyi and Dossa, Rousslan Fernand Julien and Ye, Chang and Braga, Jeff and Chakraborty, Dipam and Mehta, Kinal and Ara{\~A}{\v{s}}jo, Jo{\~A}{\c{G}}o GM},
  journal={Journal of Machine Learning Research},
  volume={23},
  number={274},
  pages={1--18},
  year={2022}
}

@article{weng2022tianshou,
  title={Tianshou: A highly modularized deep reinforcement learning library},
  author={Weng, Jiayi and Chen, Huayu and Yan, Dong and You, Kaichao and Duburcq, Alexis and Zhang, Minghao and Su, Yi and Su, Hang and Zhu, Jun},
  journal={Journal of Machine Learning Research},
  volume={23},
  number={267},
  pages={1--6},
  year={2022}
}

@article{mittal2025isaaclab,
  title={Isaac Lab: A GPU-Accelerated Simulation Framework for Multi-Modal Robot Learning},
  author={Mayank Mittal and Pascal Roth and James Tigue and Antoine Richard and Octi Zhang and Peter Du and Antonio Serrano-Muñoz and Xinjie Yao and René Zurbrügg and Nikita Rudin and Lukasz Wawrzyniak and Milad Rakhsha and Alain Denzler and Eric Heiden and Ales Borovicka and Ossama Ahmed and Iretiayo Akinola and Abrar Anwar and Mark T. Carlson and Ji Yuan Feng and Animesh Garg and Renato Gasoto and Lionel Gulich and Yijie Guo and M. Gussert and Alex Hansen and Mihir Kulkarni and Chenran Li and Wei Liu and Viktor Makoviychuk and Grzegorz Malczyk and Hammad Mazhar and Masoud Moghani and Adithyavairavan Murali and Michael Noseworthy and Alexander Poddubny and Nathan Ratliff and Welf Rehberg and Clemens Schwarke and Ritvik Singh and James Latham Smith and Bingjie Tang and Ruchik Thaker and Matthew Trepte and Karl Van Wyk and Fangzhou Yu and Alex Millane and Vikram Ramasamy and Remo Steiner and Sangeeta Subramanian and Clemens Volk and CY Chen and Neel Jawale and Ashwin Varghese Kuruttukulam and Michael A. Lin and Ajay Mandlekar and Karsten Patzwaldt and John Welsh and Huihua Zhao and Fatima Anes and Jean-Francois Lafleche and Nicolas Moënne-Loccoz and Soowan Park and Rob Stepinski and Dirk Van Gelder and Chris Amevor and Jan Carius and Jumyung Chang and Anka He Chen and Pablo de Heras Ciechomski and Gilles Daviet and Mohammad Mohajerani and Julia von Muralt and Viktor Reutskyy and Michael Sauter and Simon Schirm and Eric L. Shi and Pierre Terdiman and Kenny Vilella and Tobias Widmer and Gordon Yeoman and Tiffany Chen and Sergey Grizan and Cathy Li and Lotus Li and Connor Smith and Rafael Wiltz and Kostas Alexis and Yan Chang and David Chu and Linxi "Jim" Fan and Farbod Farshidian and Ankur Handa and Spencer Huang and Marco Hutter and Yashraj Narang and Soha Pouya and Shiwei Sheng and Yuke Zhu and Miles Macklin and Adam Moravanszky and Philipp Reist and Yunrong Guo and David Hoeller and Gavriel State},
  journal={arXiv preprint arXiv:2511.04831},
  year={2025},
  url={https://arxiv.org/abs/2511.04831}
}

@misc{posttraining_robotics_2026,
  title        = {Post-Training for Robotics Foundation Models Dataset and Challenge},
  author       = {Zhang, Shiduo and Wang, Yue and Chang, Haonan and Zhao, Hang and
                  Liu, Yicheng and Guizilini, Vitor and Bobu, Andreea and Wagenmaker, Andrew and
                  Dixit, Anushri and Yu, Chao and Shah, Dhruv and Simchowitz, Max},
  year         = {2026},
  howpublished = {RSS 2026 Workshop \& Challenge},
  url          = {https://posttraining-for-robotics.github.io}
}

@misc{intelligence2025pi05visionlanguageactionmodelopenworld,
      title={$\pi_{0.5}$: a Vision-Language-Action Model with Open-World Generalization}, 
      author={Physical Intelligence and Kevin Black and Noah Brown and James Darpinian and Karan Dhabalia and Danny Driess and Adnan Esmail and Michael Equi and Chelsea Finn and Niccolo Fusai and Manuel Y. Galliker and Dibya Ghosh and Lachy Groom and Karol Hausman and Brian Ichter and Szymon Jakubczak and Tim Jones and Liyiming Ke and Devin LeBlanc and Sergey Levine and Adrian Li-Bell and Mohith Mothukuri and Suraj Nair and Karl Pertsch and Allen Z. Ren and Lucy Xiaoyang Shi and Laura Smith and Jost Tobias Springenberg and Kyle Stachowicz and James Tanner and Quan Vuong and Homer Walke and Anna Walling and Haohuan Wang and Lili Yu and Ury Zhilinsky},
      year={2025},
      eprint={2504.16054},
      archivePrefix={arXiv},
      primaryClass={cs.LG},
      url={https://arxiv.org/abs/2504.16054}, 
}

@article{ding2023survey,
  title={A survey on safety-critical driving scenario generation—A methodological perspective},
  author={Ding, Wenhao and Xu, Chejian and Arief, Mansur and Lin, Haohong and Li, Bo and Zhao, Ding},
  journal={IEEE Transactions on Intelligent Transportation Systems},
  volume={24},
  number={7},
  pages={6971--6988},
  year={2023},
  publisher={IEEE}
}

@article{liu2015robustness,
  title={On robustness of noninferiority clinical trial designs against bias, variability, and nonconstancy},
  author={Liu, Qing and Li, Yulan and Odem-Davis, Katherine},
  journal={Journal of biopharmaceutical statistics},
  volume={25},
  number={1},
  pages={206--225},
  year={2015},
  publisher={Taylor \& Francis}
}

@inproceedings{
zhang2021robust,
title={Robust Reinforcement Learning on State Observations with Learned Optimal Adversary},
author={Huan Zhang and Hongge Chen and Duane S Boning and Cho-Jui Hsieh},
booktitle={International Conference on Learning Representations},
year={2021},
url={https://openreview.net/forum?id=sCZbhBvqQaU}
}

@article{han2022solution,
  title={What is the Solution for State Adversarial Multi-Agent Reinforcement Learning?},
  author={Han, Songyang and Su, Sanbao and He, Sihong and Han, Shuo and Yang, Haizhao and Miao, Fei},
  journal={arXiv preprint arXiv:2212.02705},
  year={2022}
}

@article{sun2021exploring,
  title={Exploring the Training Robustness of Distributional Reinforcement Learning against Noisy State Observations},
  author={Sun, Ke and Liu, Yi and Zhao, Yingnan and Yao, Hengshuai and Jui, Shangling and Kong, Linglong},
  journal={arXiv preprint arXiv:2109.08776},
  year={2021}
}

@article{xiong2022defending,
  title={Defending Observation Attacks in Deep Reinforcement Learning via Detection and Denoising},
  author={Xiong, Zikang and Eappen, Joe and Zhu, He and Jagannathan, Suresh},
  journal={arXiv preprint arXiv:2206.07188},
  year={2022}
}

@article{klopp2017robust,
  title={Robust matrix completion},
  author={Klopp, Olga and Lounici, Karim and Tsybakov, Alexandre B},
  journal={Probability Theory and Related Fields},
  volume={169},
  number={1-2},
  pages={523--564},
  year={2017},
  publisher={Springer}
}

@inproceedings{mahmood2018benchmarking,
  title={Benchmarking reinforcement learning algorithms on real-world robots},
  author={Mahmood, A Rupam and Korenkevych, Dmytro and Vasan, Gautham and Ma, William and Bergstra, James},
  booktitle={Conference on robot learning},
  pages={561--591},
  year={2018},
  organization={PMLR}
}

@article{bertsimas2019adaptive,
  title={Adaptive distributionally robust optimization},
  author={Bertsimas, Dimitris and Sim, Melvyn and Zhang, Meilin},
  journal={Management Science},
  volume={65},
  number={2},
  pages={604--618},
  year={2019},
  publisher={INFORMS}
}

@article{tversky1974judgment,
  title={Judgment under Uncertainty: Heuristics and Biases: Biases in judgments reveal some heuristics of thinking under uncertainty.},
  author={Tversky, Amos and Kahneman, Daniel},
  journal={science},
  volume={185},
  number={4157},
  pages={1124--1131},
  year={1974},
  publisher={American association for the advancement of science}
}

@article{arthur1991designing,
  title={Designing economic agents that act like human agents: A behavioral approach to bounded rationality},
  author={Arthur, W Brian},
  journal={The American economic review},
  volume={81},
  number={2},
  pages={353--359},
  year={1991},
  publisher={JSTOR}
}

@inproceedings{li2021reinforcement,
  title={Reinforcement learning for robust parameterized locomotion control of bipedal robots},
  author={Li, Zhongyu and Cheng, Xuxin and Peng, Xue Bin and Abbeel, Pieter and Levine, Sergey and Berseth, Glen and Sreenath, Koushil},
  booktitle={2021 IEEE International Conference on Robotics and Automation (ICRA)},
  pages={2811--2817},
  year={2021},
  organization={IEEE}
}

@article{kozak2023stability,
  title={Stability, reliability, and robustness of GaN power devices: A review},
  author={Kozak, Joseph Peter and Zhang, Ruizhe and Porter, Matthew and Song, Qihao and Liu, Jingcun and Wang, Bixuan and Wang, Rudy and Saito, Wataru and Zhang, Yuhao},
  journal={IEEE Transactions on Power Electronics},
  volume={38},
  number={7},
  pages={8442--8471},
  year={2023},
  publisher={IEEE}
}

@misc{minari,
  author = {Younis, Omar G. and Perez-Vicente, Rodrigo and Balis, John U. and Dudley, Will and Davey, Alex and Terry, Jordan K},
  title = {{Minari}},
  publisher = {Zenodo},
  year = {2024},
  doi = {10.5281/zenodo.13767625},
  url = {https://doi.org/10.5281/zenodo.13767625},
  note = {Software, version 0.5.0}
}

@article{lu2021revisiting,
  title={Revisiting design choices in offline model-based reinforcement learning},
  author={Lu, Cong and Ball, Philip J and Parker-Holder, Jack and Osborne, Michael A and Roberts, Stephen J},
  journal={arXiv preprint arXiv:2110.04135},
  year={2021}
}

@article{schwarke2025rslrl,
  title={RSL-RL: A Learning Library for Robotics Research},
  author={Schwarke, Clemens and Mittal, Mayank and Rudin, Nikita and Hoeller, David and Hutter, Marco},
  journal={arXiv preprint arXiv:2509.10771},
  year={2025}
}

@article{shi2026curious,
  title={The Curious Price of Distributional Robustness in Reinforcement Learning with a Generative Model},
  author={Shi, Laixi and Li, Gen and Wei, Yuting and Chen, Yuxin and Geist, Matthieu and Chi, Yuejie},
  journal={Operations Research},
  volume={74},
  number={5},
  pages={2438--2459},
  year={2026},
  publisher={INFORMS}
}

@article{wang2024sample,
  title={Sample complexity of offline distributionally robust linear markov decision processes},
  author={Wang, He and Shi, Laixi and Chi, Yuejie},
  journal={arXiv preprint arXiv:2403.12946},
  year={2024}
}

@article{shi2024distributionally,
  title={Distributionally robust model-based offline reinforcement learning with near-optimal sample complexity},
  author={Shi, Laixi and Chi, Yuejie},
  journal={Journal of Machine Learning Research},
  volume={25},
  number={200},
  pages={1--91},
  year={2024}
}

@misc{sharma2026demonstrationobjectsgeneralizingmanipulation,
      title={One Demonstration, Many Objects: Generalizing Manipulation via Local Contact Geometry}, 
      author={Satvik Sharma and Samrat Sahoo and Huang Huang and Fei-Fei Li and Jiajun Wu and Dorsa Sadigh and Jeannette Bohg},
      year={2026},
      eprint={2609.01938},
      archivePrefix={arXiv},
      primaryClass={cs.RO},
      url={https://arxiv.org/abs/2609.01938}, 
}

\appendix

\section{Benchmark construction}
\label{sec:benchmark}

\subsection{Benchmark Comparison}
\label{app:benchmark_comparison}

\providecommand{\cmpYes}{\textcolor{green!70!black}{\ding{51}}}
\providecommand{\cmpPart}{\ding{109}}
\providecommand{\cmpNo}{\ding{55}}

\newcolumntype{S}{>{\centering\arraybackslash}p{0.0925\textwidth}}
\newcolumntype{W}{>{\centering\arraybackslash}p{0.1006\textwidth}}
\newcolumntype{A}{>{\columncolor{red!8}\centering\arraybackslash}m{0.13\textwidth}}

\begin{table*}[h]
\centering
\footnotesize
\setlength{\tabcolsep}{2pt}
\renewcommand{\arraystretch}{1.18}
\newcommand{\tablehead}[1]{{\scriptsize\bfseries #1}}
\begin{tabular}{@{}
    >{\raggedright\arraybackslash}p{0.3285\textwidth}
    A *{4}{S} W@{}}
\toprule
\multicolumn{1}{c}{\multirow{2}{*}{\tablehead{Benchmark}}}
&
& \multicolumn{4}{c}{\tablehead{Shifts}}
& \multirow{2}{*}{\shortstack[c]{
    \tablehead{Task}\\
    \tablehead{expansion}
  }}
\\
\cline{3-6}
&
\multirow{-2}{*}{\shortstack[c]{
    \tablehead{Robust}\\
    \tablehead{algorithms}
  }}
& \tablehead{MDP}
& \tablehead{Latency}
& \tablehead{Semantic}
& \tablehead{Compound}
&
\\
\midrule

{\scriptsize RLlib \citep{liang2018rllib}}
& 0
& 0/4 & \cmpNo & \cmpNo & \cmpNo
& \cmpYes
\\

{\scriptsize RRLS \citep{zouitine2024rrls}}
& 4
& 2/4 & \cmpNo & \cmpNo & \cmpPart
& \cmpYes
\\

{\scriptsize ODRL \citep{lyu2024odrl}}
& 0
& 1/4 & \cmpNo & \cmpNo & \cmpNo
& \cmpNo
\\

{\scriptsize Robust-Gymnasium \citep{gu2025robust}}
& 4
& 4/4 & \cmpNo & \cmpYes & \cmpYes
& \cmpYes
\\

{\scriptsize RWRL Suite \citep{dulac2020empirical}}
& 0
& 3/4 & \cmpYes & \cmpNo & \cmpYes
& \cmpPart
\\

{\scriptsize RoAd-RL \citep{mohan2026road}}
& 0
& 1/4 & \cmpNo & \cmpNo & \cmpNo
& \cmpYes
\\

\rowcolor{blue!6}
{\scriptsize\bfseries \ouralg{} (ours)}
& \cellcolor{red!14}\textbf{17}
& \textbf{4/4}
& \cmpYes
& \cmpYes
& \cmpYes
& \cmpYes
\\

\bottomrule
\end{tabular}

\caption{\textbf{Comparison with RL libraries and robustness benchmarks.}
\emph{Robust algorithms} counts robust and robust safe RL algorithms evaluated by each
benchmark. A checkmark denotes explicit,
evaluated support; \cmpPart\ denotes partial  support;
\cmpNo\ indicates that the capability is absent or not demonstrated in the
cited release.
\emph{MDP} counts, out of four, the MDP components on which shift is
evaluated: observations, actions, transitions, and reward/cost signals.
\emph{Compound} requires multiple shift sources to be active jointly.
\emph{Task expansion} indicates an extensible interface for carrying the
benchmark to additional tasks.}
\label{tab:benchmark_comparison}
\end{table*}

Table~\ref{tab:benchmark_comparison} compares evaluated capabilities with 6 related benchmarks.
Algorithm counts include robust and robust safe RL method. Shift entries require reported evaluations, whereas
task expansion denotes a documented interface for adding task backends.
These criteria distinguish a benchmark's demonstrated scope from its
potential extensibility.

The compared systems standardize different parts of robust RL evaluation.
RLlib provides general-purpose algorithm infrastructure; RRLS and ODRL
compare methods primarily under dynamics-related uncertainty
\citep{liang2018rllib,zouitine2024rrls,lyu2024odrl}.
Robust-Gymnasium offers broad disrupted environments, including semantic
tasks, while the RWRL Suite evaluates practical stressors such as latency
and their combination
\citep{gu2025robust,dulac2020empirical}.
RoAd-RL focuses on observation attacks and inference-time defenses
\citep{mohan2026road}.
Rather than claiming these capabilities individually as new, \ouralg
combines a structured robust online, offline, and safe  RL library with
isolated and composed shift evaluations under a shared protocol. Its
task-specific adapters further demonstrate how the evaluation methodology
extends from standard control to advanced robotics tasks such as Isaac Lab and a VLA pipeline.

\subsection{Disruptor toolbox: shift modes, operators and implementations}
\label{app:benchmark_construction}

A perturbation configuration is specified by
(\textbf{source}, \textbf{factor}, \textbf{mode}, \textbf{magnitude}, \textbf{resampling rule}, \textbf{optional schedule}, \textbf{seed}).
The \textbf{source} identifies where the intervention enters the agent--environment
interaction (Section~\ref{sec:disruptor}), the \textbf{factor} names the perturbed
quantity (e.g., gravity, friction, or an observation channel), the \textbf{mode}
specifies how the intervention is constructed, and the \textbf{magnitude} sets how far
it departs from nominal in the operator's natural units. The \textbf{resampling rule}
determines whether a value is redrawn at each step, held for an episode, or fixed.
The optional \textbf{schedule} controls when the intervention is active
and how its magnitude evolves, while the \textbf{seed} makes stochastic
draws reproducible. The main text summarizes five shift modes;
here we detail their source-specific modes, indexed in
Table~\ref{tab:shift_operators}.

\begin{table*}[!t]
\centering
\footnotesize
\setlength{\tabcolsep}{4pt}
\renewcommand{\arraystretch}{1.12}
\begin{tabularx}{\textwidth}{@{}
    >{\raggedright\arraybackslash}p{0.2\textwidth}
    >{\raggedright\arraybackslash}p{0.15\textwidth}
    >{\raggedright\arraybackslash}X
    >{\raggedright\arraybackslash}p{0.22\textwidth}@{}}
\toprule
Source & Operator & Intervention & Magnitude and resampling \\
\midrule
\multicolumn{4}{@{}l}{\textit{Stochastic}} \\
Observation, action, reward/cost
& \textbf{gauss}, \textbf{uniform}
& Add componentwise signal noise
& Standard deviation or sampling range; per step \\
Observation
& \textbf{relative}
& Scale noise by each feature's reference spread, $\sigma_i=c\hat{\sigma}_i$
& Relative scale $c$; per step \\
Observation
& \textbf{bias}
& Draw and hold a sensor offset
& Offset range; per episode \\
Dynamics
& \textbf{gauss}, \textbf{uniform}, \textbf{loguniform}
& Sample physical parameters or their multipliers
& Parameter range; per reset \\
Latency
& \textbf{buffer}, \textbf{interp}
& Delay a buffered command or interpolate timestamped observations
& Control steps or milliseconds; configured refresh \\
Latency
& \textbf{substep}
& Vary how many simulator substeps hold one policy action
& Substeps; per policy decision \\
\midrule
\multicolumn{4}{@{}l}{\textit{Adversarial}} \\
Observation
& \textbf{adversarial}
& Search for a policy-dependent perturbation within an $\ell_\infty$ ball
& Radius $\epsilon$; per step \\
Action
& \textbf{oppose}
& Subtract a bounded component along the chosen action
& Opposition magnitude; per step \\
Action
& \textbf{rotate}
& Change action direction while preserving magnitude; 2-D actions only
& Rotation angle; per step \\
Action
& \textbf{oppose\_goal}
& Remove part of the goal-directed action component
& Opposition strength; per step \\
\midrule
\multicolumn{4}{@{}l}{\textit{Parametric}} \\
Observation, action, reward/cost
& \textbf{shift}
& Add a specified signal offset
& Signal units; fixed \\
Dynamics
& \textbf{scale}, \textbf{set}
& Multiply or assign a named model parameter
& Ratio or parameter value; fixed \\
Dynamics, semantics
& \textbf{translate}
& Move a named body, object, or goal
& Metric displacement; fixed \\
Semantics
& \textbf{hue}, \textbf{tint}, \textbf{swap},
  \textbf{brightness}, \textbf{blur}, \textbf{translate}
& Change appearance or relocate a named goal/object
& Hue angle, tint colour, named swap, brightness factor,
  blur radius (px), or displacement (m); fixed \\
\midrule
\multicolumn{4}{@{}l}{\textit{Non-stationary}} \\
Any compatible source
& \textbf{schedule}
& Vary an operator's severity by $g(t)$
& Ramp, step, or sinusoidal profile \\
Dynamics
& \textbf{push}
& Apply an intermittent force to a named body
& Force, duration, and interval; per event \\
Reward/cost
& \textbf{delay}
& Release accumulated feedback every $k$ steps or shift it by $k$ steps
& Delay $k$; flush at episode end \\
\midrule
\multicolumn{4}{@{}l}{\textit{Composition}} \\
Compatible sources
& \textbf{ordered list}
& Stack specifications with independent random streams
& Explicit order and seeds \\
\bottomrule
\end{tabularx}
\caption{\textbf{Shift operator index.}
The first three families construct interventions; non-stationary and
composition determine how interventions evolve and combine. Each row gives
the compatible source, mode, intervention, and magnitude or resampling rule.}
\label{tab:shift_operators}
\end{table*}

\subsubsection{Shift source: Latency}
\label{app:latency}
We treat latency as a composable deployment-shift source within this unified
robust RL evaluation framework. Specifically, we expose observation delay,
action delay, and control-period variation through
$\tilde o_t=\mathcal{I}(\mathcal{H}_t,t-\delta_o),
\tilde a_t=a_{t-\delta_a},
s_{t+1}\sim P_0^{(\kappa_t)}(\cdot\mid s_t,\tilde a_t)$,
capturing delayed sensing, delayed execution, and variable control rates,
respectively. The three operators act at different points of one decision and
are implemented accordingly. 

\begin{itemize}[leftmargin=1.4em,itemsep=2pt,topsep=3pt]
    \item \textbf{Observation latency.}
    A timestamped history $\mathcal H_t$ is queried at $t-\delta_o$ using
    linear interpolation, so sub-step delays are not rounded away
    \citep{tan2018sim}. If the requested time precedes reset, the history
    returns the reset observation.

    \item \textbf{Action latency.}
    An integer buffer executes the command from $\delta_a$ control steps
    earlier, $\tilde a_t=a_{\max(t-\delta_a,0)}$, using the initial command
    to warm-start the buffer.

    \item \textbf{Control-period variation.}
    Following \citet{peng2018sim}, we set
    $\Delta t_t=\Delta t_0+\xi_t$, where
    $\xi_t\sim\operatorname{Exp}(\lambda)$; $\lambda$ is drawn per episode
    and $\xi_t$ per decision. One command is held for
    $\kappa_t=\operatorname{round}(\Delta t_t/h_{\mathrm{sim}})$ simulator
    substeps, changing its duration time.
\end{itemize}

\subsubsection{Operator details}
\begin{itemize}[leftmargin=1.4em,itemsep=2pt,topsep=3pt]
    \item \textbf{Stochastic.}
    Signal noise is redrawn per step, whereas \textbf{bias} persists for an
    episode and dynamics parameters are sampled at reset.
    \textbf{loguniform} samples in log space to balance reciprocal
    multipliers. Latency draws act on command buffers, timestamped
    observations, or action-hold duration.

    \item \textbf{Adversarial.}
    The observation operator searches for an input that changes the current
    actor's output within radius $\epsilon$; it requires a differentiable
    feed-forward actor. The three action operators respectively oppose the
    chosen command, rotate a 2-D command, or suppress its goal-directed
    component. Commands are clipped to the action space before perturbation.

    \item \textbf{Parametric.}
    These operators apply a specified value rather than draw one.
    \textbf{set} handles parameters whose nominal value is zero, while
    \textbf{translate} expresses a displacement in physical units.
    Semantic operators change appearance or task referents.

    \item \textbf{Non-stationary.}
    A \textbf{schedule} changes the shift magnitude over time, whereas latency operators
    change when signals enter the control loop. \textbf{push} creates an
    intermittent event. For feedback \textbf{delay}, interval release sums
    rewards over $k$ steps, while shifted release delivers each reward $k$
    steps later; both flush pending feedback at episode end.

    \item \textbf{Composition.}
    An ordered list specifies wrapper order explicitly. Each specification
    uses its own seeded random stream, making compound conditions reproducible.
\end{itemize}
\section{Evaluation protocol}
\label{app:evaluation_grid}

\begin{table}[tbp]
\centering
\footnotesize
\setlength{\tabcolsep}{2.5pt}
\renewcommand{\arraystretch}{1.15}

\begin{tabularx}{\textwidth}{
    @{}
    >{\raggedright\arraybackslash}p{0.235\textwidth}
    >{\raggedright\arraybackslash}p{0.14\textwidth}
    c
    >{\centering\arraybackslash}X
    >{\centering\arraybackslash}X
    >{\centering\arraybackslash}X
    >{\centering\arraybackslash}X
    @{}
}
\toprule
Environment & Shift factor & Unit &
$Q1$ & $Q2$ & $Q3$ & $Q4$  \\
\midrule

\multicolumn{7}{@{}l}{\textit{Online}} \\[1pt]

\multirow[t]{3}{0.235\textwidth}{%
    \raggedright
    \textbf{Hopper-v5}\\
}
& Gravity & $\times$ & 0.9, 1.1 & 0.8, 1.2 & 0.7, 1.3 & 0.6, 1.4 \\
& Morphology & $\times$ & 0.9, 1.1 & 0.8, 1.2 & 0.7, 1.3 & 0.6, 1.4 \\
& Actuator gear & $\times$ & 0.9, 1.1 & 0.8, 1.2 & 0.7, 1.3 & 0.6, 1.4 \\

\midrule

\multirow[t]{3}{0.235\textwidth}{%
    \raggedright
    \textbf{HalfCheetah-v5}\\
}
& Gravity & $\times$ & 0.9, 1.1 & 0.8, 1.2 & 0.7, 1.3 & 0.6, 1.4 \\
& Morphology & $\times$ & 0.9, 1.1 & 0.8, 1.2 & 0.7, 1.3 & 0.6, 1.4 \\
& Actuator gear & $\times$ & 0.9, 1.1 & 0.8, 1.2 & 0.7, 1.3 & 0.6, 1.4 \\

\midrule

\multirow[t]{3}{0.235\textwidth}{%
    \raggedright
    \textbf{Walker2d-v5}
}
& Gravity & $\times$ & 0.9, 1.1 & 0.8, 1.2 & 0.7, 1.3 & 0.6, 1.4 \\
& Morphology & $\times$ & 0.9, 1.1 & 0.8, 1.2 & 0.7, 1.3 & 0.6, 1.4 \\
& Actuator gear & $\times$ & 0.9, 1.1 & 0.8, 1.2 & 0.7, 1.3 & 0.6, 1.4 \\

\midrule

\textbf{FetchReachDense-v3}
& Morphology & $\times$ & 0.8 & 1.2 & 1.4 & 1.6 \\
& Actuator gear & $\times$ & 0.4 & 0.2 & 0.1 & 2 \\

\addlinespace[2pt]
\midrule

\textbf{PointMaze\_UMaze-v3}
& Actuator gear & $\times$ & 0.8 & 1.2 & 1.4 & 0.6 \\
& Dry friction & N & 10 & 20 & 30 & 40 \\
& Adversarial action & -- & oppose\,(0.6) & oppose-goal\,(0.6) & oppose-goal\,(0.8) & rotate\,(70) \\

\addlinespace[2pt]
\midrule

\textbf{CarRacing-v2}
& Morphology & $\times$ & 0.8, 1.2 & 0.6 & 1.4 & 1.6, 1.8 \\
& Actuator gear & $\times$ & 0.8, 1.2 & 0.6 & 1.4 & 0.4, 1.6 \\
& Surface friction & $\times$ & 0.9, 1.1 & 1.2 & 0.8 & 0.6, 0.7 \\

\midrule
\midrule

\multicolumn{7}{@{}l}{\textit{Offline}} \\[5pt]

\multirow[t]{3}{0.235\textwidth}{%
    \raggedright
    \textbf{Hopper-v5}\\
}
& Gravity & $\times$ & 0.9, 1.1 & 0.8, 1.2 & 0.7, 1.3 & 0.6, 1.4 \\
& Morphology & $\times$ & 0.9, 1.1 & 0.8, 1.2 & 0.7, 1.3 & 0.6, 1.4 \\
& Actuator gear & $\times$ & 0.9, 1.1 & 0.8, 1.2 & 0.7, 1.3 & 0.6, 1.4 \\

\midrule

\multirow[t]{3}{0.235\textwidth}{%
    \raggedright
    \textbf{HalfCheetah-v5}\\
}
& Gravity & $\times$ & 0.9, 1.1 & 0.8, 1.2 & 0.7, 1.3 & 0.6, 1.4 \\
& Morphology & $\times$ & 0.9, 1.1 & 0.8, 1.2 & 0.7, 1.3 & 0.6, 1.4 \\
& Actuator gear & $\times$ & 0.9, 1.1 & 0.8, 1.2 & 0.7, 1.3 & 0.6, 1.4 \\

\midrule

\multirow[t]{3}{0.235\textwidth}{%
    \raggedright
    \textbf{Walker2d-v5}
}
& Gravity & $\times$ & 0.9, 1.1 & 0.8, 1.2 & 0.7, 1.3 & 0.6, 1.4 \\
& Morphology & $\times$ & 0.9, 1.1 & 0.8, 1.2 & 0.7, 1.3 & 0.6, 1.4 \\
& Actuator gear & $\times$ & 0.9, 1.1 & 0.8, 1.2 & 0.7, 1.3 & 0.6, 1.4 \\

\midrule

\multirow[t]{2}{0.235\textwidth}{%
    \raggedright\textbf{AdroitHandDoor-v1}
}
& Actuator gear & $\times$ & 0.8 & 1.2 & 1.4 & 0.6 \\
& Latch friction & $\times$ & 2 & 3 & 4 & 5 \\

\midrule

\multirow[t]{3}{0.235\textwidth}{%
    \raggedright\textbf{PointMaze\_Open-v3}
}
& Actuator gear & $\times$ & 0.4 & 0.2 & 0.1 & 0.05 \\
& Dry friction & N & 30 & 45 & 60 & 75 \\
& Adversarial action & -- & oppose-goal & rotate & rotate\,(90) & oppose \\

\midrule

\multirow[t]{3}{0.235\textwidth}{%
    \raggedright
    \textbf{LunarLander}\\
    \textbf{Continuous-v3}
}
& Gravity & $\times$ & 0.9, 1.1 & 1.15 & 0.85 & 0.75, 1.25 \\
& Engine power & $\times$ & 0.9, 1.1 & 0.75 & 1.25 & 0.6, 1.4 \\
& Wind & N & 5, 10 & 15 & 20 & 25, 30 \\

\bottomrule
\end{tabularx}

\caption{\textbf{The 220 perturbation configurations and the shift levels.} Each row
is one shift factor on one environment; the four cells list every
perturbation configuration evaluated on that factor, grouped by shift levels. $Q$ indexes displacement from nominal. Adversarial-action is discrete attack modes with no scalar
distance from nominal.}
\label{tab:evaluation_grid}
\end{table}

\begin{table}[tbp]
    \centering
    \footnotesize
    \setlength{\tabcolsep}{3pt}
    \renewcommand{\arraystretch}{1.15}
    \begin{tabularx}{\textwidth}{@{}lll@{\hspace{4pt}}>{\raggedright\arraybackslash}X@{}}
                \toprule
        Shift source & Perturbation & Mode & Setting \\
        \midrule
        \addlinespace[1pt]
        Observation & \texttt{obs\_rel\_0.10}     & stochastic & Gaussian at $0.10\,\sigma_i$, $\sigma_i$ the dimension's own std \\
         & \texttt{obs\_uni\_0.10}     & stochastic & $\mathcal{U}[-0.10,\,0.10]$ \\
         & \texttt{obs\_gauss\_0.10}   & stochastic & $\sigma=0.10$ \\
         & \texttt{mad\_0.02}          & adversarial & MAD gradient attack, $\epsilon=0.02$ \\
         & \texttt{mad\_0.05}          & adversarial & MAD gradient attack, $\epsilon=0.05$ \\
        \addlinespace[2pt]
        Action & \texttt{act\_uni\_0.10}     & stochastic & $\mathcal{U}[-0.10,\,0.10]$ on the clipped action \\
         & \texttt{act\_gauss\_0.10}   & stochastic & $\sigma=0.10$ on the clipped action \\
         & \texttt{act\_offset\_0.05}  & parametric & $+0.05$ on every actuator \\
         & \texttt{act\_oppose\_0.10}  & adversarial & $\epsilon=0.10$ against the action just chosen \\
         & \texttt{cmd\_oppose\_0.20}  & adversarial & $\epsilon=0.20$, applied to the command before clipping \\
        \addlinespace[2pt]
        Dynamics & \texttt{gear\_static\_0.8}  & parametric & actuator gear $\times\,0.8$ for the whole episode \\
         & \texttt{gear\_ramp\_0.8}    & non-stationary & gear $1.0\!\rightarrow\!0.8$, linear over $1000$ steps \\
         & \texttt{gear\_sine\_p500}   & non-stationary & gear cycling in $[0.8,\,1.2]$, period $500$ steps \\
         & \texttt{push\_60}           & non-stationary & $60$\,N on the torso, $12$ steps every $250$, sign random \\
         & \texttt{trans\_t1}          & stochastic & state noise $\sigma_q=10^{-3}$, $\sigma_v=10^{-2}$ per step \\
        \addlinespace[2pt]
        Latency & \texttt{obs\_delay\_4ms}    & stochastic & delay drawn from $[2,\,4]$\,ms, interpolated \\
         & \texttt{obs\_delay\_8ms}    & stochastic & delay drawn from $[4,\,8]$\,ms \\
         & \texttt{obs\_delay\_16ms}   & stochastic & delay drawn from $[8,\,16]$\,ms \\
         & \texttt{act\_delay\_8ms}    & parametric & fixed buffer of $1$ control step ($8$\,ms) \\
         & \texttt{act\_delay\_rand1}  & stochastic & buffer redrawn each episode, $\{0,1\}$ steps \\
        \multicolumn{4}{@{}l}{\emph{In-training perturbation}} \\
        \addlinespace[1pt]
        Reward & Gaussian, uniform, constant & signal corruption & three arms matched at $\sigma=0.5$ (uniform $\pm0.866$) \\
        Latency(Reward) & credit ladder & interval reward release & $k\in\{4,16,32,64\}$ control steps \\
        \bottomrule

    \end{tabularx}
    \caption{\textbf{44 individual shift perturbation configuration.} The 22 fixed sets on Hopper-v5 are instantiated  separately for online and
offline evaluation, composing 44 perturbation configurations used to
compute the scores in Section~\ref{sec:channels}, and the \emph{In-training perturbation} means the shift acts during learning, the evaluation is nominal. A control step is $8$\,ms
    (\texttt{frame\_skip} $4\times2$\,ms).}
    \label{tab:part2_setup}
\end{table}

\begin{table}[tbp]
    \centering
    \small
    \setlength{\tabcolsep}{4pt}
    \renewcommand{\arraystretch}{1.18}
    \begin{tabular}{@{}lll@{\hspace{9pt}}l@{}}
        \toprule
        Shift source & Shift factor & Resampling rule & Setting \\
        \midrule
        \multicolumn{4}{@{}l}{\emph{Hopper-v5} $\leftarrow$ OmniH2O \citep{he2024omnih2o}} \\
        \addlinespace[1pt]
        Dynamics & Foot/link friction & log-uniform, per episode & $\times[0.38,1.06]$ \\
        Dynamics & Link mass          & log-uniform, per episode & $\times[0.81,1.17]$ \\
        Dynamics & Actuator gear      & log-uniform, per episode & $\times[0.84,1.14]$ \\
        Dynamics & Torso push         & onset phase per episode, sign per push & $119$\,N, every $625$ steps, $10$ steps \\
        Action   & Torque noise       & Gaussian, per step, pre-clip & $\sigma=0.06$ \\
        Latency   & Action delay       & uniform integer, per episode & $[1,2]$ steps \\
        \addlinespace[3pt]
        \midrule
        \multicolumn{4}{@{}l}{\emph{Pusher-v5s} $\leftarrow$ \citet{peng2018sim} Table~I} \\
        \addlinespace[1pt]
        Dynamics    & Link mass         & log-uniform, per episode & $\times[0.03,32]$ \\
        Dynamics    & Object mass       & log-uniform, per episode & $\times[0.11,3.8]$ \\
        Dynamics    & Joint damping     & log-uniform, per episode & $\times[0.02,1.8{\times}10^{3}]$ \\
        Dynamics    & Actuator gear     & log-uniform, per episode & $\times[0.18,5.7]$ \\
        Latency      & Control period    & $\lambda$ per episode, $\mathrm{d}t$ per step & $\mathrm{d}t\!=\!\mathrm{d}t_0\!+\!\mathrm{Exp}(1/\lambda)$, $\lambda\!\in\![50,400]$ \\
        Observation & Sensor noise      & Gaussian, per step, per-feature std & $0.125\,\sigma_i$ \\
        Semantics
                    & Goal displacement & deterministic & $0.25$\,m diagonal \\
        \bottomrule
    \end{tabular}
    \caption{\textbf{Reference-derived compound shifts scenario configuration.} Every factor of a profile is applied
    jointly in one mode; this sheet lists the reference system's own factor settings.}
    \label{tab:reality_setup}
\end{table}

\subsection{Metrics}
\label{app:metric}
\paragraph{Normalized score.}
\begin{table}[tbp]
\centering
\small
\setlength{\tabcolsep}{5pt}
\renewcommand{\arraystretch}{1.1}
\begin{tabular}{@{}l l r r l@{}}
\toprule
Task & Metric & $R_{\min}$ & $R_{\max}$ & Source \\
\midrule
Hopper-v5 & episodic return & $-20.27$ & $3234.3$ & D4RL v2 random / expert \\
HalfCheetah-v5 & episodic return & $-280.18$ & $12135.0$ & D4RL v2 random / expert \\
Walker2d-v5 & episodic return & $1.63$ & $4592.3$ & D4RL v2 random / expert \\
\addlinespace[2pt]
AdroitHandDoor-v1 & episodic return & $-56.51$ & $2880.57$ & D4RL v2 random / expert \\
PointMaze\_Open-v3 & goals per 300 steps & $0.0$ & $3.28$ & no goal/ IQL nominal \\
LunarLanderContinuous-v3 & episodic return & $-213.9$ & $136.9$ & random policy / IQL nominal \\
\addlinespace[2pt]
PointMaze\_UMaze-v3 & goals per 300 steps & $0.23$ & $5.5$ & random policy / kinematic ceiling \\
FetchReachDense-v3 & neg.\ goal distance & $-8.42$ & $0.0$ & random policy / reward maximum \\
CarRacing-v2  & episodic return & $-30.44$ & $900.0$ & random policy / solved threshold \\
Pusher-v5s  & episodic return & $-146.2$ & $-23.97$ & random policy / PPO nominal \\
Unitree G1 & episodic return & 0 & 37.30 & zero floor / PPO nominal \\

\bottomrule
\end{tabular}
\caption{\textbf{Normalization anchors $(R_{\min},R_{\max})$.} The pairs
entering ~\eqref{eq:normalization}, fixed per environment and shared by
every algorithm, dataset quality, and perturbation configurations. \emph{Source} names
the construction behind each pair: D4RL~v2 where a published pair exists, and
otherwise a random-policy floor with either a task-defined or a
reference-learner ceiling. }
\label{tab:normalization_anchors}
\end{table}

To compare standard-control tasks with different native metrics, let $\bar R$
denote the mean raw task score over evaluation episodes. We report the
unclipped normalized score
\begin{equation}
\mathrm{S}(\bar R)
=100\frac{\bar R-R_{\min}}{R_{\max}-R_{\min}},
\label{eq:normalization}
\end{equation}
where $R_{\min}$ and $R_{\max}$ are fixed task-level reference scores shared by
all methods, seeds, and shift conditions. We use published D4RL references
where available and otherwise use task-defined or reference-policy anchors;
all values are listed in Table~\ref{tab:normalization_anchors}. Normalization
precedes cross-task aggregation. For the standard-control experiments, we map each task's native metric to a
common comparable scale. 

\paragraph{Performance and safety.}
We report performance and safety separately on the advanced robotic tasks.
For G1, performance is PPO-reference-normalized return over five
out-of-envelope conditions (Table~\ref{tab:normalization_anchors}), while safety is the fraction of episodes
surviving the 20\,s horizon under 20--80\,ms action delays. For Drawer,
performance is the fraction of episodes opening the drawer beyond
$0.30$\,m over fourteen out-of-envelope conditions; safety is the fraction
whose cumulative joint-limit cost remains below the per-episode budget of
$20$. We average episodes within each training seed, then report the mean
and standard deviation across seeds.

\subsection{Evaluation grids}
\label{app:eval}
The main robust RL algorithms evaluation grid varies one shift factor at a time across online and
offline tasks, grouping its non-nominal settings into four task--factor-specific
shift levels (Table~\ref{tab:evaluation_grid}). The grid is fixed and same for every algorithm: $220$ non-nominal conditions, $110$ online and
$110$ offline, plus the nominal setting of every factor. The individual shift
grid instead varies intervention modes within a shift source, allowing
method-level specialization to be diagnosed
(Table~\ref{tab:part2_setup}). Finally, the compound grid activates physical,
control, sensing, and timing factors jointly in reference-derived Hopper and
Pusher tasks (Table~\ref{tab:reality_setup}). Thus, the three grids measure
robustness breadth, shift-specific behavior, and robustness under
interacting deployment factors, respectively.

The advanced tasks bring their own grids, and both place every method on the same conditions. On Isaac Lab the conditions are positioned relative to the reference envelope~\citep{he2024omnih2o} of Tables~\ref{tab:envelope_g1} and~\ref{tab:envelope_drawer}: G1 has $31$ per seed, six inside that envelope, eighteen on its boundary, five beyond it, and flat and rough nominal. Drawer has $16$, two controls and fourteen beyond the reference ranges, each moving one axis while the rest keep their training randomization.

\begin{figure}[H]
    \centering
    \includegraphics[width=0.8\linewidth]{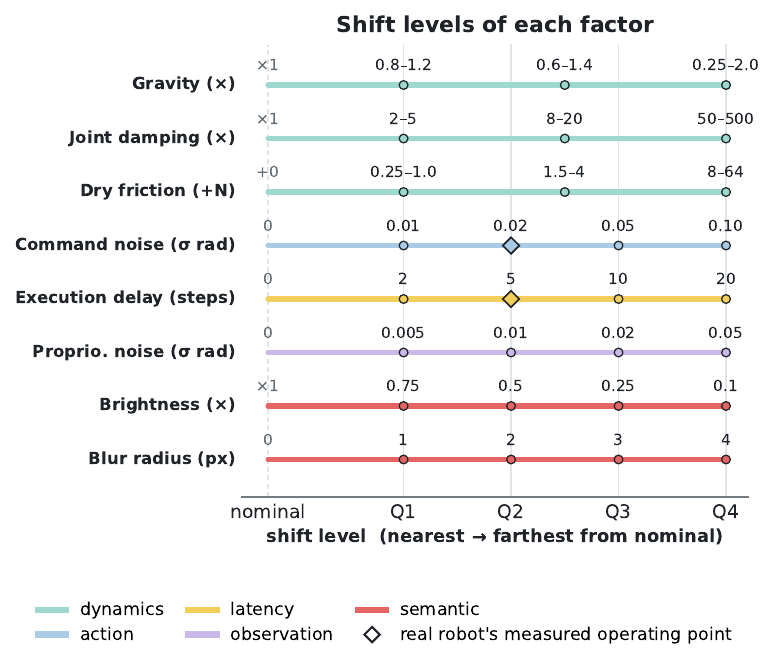}
\caption{\textbf{VLA evaluation grid.} Each row is one shift factor, coloured
by its shift source, with the nominal value on the left and the setting of each
level above its circle. Q1--Q4 order the levels from nearest to farthest from
nominal; the three dynamics shift have three levels. The diamond marks the
operating point measured on the real robot: it sits at Q2, so Q1 is milder
than the hardware and Q3--Q4 go beyond it.}

    \label{fig:vla_grid}
\end{figure}

The VLA evaluation tests eight shift factors across five shift sources
(Figure~\ref{fig:vla_grid}). Each factor has its own Q1--Q4 ladder, ordered
outward from its nominal setting. Hardware logs place execution delay ($83$\,ms) and command
noise ($0.020$\,rad) at Q2, so the sweep includes both milder and more
severe conditions than those measured on the platform. We report the
actual input change for perception and command shifts separately; the
three dynamics shifts are specified in physical units.

\begin{table}[tbp]
\centering
\footnotesize
\setlength{\tabcolsep}{6pt}
\renewcommand{\arraystretch}{1.08}
\begin{tabular}{@{}l l l@{}}
\toprule
Shift factor & Training setting & Evaluation setting \\
\midrule
Contact friction & $\mathcal{U}[0.2, 1.1]$ & $0.1$ \\
Link mass & $\times\,\mathcal{U}[0.7, 1.3]$ & $\times\,1.5$ \\
Base CoM, $x$ and $y$ & $\pm 0.1$\,m & $+0.15$\,m along $x$ \\
Torque noise & $\pm 10\%$ of the effort limit & $20\%$ \\
Action delay & $1$--$3$ policy steps ($20$--$60$\,ms) & $4$ steps ($80$\,ms) \\
Actuator $K_p$, $K_d$ & $\times\,\mathcal{U}[0.75, 1.25]$ & boundary only, $0.75$ and $1.25$ \\
Velocity push & $1$\,m/s, random direction, every $5$\,s & boundary only, $1$\,m/s \\
\bottomrule
\end{tabular}
\caption{\textbf{G1: training and evaluation configuration.}}
\label{tab:envelope_g1}
\end{table}

\begin{table}[tbp]
\centering
\footnotesize
\setlength{\tabcolsep}{6pt}
\renewcommand{\arraystretch}{1.08}
\begin{tabular}{@{}l l l@{}}
\toprule
Shift factor & Training setting& Evaluation setting\\
\midrule
Robot friction & static and dynamic $\mathcal{U}[0.8, 1.25]$ & both $0.4$; both $1.8$ \\
Handle friction & static $\mathcal{U}[1.0, 1.25]$, dynamic $\mathcal{U}[1.25, 1.5]$ & both $0.5$; both $2.0$ \\
Cabinet mass & $\times\,\mathcal{U}[0.8, 1.2]$ & $\times\,0.6$; $\times\,1.4$ \\
Arm $K_p$ & $\times\,\mathcal{U}[0.9, 1.1]$ & $\times\,0.75$; $\times\,1.25$ \\
Arm $K_d$ & $\times\,\mathcal{U}[0.9, 1.1]$ & $\times\,0.75$; $\times\,1.25$ \\
Drawer damping & $\times\,\mathcal{U}[0.5, 1.5]$ & $\times\,0.25$; $\times\,2.0$ \\
Action delay & $0$--$2$ policy steps & $3$ steps ($50$\,ms); $4$ steps ($66.7$\,ms) \\
\bottomrule
\end{tabular}
\caption{\textbf{Drawer: training and evaluation configuration.}}
\label{tab:envelope_drawer}
\end{table}

\section{Benchmark tasks and datasets}
\label{app:tasks}

In this section we present an overview of the environments the benchmark
evaluates on, the control interface each of them exposes, and the offline datasets learn from. 
The main experiment in Section~\ref{sec:exp_part1} covers locomotion, manipulation, navigation, and
vehicle control; Pusher and DoorCausal provide the additional semantic-channel tests and compound-profile in Section~\ref{sec:channels}. Table~\ref{tab:task_inventory} gives the
default interface of every task we included in our benchmark.

Offline methods learn from the fixed datasets in Table~\ref{tab:offline_datasets}.
We use Minari~\citep{minari}, including its D4RL ports~\citep{fu2020d4rl},
wherever a corresponding dataset is available; only for LunarLander do we
collect our own dataset.

\begin{table*}[!htbp]
\centering
\scriptsize
\setlength{\tabcolsep}{4pt}
\renewcommand{\arraystretch}{1.12}
\begin{tabularx}{\textwidth}{@{}
    >{\raggedright\arraybackslash}p{0.24\textwidth}
    >{\raggedright\arraybackslash}X
    >{\raggedright\arraybackslash}p{0.13\textwidth}
    c
    >{\raggedright\arraybackslash}p{0.15\textwidth}@{}}
\toprule
Task & Observation & Action (rate) & Horizon & Task type \\
\midrule
\multicolumn{5}{@{}l}{\textbf{Standard control}} \\
\textit{Hopper-v5}
    & 11-D state & 3-D & 1,000 & Locomotion \\
\textit{Walker2d-v5}
    & 17-D state & 6-D & 1,000 & Locomotion \\
\textit{HalfCheetah-v5}
    & 17-D state & 6-D & 1,000 & Locomotion \\
\textit{AdroitHandDoor-v1}
    & 39-D state & 28-D & 200 & Manipulation \\
\textit{FetchReachDense-v3}
    & 13-D proprioception and relative goal
    & 4-D & 50 & Manipulation \\
\textit{Pusher-v5s}
    & 23-D state & 7-D & 100 & Manipulation \\
\textit{DoorCausal}
    & 54-D State & 8-D & 300 & Manipulation \\
\textit{PointMaze\_UMaze-v3}
    & 6-D state and desired goal
    & 2-D & 300 & Navigation \\
\textit{PointMaze\_Open-v3}
    & 6-D state and desired goal
    & 2-D & 300 & Navigation \\
\textit{CarRacing-v2}
    & $96 \times 96 \times 3$ pixels
    & 3-D & 1,000 & Driving \\
\textit{LunarLanderContinuous-v3}
    & 8-D state & 2-D & 1,000 & Driving \\
\midrule
\multicolumn{5}{@{}l}{\textbf{Advanced robotics}} \\
\textit{Unitree G1}
    & 123-D state & 37-D (50\,Hz) & 1,000 & Locomotion \\
\textit{Franka Drawer}
    & 31-D state & 8-D (60\,Hz) & 480 & Manipulation \\
\textit{insert-mouse-battery}
    & 14-D joint state and three $180\!\times\!320\!\times\!3$ camera streams
    & 14-D (60\,Hz) & Demo & Bimanual manipulation \\
\bottomrule
\end{tabularx}
\caption{\textbf{Benchmark tasks.}
Observation and Action specify the default policy interface; control rates
are shown in parentheses where reported. Numeric horizons are maximum
environment steps per episode, while \emph{Demo} follows the length of the
recorded demonstration used for VLA evaluation.}

\label{tab:task_inventory}
\label{tab:advanced_task_inventory}
\end{table*}

\begin{table*}[!htbp]
\centering
\footnotesize
\setlength{\tabcolsep}{4pt}
\renewcommand{\arraystretch}{1.12}
\begin{tabularx}{\textwidth}{@{}
>{\raggedright\arraybackslash}p{0.23\textwidth}
>{\raggedright\arraybackslash}p{0.36\textwidth}
>{\raggedright\arraybackslash}X@{}}
\toprule
Task & Dataset & Source or collection \\
\midrule
Hopper, Walker2d, HalfCheetah
& Minari \texttt{medium-v0}
& Collected in Gymnasium~v5 environments \\
AdroitHandDoor
& \path{D4RL/door/expert-v2}
& Minari port of D4RL trajectories \\
PointMaze\_Open
& \path{D4RL/pointmaze/open-v3}
& Minari port of D4RL trajectories \\
LunarLanderContinuous
& Collected for this benchmark
& Deterministic SAC checkpoints trained for $1.5$ million steps;
  selected medium corpus of $1{,}741$ episodes \\
\bottomrule
\end{tabularx}
\caption{\textbf{Offline datasets.} Each dataset contains approximately
one million transitions.}
\label{tab:offline_datasets}
\end{table*}

\begin{figure}[!ht]
    \centering
    \includegraphics[width=0.5\linewidth]{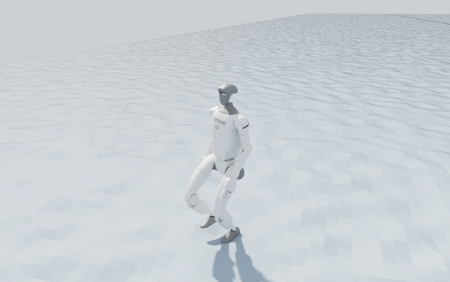}\hfill
    \includegraphics[width=0.5\linewidth]{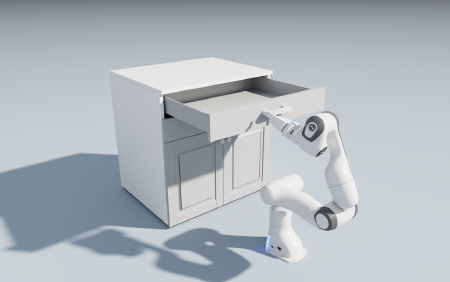}
    \caption{\textbf{Isaac Lab tasks.} \emph{Left}: Humanoid G1 locomotion task; \emph{right}: Franka drawer manipulation  task.}
    \label{fig:isaac_lab_tasks}
\end{figure}

\begin{figure}[!ht]
    \centering
    \includegraphics[width=1\linewidth]{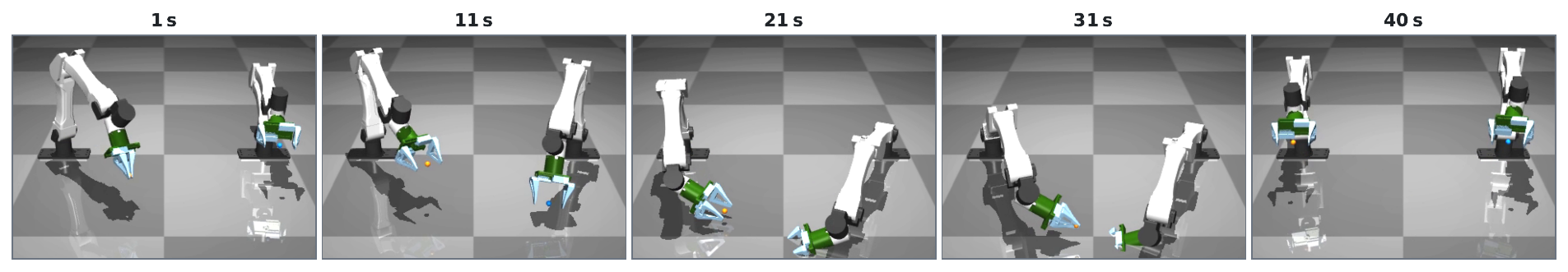}
    \caption{\textbf{VLA task.} Bimanual \emph{insert-mouse-battery} demonstrations of the
    RSS 2026 post-training challenge~\citep{posttraining_robotics_2026}.}
    \label{fig:vla_task}
\end{figure}

\paragraph{Advanced robotic tasks.}
These tasks test robust safe algorithms and show our interface extensibility meanwhile. In Isaac Lab~\citep{mittal2025isaaclab}, Unitree G1 (Figure~\ref{fig:isaac_lab_tasks})
locomotion measures return and survival under delay, and Franka drawer
manipulation measures success and joint-limit cost under physical shifts. The
VLA task (Figure~\ref{fig:vla_task}) fine-tunes $\pi_{0.5}$~\citep{intelligence2025pi05visionlanguageactionmodelopenworld} on expert
demonstrations for the bimanual \emph{insert-mouse-battery} task from the RSS
2026 post-training challenge~\citep{posttraining_robotics_2026}, then evaluates
perception and execution shifts through camera and robot-state interfaces.

\section{Algorithms}
\label{app:algorithms}

The library holds six standard RL baselines without any robust design,
thirteen robust RL algorithms, and two robust safe algorithms.
They are organized by the taxonomy of Section~\ref{sec:library_objective}.

\subsection{Offline algorithms}
\label{app:algorithms_offline}

\subsubsection{Standard RL baselines.}

\paragraph{IQL}
IQL~\citep{kostrikov2021offline} avoids querying out-of-distribution actions
altogether: a value function is fitted to the critic by expectile regression, the twin critics bootstrap from that value function, and the
actor is extracted by advantage-weighted regression.

\paragraph{TD3+BC}
TD3+BC~\citep{fujimoto2021minimalist} adds a behavior-cloning term to the TD3 actor loss and normalizes states. The term is balanced against the critic by dividing by the mean absolute $Q$ over the batch.

\paragraph{MOPO}
MOPO~\citep{yu2020mopo} learns an ensemble dynamics model, branches short
rollouts from dataset states, and penalizes the rollout reward by the model's
predicted uncertainty before running SAC on the mixture of real and model
data.

\paragraph{SynthER}
SynthER~\citep{lu2023synthetic} fits a diffusion model to logged transitions and trains an unmodified offline learner on a much larger sample from it. We count it as data augmentation rather than a generative method: it samples individual transitions to enlarge the dataset, where FWM and PLR-PVL sample multi-step futures conditioned on the current state and turn them into critic targets.

\subsubsection{Learner-centric algorithms.}

\paragraph{RFQI}
RFQI~\citep{panaganti2022robust} replaces the fitted-$Q$ target by a
worst-case Bellman backup over a total-variation uncertainty set of radius
$\rho$ around the empirical transitions; the inner maximization is solved
through a dual variable fitted per batch, and the actor and critic follow the
release's BCQ-style architecture. 

\paragraph{RORL}
RORL~\citep{yang2022rorl} makes an SAC learner conservative through
smoothing: the critic and policy are regularized to vary little within an
$\epsilon$-ball of dataset states, and $Q$-values at perturbed states are
penalized by the disagreement of a ten-critic ensemble. 

\paragraph{ATLA-IQL}
The offline ATLA adaptation~\citep{zhang2021robust} keeps ATLA's principle,
a learned observation adversary alternated with the learner, but the adversary
is trained by a differentiable objective on the critic  by minimizing the
minimum $Q$ at the policy's action under the perturbed state since no environment is available. 

\subsubsection{Data-centric algorithms.}

\paragraph{RSC-IQL}
RSC-IQL~\citep{ding2023seeing} applies RSC's counterfactual rewriting to logged
data where a causal mask over state dimensions is fitted from the dataset, and
transitions are rewritten by permuting or perturbing the dimensions the mask
marks as non-causal.

\paragraph{RAMBO}
RAMBO~\citep{rigter2022rambo} augments the dataset with short rollouts from
a learned dynamics ensemble and replaces the uncertainty penalty
by adversarial training of the model: the ensemble is updated to lower the
critic's value of the policy's transitions while a maximum-likelihood term keeps
it close to the data.

\subsubsection{Generative model.}

\paragraph{ROMB}
ROMB~\citep{chen2025policy} adapts a learned world model against the policy
under a constrained maximin objective: the model is pushed toward transitions
that lower the policy's value while a supervised loss keeps it on the data, and
the policy is trained on rollouts from the adapted model.

\paragraph{FWM}

FWM transfers the multi-step future modeling of Diffusion World
Model~\citep{ding2024diffusion} to a flow-matching~\citep{lipman2022flow}
world model: instead of a one-step bootstrap, the critic regresses an $h$-step
return whose future is sampled from the generative model conditioned on the
current state, action and return-to-go. Writing $x_1$ for a length-$T$ window of states and rewards
and $x_0 \sim \mathcal{N}(0, I)$, the diffusion model's denoising objective
becomes a regression onto the straight-line velocity of the linear path
$x_t = (1-t)\,x_0 + t\,x_1$,
\begin{equation}
    \mathcal{L}
    = \mathbb{E}_{t \sim \mathcal{U}[0,1],\, x_0,\, (x_1, c, g)}
      \big\lVert v_\theta(x_t, t, c, g) - (x_1 - x_0) \big\rVert_2^2 ,
    \label{eq:fwm_flow}
\end{equation}
with $c$ the current state and action and $g$ the return-to-go, dropped with
probability $0.2$ so one network carries both the conditional and the
unconditional velocity. Sampling then replaces the reverse diffusion chain by an
ODE integrated from $t=0$ to $t=1$ in $N=5$ midpoint steps on the
guidance-mixed velocity:
\begin{equation}
    \tilde{v}_\theta = w\, v_\theta(\cdot \mid c, g) + (1 - w)\, v_\theta(\cdot \mid \varnothing),
    \qquad w = 2.0 .
    \label{eq:fwm_guidance}
\end{equation}

\paragraph{PLR-PVL}
PLR-PVL~\citep{berdica2024robust} trains a collection of world models consistent with the
offline data, treats each as a level in the sense of unsupervised environment
design, and prioritizes them during policy learning by the value loss they induce.
We adapt the method by replacing their world model with the flow-matching model
of ~\eqref{eq:fwm_flow}: the ensemble is $K$ flow models differing only
in training seed, each proposing one $h$-step future for the same conditioning,
and the critic target is taken per sample from the model whose rollout carries
the largest positive value loss,
\begin{equation}
  k^\star = \arg\max_k \; \frac{1}{h}\sum_{t=1}^{h}
            \max\!\big(\delta^{(k)}_t,\, 0\big),
  \label{eq:plr_pvl}
\end{equation}
where $\delta^{(k)}_t$ is the GAE-$\lambda$ residual of the current critic along
the $k$-th rollout.

\subsection{Online algorithms}
\label{app:algorithms_online}

\subsubsection{Standard RL baselines.}

\paragraph{SAC}
SAC~\citep{haarnoja2018soft} is the off-policy reference: twin critics with a
clipped double-$Q$ target and a stochastic actor whose objective adds an
entropy bonus with automatically tuned temperature. 

\paragraph{PPO}
PPO~\citep{schulman2017proximal} is the on-policy reference: a clipped
probability-ratio surrogate on trajectories from the current policy, with
generalized advantage estimation. 

\paragraph{TRPO}
TRPO~\citep{schulman2015trust} replaces PPO's clipped surrogate by an explicit
KL trust region solved with conjugate gradients. It appears only as the base
of RARL-TRPO.

\subsubsection{Learner-centric algorithms}

\paragraph{ATLA}
ATLA~\citep{zhang2021robust} trains a policy against an \emph{optimal}
observation adversary in the state-adversarial MDP~\citep{zhang2020robust}:
the adversary is itself an RL agent that outputs a bounded perturbation of the
observation, is rewarded by the protagonist's negative return, and is updated
in alternation with the protagonist. 

\paragraph{ATLA-SA}
ATLA-SA~\citep{zhang2021robust} adds the SA-PPO
regularizer~\citep{zhang2020robust} to ATLA: a KL penalty between the policy
at a state and at the worst-case neighbor found by stochastic gradient
Langevin dynamics, and a recurrent (LSTM) policy with a 100-step history so
that the agent can infer a persistent perturbation from context.

\paragraph{RSC}
RSC~\citep{ding2023seeing} learns a causal graph and dynamics model over the
state variables from replayed experience, then generates counterfactual
transitions by intervening on the variables the graph marks as non-causal for
the reward. Mixing these into replay removes the spurious correlations a
policy would otherwise exploit. 

\subsubsection{Environment-centric algorithms.}

\paragraph{RARL}
RARL~\citep{pinto2017robust} casts robustness as a two-player zero-sum game:
an adversary applies bounded external forces to the robot and is trained to
minimize the protagonist's return, and the two are updated in alternating
blocks. We implement RARL-PPO and RARL-TRPO in our library.
\paragraph{DR}
Domain randomization~\citep{tobin2017domain,peng2018sim} samples simulator
parameters from fixed ranges at every episode so that the policy meets a
distribution of dynamics during training. We implement DR-PPO and DR-SAC in our library.

\subsubsection{Robust safe methods.}

\paragraph{RAMU}
RAMU~\citep{queeney2023risk} applies a risk measure over
sampled next-state perturbations to both the reward and the cost targets; the
original is defined for TD learners, and the row reported here is a PPO
adaptation that applies the same transform to PPO's value targets.

\paragraph{SPiDR}
SPiDR~\citep{as2026spidr} trains a constrained policy under domain
randomization and adds a pessimistic cost penalty proportional to the
disagreement of next-state predictions across sampled dynamics; it runs with
the authors' solver and a joint-limit cost budget.

\section{Training  Settings}
\label{app:online_training_protocol}

\subsection{Training protocol}
\label{app:train_protocol}
We train each method with five independent seeds to its budget in
Table~\ref{tab:algorithm_budget}. Unless otherwise noted, we freeze the
final checkpoint and evaluate each policy for 20 episodes per condition
using the same evaluation seed. RFQI retains its native best-checkpoint
selection rule.

Our default protocol trains on the nominal task and introduces shifts only
when evaluating the frozen policy. Reward shift and  latency
delivery are exceptions: they affect learning feedback but cannot alter
the actions of a frozen policy. We therefore apply these shifts during
training and evaluate the resulting policies on the nominal task. 

The Isaac Lab and VLA studies are reported under their own protocol, and the
departures are the following. Isaac Lab tasks use three training seeds and score $128$ episodes per condition. They match the retained primary training
interactions within each task to the budget in Table~\ref{tab:isaaclab_budget}. The five methods share one base learner, so the comparison is between
mechanisms rather than between releases. And they do not hold the shift grid out
from training: PPO and RAMU-PPO train on the nominal specification, whereas
DR-PPO, ATLA-PPO and SPiDR-PPO each train inside a randomization envelope. Table~\ref{tab:envelope_g1} and Table~\ref{tab:envelope_drawer} show the envelopes applied in our benchmark, which is transferred from a humanoid sim-to-real pipeline~\citep{he2024omnih2o}. The VLA task finetunes a single $\pi_{0.5}$ model  on demonstrations,
and evaluated by joint-space tracking against the recorded demonstration.

\subsection{Training budget alignment}
\label{sec:alignment}

\begin{figure}[H]
    \centering
    \includegraphics[width=1\linewidth]{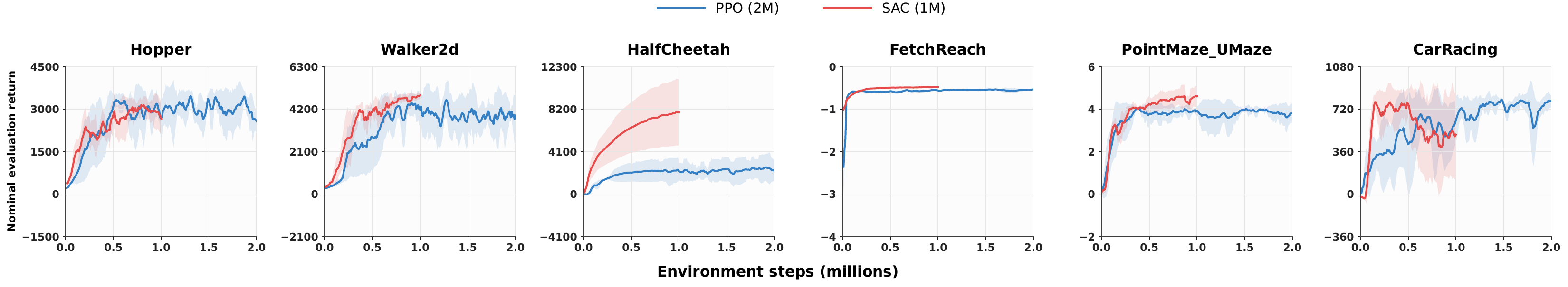}
    \caption{\textbf{Online training curves.}}
    \label{fig:online-training-curves}
\end{figure}

This part presents our training budget and provides the evidence of our training protocol alignment.
We train SAC-based algorithms for 1M environment steps and PPO-based algorithms
for 2M steps, consistent with commonly used training budgets~\citep{raffin2020generalized}.
This asymmetric budget accounts for PPO's lower sample efficiency: as an
on-policy method, PPO generally requires more environment interactions to reach
a comparable level of training maturity. Figure~\ref{fig:online-training-curves}
shows that the additional PPO steps provide meaningful learning progress.

However, more environment steps do not imply a larger computational budget.
On Walker2d, PPO at 2M steps requires only $0.80 \pm 0.29$ hours, compared with
$1.60 \pm 0.03$ hours for SAC at 1M steps on H100 GPUs
(Table~\ref{tab:walker-walltime}). Thus, the 2M-step PPO budget compensates for
its lower sample efficiency without giving it an advantage in wall-clock
compute. 

For ATLA and ATLA-SA, their official configuration remains unchanged, which is about 5M protagonist steps rather than the 2M the other PPO rows use. The asymmetry is also the one the ATLA paper itself reports under: its released configurations give vanilla PPO and SA-PPO 976 iterations of the same 2048-step rollout, i.e. the same 2M our PPO row uses, against 2441 for ATLA and ATLA-SA. Our PPO-versus-ATLA comparison therefore reproduces the budget pairing those results were established with rather than introducing a new one. The full online training budget is in Table~\ref{tab:algorithm_budget}.

\begin{table}[t]
\centering
\footnotesize
\setlength{\tabcolsep}{4pt}
\renewcommand{\arraystretch}{1.12}
\begin{tabular}{@{}
>{\raggedright\arraybackslash}p{0.22\linewidth}
>{\raggedright\arraybackslash}p{0.14\linewidth}
>{\centering\arraybackslash}p{0.15\linewidth}
>{\centering\arraybackslash}p{0.25\linewidth}
>{\centering\arraybackslash}p{0.12\linewidth}@{}}
\toprule
Task & Method & Env. steps & Wall-clock time (h)\\
\midrule
\multirow{3}{*}{\textit{Walker2d-v5}}
& PPO & 1M & $0.40 \pm 0.15$\\
& PPO & 2M & $0.80 \pm 0.29$\\
& SAC & 1M & $1.60 \pm 0.03$\\
\bottomrule
\end{tabular}
\caption{\textbf{Walker2d training wall-clock time on H100 GPUs.}
Values are mean $\pm$ sample standard deviation over three training seeds. Wall-clock time is
measured from the start of training to the corresponding checkpoint save.}
\label{tab:walker-walltime}
\end{table}

Taken together, the online budget is set by the base algorithm. Table~\ref{tab:online_budget_summary} summarizes the resulting protocol.
Methods sharing the same base algorithm are trained with the same budget,
while ATLA and ATLA-SA retain their official $\sim$5M-step configuration.

\begin{table}[t]
\centering
\small
\setlength{\tabcolsep}{5pt}
\renewcommand{\arraystretch}{1.12}
\begin{tabular}{lll}
\toprule
Budget rule & Methods & Steps \\
\midrule
SAC-based
& SAC, DR-SAC, RSC-SAC
& 1M \\

PPO/TRPO-based
& PPO, RARL-PPO, RARL-TRPO
& 2M \\

Official ATLA configuration
& ATLA, ATLA-SA
& $\sim$5M \\
\bottomrule
\end{tabular}
\caption{\textbf{Online training budget summary.}}
\label{tab:online_budget_summary}
\end{table}

\begin{table}[H]
\centering
\scriptsize
\setlength{\tabcolsep}{4pt}
\renewcommand{\arraystretch}{1.12}
\begin{tabular}{@{}l l l l@{}}
\toprule
Algorithm & Feature & Base algorithm& Budget \\
\midrule
\multicolumn{4}{@{}l}{\textit{Online}} \\
SAC & Standard & --- & 1M steps \\
PPO & Standard & --- & 2M steps \\
ATLA & Learner-centric & PPO & $2441{\times}2048$ steps \\
ATLA-SA & Learner-centric & PPO (LSTM) & $2441{\times}2048$ steps \\
RSC-SAC & Learner-centric & SAC & 1M steps \\
RARL-PPO / -TRPO & Env.-centric & PPO / TRPO & 2M steps \\
DR-SAC & Env.-centric & SAC & 1M steps \\
RAMU & Safe RL & PPO & Isaac Lab budget \\
SPiDR & Safe-robust & PPO & Isaac Lab budget \\
\midrule
\multicolumn{4}{@{}l}{\textit{Offline}} \\
IQL & Standard & --- & 1M, batch 256 \\
TD3+BC & Standard & --- & 1M, batch 256 \\
MOPO & Standard & SAC & 1M, batch 256 \\
SynthER & Standard & IQL & 1M, batch 256 \\
RFQI & Learner-centric & FQI & 500k, batch 1000 \\
RORL & Learner-centric & SAC & 3M (MuJoCo, Door) / 1M \\
ATLA-IQL & Learner-centric & IQL & 1M, batch 256 \\
RSC-IQL & Data-centric & IQL & 1M, batch 256 \\
RAMBO & Data-centric & SAC & 2M (MuJoCo, Door) / 1M \\
ROMB & Generative model & IQL & 1M, batch 256 \\
FWM & Generative model & IQL & 1M, batch 256 \\
PLR-PVL & Generative model & IQL & 1M, batch 256 \\
\bottomrule
\end{tabular}
\caption{\textbf{Training budget.} Budgets are environment
steps for online methods and policy/critic gradient updates for offline
methods.  Training budget alignment explanation is in Section~\ref{sec:alignment}. }
\label{tab:algorithm_budget}
\end{table}

For the offline algorithms' training budgets, Table~\ref{tab:algorithm_budget} shows nine of the twelve offline columns share the same 1M-update budget. RAMBO and RORL~\citep{rigter2022rambo,yang2022rorl} follow their official budgets on the tasks their releases cover and keep the 1M budget elsewhere. Thus, our MuJoCo comparisons with 1M-update
references preserve native budgets. RFQI~\citep{panaganti2022robust} follows its official budget and reports its own best checkpoint, because forcing it onto the common protocol would break its mechanism.

\begin{table}[t]
    \centering
    \footnotesize
    \setlength{\tabcolsep}{5pt}
    \renewcommand{\arraystretch}{1.1}
    \begin{tabular}{@{}l ccc r@{}}
        \toprule
        Task & PPO, DR-PPO, RAMU-PPO & ATLA-PPO & SPiDR-PPO & $B$ \\
        \midrule
        G1 & $1500 \times 4096 \times 24$ & $750 \times 2 \times 4096 \times 24$ & $900 \times 512 \times 320$ & $147{,}456{,}000$ \\
        Drawer & $3200 \times 512 \times 96$ & $1600 \times 2 \times 512 \times 96$ & $3200 \times 512 \times 96$ & $157{,}286{,}400$ \\
        \bottomrule
    \end{tabular}
    \caption{\textbf{Training budgets for Isaac Lab tasks.} Iterations are chosen so that the retained primary
    training interactions $B$ match within each task.}
    \label{tab:isaaclab_budget}
\end{table}

\newsavebox{\hparamfittablebox}
\newenvironment{hparamfittabular}[1]{%
  \begin{lrbox}{\hparamfittablebox}%
  \begin{tabular}{#1}%
}{%
  \end{tabular}%
  \end{lrbox}%
  \ifdim\wd\hparamfittablebox>\linewidth
    \resizebox{\linewidth}{!}{\usebox{\hparamfittablebox}}%
  \else
    \usebox{\hparamfittablebox}%
  \fi
}

\section{Hyperparameter settings}
\label{app:hparam_set}
\subsection{Online algorithm hyperparameters}
\label{app:online_algorithms_hyperparameters}
\label{app:online_experimental_setup}

The tables below report algorithm hyperparameters only. \textbf{H},  \textbf{HC}, and \textbf{W2D}
    denote \textbf{Hopper} ,  \textbf{HalfCheetah}, and \textbf{Walker2d}, respectively. A combined
 \textbf{H/HC/W2D} column is used only when all listed values agree across the
three tasks. Dashes indicate that a parameter does not apply.

\begin{table}[H]
\centering
\normalsize
\setlength{\tabcolsep}{3.5pt}
\renewcommand{\arraystretch}{1.06}
\begin{hparamfittabular}{@{}lcccc@{}}
\toprule
\textbf{Parameter} & \textbf{H/HC/W2D} & \textbf{PointMaze} & \textbf{FetchReach} & \textbf{CarRacing} \\
\midrule
Learning rate & \multicolumn{4}{c}{$3\times10^{-4}$} \\
Hidden layers & $2\times64$ & $2\times64$ & $2\times64$ & NatureCNN \\
Discount factor $\gamma$ & \multicolumn{4}{c}{$0.99$} \\
GAE parameter $\lambda$ & \multicolumn{4}{c}{$0.95$} \\
PPO clipping parameter & \multicolumn{4}{c}{$0.2$} \\
Value-loss coefficient & \multicolumn{4}{c}{$0.5$} \\
Entropy coefficient & \multicolumn{4}{c}{$0$} \\
Rollout length & \multicolumn{4}{c}{$2048$} \\
Minibatch size & \multicolumn{4}{c}{$64$} \\
Optimization epochs & \multicolumn{4}{c}{$10$} \\
Maximum gradient norm & \multicolumn{4}{c}{$0.5$} \\
\bottomrule
\end{hparamfittabular}
\caption{\textbf{PPO hyperparameters.}}
\label{tab:ppo_hparams}
\end{table}

\begin{table}[H]
\centering
\normalsize
\setlength{\tabcolsep}{3.5pt}
\renewcommand{\arraystretch}{1.06}
\begin{hparamfittabular}{@{}lcccc@{}}
\toprule
\textbf{Parameter} & \textbf{H/HC/W2D} & \textbf{PointMaze} & \textbf{FetchReach} & \textbf{CarRacing} \\
\midrule
Learning rate & $3\times10^{-4}$ & $3\times10^{-4}$ & $3\times10^{-4}$ & $10^{-4}$ \\
Hidden layers (actor / critic) & $2\times256$ & $2\times256$ & $2\times256$ & NatureCNN $+2\times256$ \\
Target smoothing $\tau$ & $0.005$ & $0.005$ & $0.005$ & $0.02$ \\
Replay-buffer size & $10^6$ & $10^6$ & $10^6$ & $2\times10^5$ \\
Learning starts & $100$ & $100$ & $100$ & $10^4$ \\
Train frequency / gradient steps & $1/1$ & $1/1$ & $1/1$ & $8/8$ \\
\midrule
Discount factor $\gamma$ & \multicolumn{4}{c}{$0.99$} \\
Target entropy & automatic & automatic & automatic & automatic ($-3$) \\
Minibatch size & \multicolumn{4}{c}{$256$} \\
\bottomrule
\end{hparamfittabular}
\caption{\textbf{SAC hyperparameters.}}
\label{tab:sac_hparams}
\end{table}

\begin{table}[H]
\centering
\normalsize
\setlength{\tabcolsep}{2.4pt}
\renewcommand{\arraystretch}{1.06}
\begin{hparamfittabular}{@{}lcccccc@{}}
\toprule
\textbf{Parameter} & \textbf{Hopper} & \textbf{Walker2d} & \textbf{HalfCheetah}
& \textbf{PointMaze} & \textbf{FetchReach} & \textbf{CarRacing} \\
\midrule
Adversary target & foot & both feet & torso and feet & particle & Cartesian action & hull \\
Adversary magnitude & $5$ N & $5$ N & $5$ N & $10$ N & $0.1$ & $5$ \\
Update ratio & $1/1$ & $1/1$ & $1/1$ & $2/1$ & $1/1$ & $2/1$ \\
Adversary warm-up & $0$ & $0$ & $0$ & $5\times10^4$ & $0$ & $4\times10^4$ \\
Zero-force probability $p_0$ & $0$ & $0$ & $0$ & $0.25$ & $0$ & $0.25$ \\
Protagonist/adversary lr & $3\times10^{-4}$ & $3\times10^{-4}$ & $3\times10^{-4}$
& $3\times10^{-4}$ & $3\times10^{-4}$ & $2.5\times10^{-4}$ \\
Entropy coefficient & $0$ & $0$ & $0$ & $0$ & $0.01$ & $0.01$ \\
\midrule
Discount factor $\gamma$ & \multicolumn{6}{c}{$0.99$} \\
GAE parameter $\lambda$ & \multicolumn{6}{c}{$0.95$} \\
PPO clipping parameter & \multicolumn{6}{c}{$0.2$} \\
\bottomrule
\end{hparamfittabular}
\caption{\textbf{RARL-PPO hyperparameters.} The update ratio denotes protagonist-to-adversary
update blocks.}
\label{tab:rarl_ppo_hparams}
\end{table}

\begin{table}[H]
\centering
\normalsize
\setlength{\tabcolsep}{2.4pt}
\renewcommand{\arraystretch}{1.06}
\begin{hparamfittabular}{@{}lcccccc@{}}
\toprule
\textbf{Parameter} & \textbf{Hopper} & \textbf{Walker2d} & \textbf{HalfCheetah}
& \textbf{PointMaze} & \textbf{FetchReach} & \textbf{CarRacing} \\
\midrule
Adversary target & foot & both feet & torso and feet & particle & Cartesian action & hull \\
Adversary magnitude & $5$ N & $5$ N & $5$ N & $10$ N & $0.1$ & $2$ \\
Update ratio & $1/1$ & $1/1$ & $1/1$ & $2/1$ & $1/1$ & $5/1$ \\
Adversary warm-up & $0$ & $0$ & $0$ & $5\times10^4$ & $0$ & $3\times10^5$ \\
Zero-force probability $p_0$ & $0$ & $0$ & $0$ & $0.25$ & $0$ & $0.5$ \\
Protagonist learning rate & $10^{-3}$ & $10^{-3}$ & $10^{-3}$ & $10^{-3}$ & $10^{-3}$ & $10^{-4}$ \\
Adversary learning rate & $10^{-3}$ & $10^{-3}$ & $10^{-3}$ & $10^{-3}$ & $10^{-3}$ & $5\times10^{-5}$ \\
\midrule
Discount factor $\gamma$ & \multicolumn{6}{c}{$0.99$} \\
GAE parameter $\lambda$ & \multicolumn{6}{c}{$0.95$} \\
Target KL divergence & \multicolumn{6}{c}{$0.01$} \\
CG steps / damping & \multicolumn{6}{c}{$15/0.1$} \\
Critic updates / policy update & \multicolumn{6}{c}{$10$} \\
\bottomrule
\end{hparamfittabular}
\caption{\textbf{RARL-TRPO hyperparameters.} The update ratio denotes protagonist-to-adversary
update blocks; CG denotes conjugate gradient.}
\label{tab:rarl_trpo_hparams}
\end{table}

\begin{table}[H]
\centering
\normalsize
\setlength{\tabcolsep}{3.5pt}
\renewcommand{\arraystretch}{1.06}
\begin{hparamfittabular}{@{}lcccc@{}}
\toprule
\textbf{Parameter} & \textbf{H/HC/W2D} & \textbf{PointMaze} & \textbf{FetchReach} & \textbf{CarRacing} \\
\midrule
Randomized parameters & $g$/morph./gear & friction/gear & morph./gear & friction/morph./gear \\
Randomization range & $[0.9,1.1]$ & $[0,7.5]/[0.9,1.1]$ & $[0.9,1.1]$ & $[0.9,1.1]$ \\
Learning rate & $3\times10^{-4}$ & $3\times10^{-4}$ & $10^{-3}$ & $10^{-4}$ \\
Learning-rate schedule & constant & constant & constant & linear to $3\times10^{-5}$ \\
Discount factor $\gamma$ & $0.99$ & $0.99$ & $0.95$ & $0.99$ \\
\midrule
Target smoothing $\tau$ & \multicolumn{4}{c}{$0.005$} \\
Entropy coefficient & \multicolumn{4}{c}{automatic tuning} \\
Target entropy & automatic & automatic & automatic & $-2$ (fixed) \\
\bottomrule
\end{hparamfittabular}
\caption{\textbf{DR-SAC hyperparameters.} Physical parameters are sampled independently
at reset and remain fixed within an episode. Morph. represents Morphology}
\label{tab:dr_sac_hparams}
\end{table}

\begin{table}[H]
\centering
\normalsize
\setlength{\tabcolsep}{3.5pt}
\renewcommand{\arraystretch}{1.06}
\begin{hparamfittabular}{@{}lcccc@{}}
\toprule
\textbf{Parameter} & \textbf{H/HC/W2D} & \textbf{PointMaze} & \textbf{FetchReach} & \textbf{CarRacing} \\
\midrule
Actor learning rate & $3\times10^{-4}$ & $3\times10^{-4}$ & $3\times10^{-4}$ & $10^{-4}$ \\
Critic learning rate & $10^{-3}$ & $3\times10^{-4}$ & $10^{-3}$ & $10^{-4}$ \\
Learning-rate schedule & constant & constant & constant & linear to $3\times10^{-5}$ \\
Causal/RSC warm-up & $10^5$ & $3\times10^5$ & $10^5$ & $4\times10^5$ \\
Replay augmentation / RSC mix & $0.25$ & $0.10$ & $0.25$ & $0.05$ \\
Selected fraction & -- & $0.25$ & -- & $0.25$ \\
Reference size & $128$ & -- & $128$ & -- \\
Dynamics learning rate & -- & $10^{-4}$ & -- & $10^{-4}$ \\
\midrule
Discount factor $\gamma$ & \multicolumn{4}{c}{$0.99$} \\
\bottomrule
\end{hparamfittabular}
\caption{\textbf{RSC-SAC hyperparameters.}}
\label{tab:rsc_sac_hparams}
\end{table}

\begin{table}[H]
\centering
\normalsize
\setlength{\tabcolsep}{2.4pt}
\renewcommand{\arraystretch}{1.06}
\begin{hparamfittabular}{@{}lcccccc@{}}
\toprule
\textbf{Parameter} & \textbf{Hopper} & \textbf{Walker2d} & \textbf{HalfCheetah}
& \textbf{PointMaze} & \textbf{FetchReach} & \textbf{CarRacing} \\
\midrule
Observation radius $\epsilon$ & $0.075$ & $0.05$ & $0.15$ & $0.075$ & $0.075$ & $8/255$ \\
Policy learning rate & $3\times10^{-4}$ & $4\times10^{-4}$ & $10^{-4}$
& $3\times10^{-4}$ & $3\times10^{-4}$ & $10^{-4}$ \\
Value learning rate & $2.5\times10^{-4}$ & $3\times10^{-4}$ & $10^{-3}$
& $2.5\times10^{-4}$ & $2.5\times10^{-4}$ & $10^{-4}$ \\
Adversary policy lr & $3\times10^{-3}$ & $10^{-3}$ & $3\times10^{-3}$
& $3\times10^{-3}$ & $3\times10^{-3}$ & $10^{-4}$ \\
Adversary value lr & $3\times10^{-5}$ & $10^{-4}$ & $3\times10^{-4}$
& $3\times10^{-5}$ & $3\times10^{-5}$ & -- \\
Adversary updates per block & $1$ & $1$ & $1$ & $1$ & $1$ & $4$ \\
Initial policy std. dev. & $1.0$ & $1.0$ & $0.5$ & $1.0$ & $1.0$ & -- \\
\midrule
Discount factor $\gamma$ & \multicolumn{6}{c}{$0.99$} \\
GAE parameter $\lambda$ & \multicolumn{6}{c}{$0.95$} \\
PPO clipping parameter & \multicolumn{6}{c}{$0.2$} \\
\bottomrule
\end{hparamfittabular}
\caption{\textbf{ATLA-PPO hyperparameters.} }
\label{tab:atla_hparams}
\end{table}

\begin{table}[H]
\centering
\normalsize
\setlength{\tabcolsep}{2.4pt}
\renewcommand{\arraystretch}{1.06}
\begin{hparamfittabular}{@{}lcccccc@{}}
\toprule
\textbf{Parameter} & \textbf{Hopper} & \textbf{Walker2d} & \textbf{HalfCheetah}
& \textbf{PointMaze} & \textbf{FetchReach} & \textbf{CarRacing} \\
\midrule
Observation radius $\epsilon$ & $0.075$ & $0.05$ & $0.15$ & $0.075$ & $0.075$ & $8/255$ \\
SA regularization weight & $0.3$ & $0.3$ & $0.03$ & $0.3$ & $0.3$ & $0.1$ \\
Policy learning rate & $10^{-3}$ & $10^{-3}$ & $10^{-3}$ & $10^{-3}$ & $10^{-3}$ & $10^{-4}$ \\
Value learning rate & $10^{-2}$ & $3\times10^{-2}$ & $3\times10^{-3}$
& $10^{-2}$ & $10^{-2}$ & $10^{-4}$ \\
Adversary policy lr & $10^{-2}$ & $3\times10^{-4}$ & $10^{-2}$
& $10^{-2}$ & $10^{-2}$ & $10^{-4}$ \\
Adversary value lr & $10^{-2}$ & $10^{-2}$ & $3\times10^{-4}$
& $10^{-2}$ & $10^{-2}$ & -- \\
Adversary updates per block & $1$ & $1$ & $1$ & $1$ & $1$ & $4$ \\
Initial policy std. dev. & $1.0$ & $1.0$ & $0.5$ & $1.0$ & $1.0$ & -- \\
History length & $100$ & $100$ & $100$ & $100$ & $100$ & -- \\
\midrule
Discount factor $\gamma$ & \multicolumn{6}{c}{$0.99$} \\
GAE parameter $\lambda$ & \multicolumn{6}{c}{$0.95$} \\
PPO clipping parameter & \multicolumn{6}{c}{$0.2$} \\
State-adversarial inner steps & \multicolumn{6}{c}{$2$} \\
\bottomrule
\end{hparamfittabular}
\caption{\textbf{ATLA-SA-PPO hyperparameters.} State-task $\epsilon$
values act on normalized observations; CarRacing uses $8$ raw pixel levels,
equivalent to $8/255$ after image normalization.}
\label{tab:atla_sa_hparams}
\end{table}

\subsection{Offline algorithm hyperparameters}
\label{app:offline_hparams}

The tables below give the algorithm hyperparameters behind the reported
offline results. Dashes indicate that a parameter does not apply.

\begin{table}[H]
\centering
\normalsize
\setlength{\tabcolsep}{3pt}
\renewcommand{\arraystretch}{1.06}
\begin{hparamfittabular}{@{}lcccccc@{}}
\toprule
\textbf{Parameter} & \textbf{Hopper} & \textbf{Walker2d} & \textbf{HalfCheetah}
& \textbf{Door} & \textbf{PointMaze} & \textbf{LunarLander} \\
\midrule
AWR temperature $\beta$ & $3.0$ & $3.0$ & $3.0$ & $0.5$ & $3.0$ & $3.0$ \\
Policy dropout & --- & --- & --- & $0.1$ & --- & --- \\
\midrule
Expectile $\tau$ & \multicolumn{6}{c}{$0.7$} \\
Advantage clip & \multicolumn{6}{c}{$100$} \\
Learning rate & \multicolumn{6}{c}{$3\times10^{-4}$} \\
Hidden layers & \multicolumn{6}{c}{$2\times256$} \\
Discount factor $\gamma$ & \multicolumn{6}{c}{$0.99$} \\
Target EMA rate & \multicolumn{6}{c}{$0.005$} \\
Batch size & \multicolumn{6}{c}{$256$} \\
\bottomrule
\end{hparamfittabular}
\caption{\textbf{IQL hyperparameters.} }
\label{tab:hparams_iql}
\end{table}

\begin{table}[H]
\centering
\normalsize
\setlength{\tabcolsep}{6pt}
\renewcommand{\arraystretch}{1.06}
\begin{hparamfittabular}{@{}lclc@{}}
\toprule
\textbf{Parameter} & \textbf{Value} & \textbf{Parameter} & \textbf{Value} \\
\midrule
BC weight $\alpha$ & $2.5$ & Target policy noise & $0.2$ \\
Noise clip & $0.5$ & Policy delay & $2$ \\
Learning rate & $3\times10^{-4}$ & Hidden layers & $2\times256$ \\
Discount factor $\gamma$ & $0.99$ & Target EMA rate & $0.005$ \\
State normalization & on & Batch size & $256$ \\
\bottomrule
\end{hparamfittabular}
\caption{\textbf{TD3+BC hyperparameters.} }
\label{tab:hparams_td3bc}
\end{table}

\begin{table}[H]
\centering
\normalsize
\setlength{\tabcolsep}{3pt}
\renewcommand{\arraystretch}{1.06}
\begin{hparamfittabular}{@{}lcccccc@{}}
\toprule
\textbf{Parameter} & \textbf{Hopper} & \textbf{Walker2d} & \textbf{HalfCheetah}
& \textbf{Door} & \textbf{PointMaze} & \textbf{LunarLander} \\
\midrule
Rollout horizon $h$ & $42$ & $3$ & $5$ & $1$ & $5$ & $5$ \\
Penalty coeff.\ $\lambda$ & $37.28$ & $1.0$ & $1.0$ & $5.0$ & $1.0$ & $1.0$ \\
Penalty type & ens.-std & ens.-std & max-aleat. & max-aleat. & max-aleat. & max-aleat. \\
Real data ratio & $0.05$ & $0.5$ & $0.05$ & $0.5$ & $0.05$ & $0.05$ \\
\midrule
Dynamics ensemble & \multicolumn{6}{c}{$7$ networks, $5$ elites} \\
Rollouts per refresh & \multicolumn{6}{c}{$5\times10^{4}$} \\
Target entropy & \multicolumn{6}{c}{$-3$} \\
Learning rate & \multicolumn{6}{c}{$3\times10^{-4}$} \\
Discount factor $\gamma$ & \multicolumn{6}{c}{$0.99$} \\
Target EMA rate & \multicolumn{6}{c}{$0.005$} \\
Batch size & \multicolumn{6}{c}{$256$} \\
\bottomrule
\end{hparamfittabular}
\caption{\textbf{MOPO hyperparameters.} The penalty type
names the ensemble statistic the reward is discounted by: \texttt{max-aleat.}
is the maximum per-member aleatoric standard deviation of the original paper,
\texttt{ens.-std} the standard deviation across members, which \citep{lu2021revisiting} introduce and find to work better than the original penalty at much longer rollout horizons. The Hopper column takes its three values directly.}
\label{tab:hparams_mopo}
\end{table}

\begin{table}[H]
\centering
\normalsize
\setlength{\tabcolsep}{6pt}
\renewcommand{\arraystretch}{1.06}
\begin{hparamfittabular}{@{}lclc@{}}
\toprule
\textbf{Parameter} & \textbf{Value} & \textbf{Parameter} & \textbf{Value} \\
\midrule
Denoiser & residual MLP & Width & $2048$ \\
Layers & $6$ & Time embedding & $128$ \\
Fourier features & random & Sampling steps & $128$ \\
$\sigma_{\text{data}}$ & $1.0$ & $S_{\text{churn}}$ & $80$ \\
Learning rate & $3\times10^{-4}$ & Schedule & cosine \\
Batch size & $1024$ & Terminals modeled & yes \\
\bottomrule
\end{hparamfittabular}
\caption{\textbf{SynthER hyperparameters.} The consumer is the IQL agent of Table~\ref{tab:hparams_iql},
trained on the generated dataset with normalization statistics taken from the
real one.}
\label{tab:hparams_synther}
\end{table}

\begin{table}[H]
\centering
\normalsize
\setlength{\tabcolsep}{3pt}
\renewcommand{\arraystretch}{1.06}
\begin{hparamfittabular}{@{}lcccccc@{}}
\toprule
\textbf{Parameter} & \textbf{Hopper} & \textbf{Walker2d} & \textbf{HalfCheetah}
& \textbf{Door} & \textbf{PointMaze} & \textbf{LunarLander} \\
\midrule
Adversary weight $\lambda_{\text{adv}}$ & $3\times10^{-4}$ & $0$ & $3\times10^{-4}$
& $3\times10^{-4}$ & $3\times10^{-4}$ & $3\times10^{-4}$ \\
Horizon $h$ & $5$ & $5$ & $5$ & $1$ & $1$ & $1$ \\
BC anchor weight & $1.0$ & --- & --- & $1.0$ & $1.0$ & $1.0$ \\
Target entropy & $-3$ & $-6$ & $-6$ & $-90$ & $-2$ & $-2$ \\
\midrule
Real data ratio & \multicolumn{6}{c}{$0.5$} \\
Dynamics ensemble & \multicolumn{6}{c}{$7$ networks, $5$ elites} \\
Rollouts per refresh & \multicolumn{6}{c}{$5\times10^{4}$} \\
Critic / actor lr & \multicolumn{6}{c}{$3\times10^{-4}$ / $1\times10^{-4}$} \\
Adversary lr & \multicolumn{6}{c}{$3\times10^{-4}$} \\
Discount factor $\gamma$ & \multicolumn{6}{c}{$0.99$} \\
Target EMA rate & \multicolumn{6}{c}{$5\times10^{-3}$} \\
Batch size & \multicolumn{6}{c}{$256$} \\
\bottomrule
\end{hparamfittabular}
\caption{\textbf{RAMBO hyperparameters.} The BC anchor is a behaviour-cloning term kept on the
actor throughout training. }
\label{tab:hparams_rambo}
\end{table}

\begin{table}[H]
\centering
\normalsize
\setlength{\tabcolsep}{3pt}
\renewcommand{\arraystretch}{1.06}
\begin{hparamfittabular}{@{}lcccccc@{}}
\toprule
\textbf{Parameter} & \textbf{Hopper} & \textbf{Walker2d} & \textbf{HalfCheetah}
& \textbf{Door} & \textbf{PointMaze} & \textbf{LunarLander} \\
\midrule
Uncertainty radius $\rho$ & $0.5$ & $0.3$ & $0.3$ & $0.5$ & $0.3$ & $0.3$ \\
Actor lr & $1\times10^{-4}$ & $3\times10^{-4}$ & $3\times10^{-4}$
& $3\times10^{-4}$ & $3\times10^{-4}$ & $3\times10^{-4}$ \\
Critic lr & $6\times10^{-4}$ & $8\times10^{-4}$ & $8\times10^{-4}$
& $8\times10^{-4}$ & $8\times10^{-4}$ & $8\times10^{-4}$ \\
\midrule
Perturbation limit $\Phi$ & \multicolumn{6}{c}{$0.1$} \\
Convex weight $\lambda$ & \multicolumn{6}{c}{$0.75$} \\
Dual tolerance / max steps & \multicolumn{6}{c}{$10^{-3}$ / $10^{4}$} \\
Target / policy samples & \multicolumn{6}{c}{$10$ / $100$} \\
Discount factor $\gamma$ & \multicolumn{6}{c}{$0.99$} \\
Target EMA rate & \multicolumn{6}{c}{$0.005$} \\
Batch size & \multicolumn{6}{c}{$1000$} \\
\bottomrule
\end{hparamfittabular}
\caption{\textbf{RFQI hyperparameters.}}
\label{tab:hparams_rfqi}
\end{table}

\begin{table}[H]
\centering
\normalsize
\setlength{\tabcolsep}{3pt}
\renewcommand{\arraystretch}{1.06}
\begin{hparamfittabular}{@{}lcccccc@{}}
\toprule
\textbf{Parameter} & \textbf{Hopper} & \textbf{Walker2d} & \textbf{HalfCheetah}
& \textbf{Door} & \textbf{PointMaze} & \textbf{LunarLander} \\
\midrule
Smoothing samples & $20$ & $20$ & $10$ & $20$ & $20$ & $20$ \\
Policy smooth $\epsilon$ & $0.005$ & $0.01$ & $0.001$ & $0.01$ & $0.005$ & $0.005$ \\
Policy smooth weight & $0.1$ & $1.0$ & $0.1$ & $1.0$ & $0.1$ & $0.1$ \\
$Q$ smooth $\epsilon$ & $0.005$ & $0.01$ & $0.001$ & $0.01$ & $0.005$ & $0.005$ \\
$Q$ OOD $\epsilon$ & $0.01$ & $0.01$ & $0$ & $0.01$ & $0.01$ & $0.01$ \\
$Q$ OOD weight & $0.5$ & $0.1$ & $0$ & $0.1$ & $0.5$ & $0.5$ \\
Uncert.\ weight, init / min & $2.0$ / $0.1$ & $1.0$ / $0.1$ & $0$ / $0$
& $1.0$ / $0.1$ & --- & --- \\
Uncertainty decay & $10^{-6}$ & $5\times10^{-7}$ & $0$ & $5\times10^{-7}$ & --- & --- \\
\midrule
BC anchor weight & \multicolumn{6}{c}{$1.0$, from update $2\times10^{4}$} \\
Critic ensemble & \multicolumn{6}{c}{$10$ critics} \\
$Q$ smooth weight / $\tau$ & \multicolumn{6}{c}{$10^{-4}$ / $0.2$} \\
Hidden layers & \multicolumn{6}{c}{$3\times256$} \\
Policy / critic lr & \multicolumn{6}{c}{$3\times10^{-4}$} \\
Discount factor $\gamma$ & \multicolumn{6}{c}{$0.99$} \\
Target EMA rate & \multicolumn{6}{c}{$0.005$} \\
Batch size & \multicolumn{6}{c}{$256$} \\
\bottomrule
\end{hparamfittabular}
\caption{\textbf{RORL hyperparameters.}}
\label{tab:hparams_rorl}
\end{table}

\begin{table}[H]
\centering
\normalsize
\setlength{\tabcolsep}{6pt}
\renewcommand{\arraystretch}{1.06}
\begin{hparamfittabular}{@{}lclc@{}}
\toprule
\textbf{Parameter} & \textbf{Value} & \textbf{Parameter} & \textbf{Value} \\
\midrule
Adversary radius $\epsilon$ & $0.025$ & $\epsilon$ ramp length & $1.5\times10^{4}$ updates \\
Adversarial  actor loss & $0.1$ & Adversary lr & $3\times10^{-4}$ \\
$L_2$ regulariser & $0.02$ & EMA & $0.995$ \\
\bottomrule
\end{hparamfittabular}
\caption{\textbf{ATLA-IQL hyperparameters. }}
\label{tab:hparams_atla_iql}
\end{table}

\begin{table}[H]
\centering
\normalsize
\setlength{\tabcolsep}{6pt}
\renewcommand{\arraystretch}{1.06}
\begin{hparamfittabular}{@{}lclc@{}}
\toprule
\textbf{Parameter} & \textbf{Value} & \textbf{Parameter} & \textbf{Value} \\
\midrule
Rewrite mode & causal permutation & Rewritten fraction of batch & $0.25$ \\
Mask selection threshold & $0.7$ & Perturbation probability scale & $0.3$ \\
Mask sparsity weight & $0.5$  \\
\bottomrule
\end{hparamfittabular}
\caption{\textbf{RSC-IQL hyperparameters.}}
\end{table}

\begin{table}[H]
\centering
\normalsize
\setlength{\tabcolsep}{6pt}
\renewcommand{\arraystretch}{1.06}
\begin{hparamfittabular}{@{}lclc@{}}
\toprule
\textbf{Parameter} & \textbf{Value} & \textbf{Parameter} & \textbf{Value} \\
\midrule
Dynamics ensemble & $7$ networks, $5$ elites & Rollout horizon & $5$ \\
GAE parameter $\lambda$ & $0.95$ & Adversary clip range & $0.2$ \\
Supervised anchor weight & $3000$ & Model share of critic target & $0.5$ \\
Stackelberg coefficient & $0$ & & \\
\bottomrule
\end{hparamfittabular}
\caption{\textbf{ROMB hyperparameters.}}
\label{tab:hparams_romb}
\end{table}

\begin{table}[H]
\centering
\normalsize
\setlength{\tabcolsep}{6pt}
\renewcommand{\arraystretch}{1.06}
\begin{hparamfittabular}{@{}lclc@{}}
\toprule
\textbf{Parameter} & \textbf{Value} & \textbf{Parameter} & \textbf{Value} \\
\midrule
Horizon $h$ & $5$ & Guidance weight & $2.0$ \\
ODE sampling steps / solver & $5$ / midpoint & RTG conditioning & on \\
\addlinespace[2pt]
Rollout window $T$ & $12$ & Backbone & U-Net, $128$ features \\
Learning rate & $10^{-4}$ & Conditioning dropout & $0.2$ \\
Batch size & $256$ & & \\
\bottomrule
\end{hparamfittabular}
\caption{\textbf{FWM hyperparameters.}}
\label{tab:hparams_fwm}
\end{table}

\begin{table}[H]
\centering
\normalsize
\setlength{\tabcolsep}{6pt}
\renewcommand{\arraystretch}{1.06}
\begin{hparamfittabular}{@{}lclc@{}}
\toprule
\textbf{Parameter} & \textbf{Value} & \textbf{Parameter} & \textbf{Value} \\
\midrule
Flow-model ensemble & $3$ & PVL decay $\lambda$ & $0.95$ \\
PVL discount $\gamma\lambda$ & $0.9405$ & Ensemble diversity & training seeds \\
PVL horizon & $h=5$ & Advantage clipping & positive part \\
\bottomrule
\end{hparamfittabular}
\caption{\textbf{PLR-PVL hyperparameters.}}
\label{tab:hparams_plr_pvl}
\end{table}

\subsection{Robust safe algorithm hyperparameters}
\label{app:safe_algorithms_hyperparameters}

\begin{table}[H]
\centering
\normalsize
\setlength{\tabcolsep}{6pt}
\renewcommand{\arraystretch}{1.06}
\begin{hparamfittabular}{@{}lclc@{}}
\toprule
\textbf{Parameter} & \textbf{Value} & \textbf{Parameter} & \textbf{Value} \\
\midrule
\multicolumn{4}{@{}l}{\emph{G1}} \\
Actor and critic & 256-128-128, ELU & Observation normalization & off \\
Learning rate & $10^{-3}$, adaptive & KL target & $0.01$ \\
Discount $\gamma$ & $0.99$ & GAE $\lambda$ & $0.95$ \\
Entropy coefficient & $0.008$ & Clip $\epsilon$ & $0.2$ \\
\addlinespace[2pt]
\multicolumn{4}{@{}l}{\emph{Drawer}} \\
Actor and critic & 256-128-64, ELU & Observation normalization & off \\
Learning rate & $5\times10^{-4}$, fixed & KL target & $0.02$ \\
Discount $\gamma$ & $0.99$ & GAE $\lambda$ & $0.95$ \\
Entropy coefficient & $0.001$ & Clip $\epsilon$ & $0.2$ \\
\bottomrule
\end{hparamfittabular}
\caption{\textbf{PPO hyperparameters.}}
\label{tab:hparams_ppo}
\end{table}

\begin{table}[H]
\centering
\normalsize
\setlength{\tabcolsep}{6pt}
\renewcommand{\arraystretch}{1.06}
\begin{hparamfittabular}{@{}lclc@{}}
\toprule
\textbf{Parameter} & \textbf{Value} & \textbf{Parameter} & \textbf{Value} \\
\midrule
\multicolumn{4}{@{}l}{\emph{G1}} \\
Observation corruption & on & Push, every $10$--$15$\,s & $v_x, v_y \sim \mathcal{U}[-0.5, 0.5]$\,m/s \\
\addlinespace[2pt]
\multicolumn{4}{@{}l}{\emph{Drawer}} \\
Robot friction & $\mathcal{U}[0.8, 1.25]$ & Handle friction & static $\mathcal{U}[1.0, 1.25]$, dynamic $\mathcal{U}[1.25, 1.5]$ \\
Cabinet mass & $\times\,\mathcal{U}[0.8, 1.2]$ & Arm $K_p$, $K_d$ & $\times\,\mathcal{U}[0.9, 1.1]$ \\
Drawer damping & $\times\,\mathcal{U}[0.5, 1.5]$ & Action delay & $0$--$2$ policy steps \\
\bottomrule
\end{hparamfittabular}
\caption{\textbf{DR-PPO hyperparameters.}}
\label{tab:safe_drppo}
\end{table}

\begin{table}[H]
\centering
\normalsize
\setlength{\tabcolsep}{6pt}
\renewcommand{\arraystretch}{1.06}
\begin{hparamfittabular}{@{}lclc@{}}
\toprule
\textbf{Parameter} & \textbf{Value} & \textbf{Parameter} & \textbf{Value} \\
\midrule
Adversary network & 128-128, ELU & Adversary lr & $10^{-3}$ \\
Adversary clip & $0.2$ & Adversary entropy coefficient & $0$ \\
\addlinespace[2pt]
Perturbation & $\epsilon\tanh(\cdot)$, $\epsilon = 0.05$ & Phases per outer iteration & $1 + 1$ \\
\bottomrule
\end{hparamfittabular}
\caption{\textbf{ATLA-PPO hyperparameters.}}
\label{tab:hparams_atla}
\end{table}

\begin{table}[H]
\centering
\normalsize
\setlength{\tabcolsep}{6pt}
\renewcommand{\arraystretch}{1.06}
\begin{hparamfittabular}{@{}lclc@{}}
\toprule
\textbf{Parameter} & \textbf{Value} & \textbf{Parameter} & \textbf{Value} \\
\midrule
Synthetic next states & $5$ & Uncertainty $\varepsilon$ & $0.1$ \\
Wang distortion $\eta$ & $0.75$ & Cost discount & $0.9$ \\
\addlinespace[2pt]
Episode cost budget & $20$ & Initial multiplier & $0.01$ \\
Dual learning rate & $0.01$ & Cost EMA $\beta$ & $0.9$ \\
KL guard, mean / max & $0.03$ / $0.3$ & Step halvings & $8$ \\
\bottomrule
\end{hparamfittabular}
\caption{\textbf{RAMU-PPO hyperparameters. }}
\label{tab:hparams_ramu}
\end{table}

\begin{table}[H]
\centering
\normalsize
\setlength{\tabcolsep}{6pt}
\renewcommand{\arraystretch}{1.06}
\begin{hparamfittabular}{@{}lclc@{}}
\toprule
\textbf{Parameter} & \textbf{Value} & \textbf{Parameter} & \textbf{Value} \\
\midrule
Auxiliary members per env & $8$ & Uncertainty penalty & $0.075 \times \mathrm{Var}(s')$ \\
Variance clip & $10^{3}$ & Cost discount & $0.9$ \\
\addlinespace[2pt]
Cost GAE $\lambda$ & $0.95$ & Episode cost budget & $20$ \\
Initial multiplier & $0.01$ & Dual learning rate & $0.01$ \\
\bottomrule
\end{hparamfittabular}
\caption{\textbf{SPiDR-PPO hyperparameters.} }
\label{tab:hparams_spidr}
\end{table}

\section{Additional results and analysis}
\label{app:additional_results}

\subsection{Analysis of robust methods' gains}
\label{app:gain_columns}

Figure~\ref{fig:decay} reports two gains for each algorithmic feature family. The
\textbf{darker column} is the \emph{base-relative gain}: each method with a
matched base learner is compared with that base algorithm under the same conditions, and
the differences are averaged within the family. Standalone recipes without a
matched reference learner, including RFQI, RORL, and RAMBO, are excluded; the
number of included methods is reported as $n$ in
Table~\ref{tab:gain_col}. The \textbf{lighter column} is the
\emph{standard-pool gain}: the full family's mean minus the mean of all six
Standard RL references. It therefore reflects both the robustness mechanism
and the strength of its underlying learner.

Brackets denote 95\% seed-level stratified bootstrap intervals from $10{,}000$
replicates. Each training seed contributes its mean over 20 evaluation
episodes, and seeds are resampled within each (method, task) stratum, with the
same draw reused across all conditions of that task. Methods and base algorithms are
separate runs and are resampled independently. 

The two different comparisons indicate that both online families retain positive base-relative gains
($+5.5$ and $+6.6$), whereas offline gains are smaller. The generative family
offers the cleanest comparison: all three methods use IQL, yielding a $+9.8$
gain over the standard pool but $+2.5$ over IQL. Its aggregate advantage thus
reflects both the added mechanism and a strong base algorithm, while fixed-data
coverage may limit the incremental benefit available offline.

\begin{table}[h]
\centering
\footnotesize
\setlength{\tabcolsep}{6pt}
\renewcommand{\arraystretch}{1.14}

\label{tab:gain_columns}
\begin{tabular}{@{}lccc@{}}
\toprule
Algorithm feature & Paired gain & Pooled gain & $n$ paired \\
\midrule
Learner-centric (Online)  & $+5.5$ $[+2.8, +8.3]$ & $+9.5$ $[+7.2, +11.8]$ & 3 \\
Environment-centric       & $+6.6$ $[+3.8, +9.6]$ & $+9.6$ $[+7.3, +11.8]$ & 2 \\
Learner-centric (Offline) & $+3.8$ $[+0.8, +6.4]$ & $+2.9$ $[+1.3, +4.4]$  & 1 \\
Data-centric              & $-0.1$ $[-2.7, +2.1]$ & $+7.6$ $[+4.2, +11.1]$ & 1 \\
Generative model          & $+2.5$ $[-0.3, +4.9]$ & $+9.8$ $[+7.9, +11.7]$ & 3 \\
\bottomrule
\end{tabular}
\caption{\textbf{The two columns of the  algorithm feature panel, with intervals.}
Point estimates and $95\%$ seed-level stratified bootstrap intervals from
$10{,}000$ replicates. \emph{Paired} subtracts each member's own base
algorithm; \emph{pooled} subtracts the mean of all six Standard RL references.
$n$ is the number of members that have a matched base learner and therefore enter
the paired column.}
\label{tab:gain_col}
\end{table}

\FloatBarrier
\subsection{Analysis of task capability: On-policy vs Off-policy}

On-policy methods update a policy using trajectories collected by that policy or a recent version of it, as in PPO~\citep{schulman2017proximal}. Off-policy methods can also learn from data generated by earlier or different policies, often through experience replay, as in SAC~\citep{haarnoja2018soft}. This distinction concerns which data can be used for an update.

\begin{figure}[t]
    \centering
    \includegraphics[width=1\linewidth]{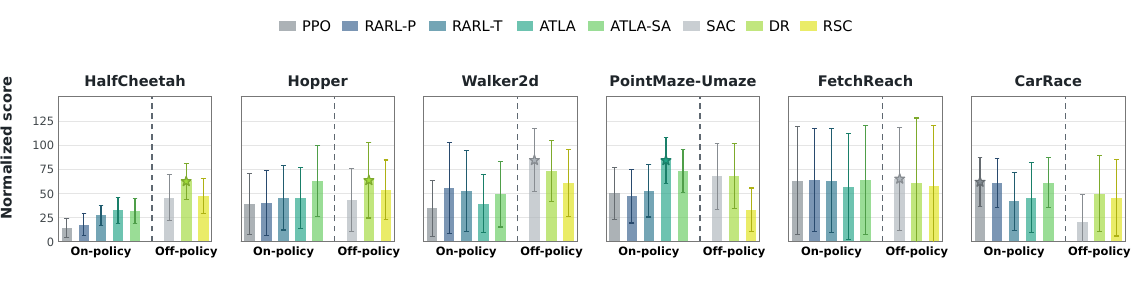}
    \caption{\textbf{On-policy and off-policy performance aggregated by task.}}
    \label{fig:algorithm_task_aggregate}
\end{figure}

Figure~\ref{fig:algorithm_task_aggregate} shows that the evaluated off-policy methods generally achieve higher scores on the MuJoCo locomotion tasks, while on-policy methods are stronger on PointMaze-UMaze and CarRace. In
locomotion, frequent reward feedback allows replay-based learners to reuse
informative transitions across many updates, which may contribute to the
off-policy methods' advantage. PointMaze requires discovering goal-reaching
trajectories, while CarRacing couples control with visual input; updates
from recent on-policy rollouts may be more useful in these settings.

\subsection{Results of advanced robotic tasks}
\label{sec:isaac_res}

\begin{figure}[t]
    \centering
    \includegraphics[width=1\linewidth]{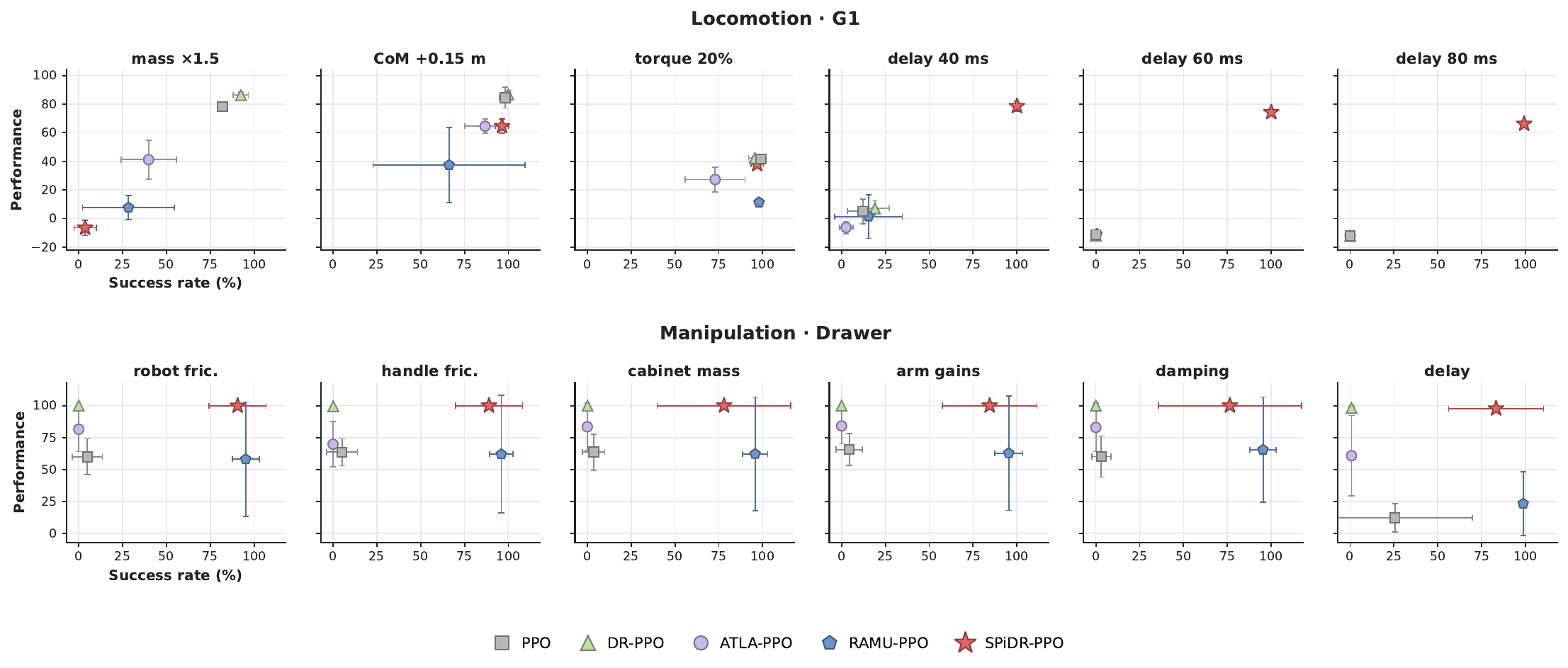}
    \caption{\textbf{Isaac Lab results.} Each panel
repeats Figure~\ref{fig:part4} on one factor: safety success rate against
performance, one point per algorithm, bars the s.d.\ over three training seeds.}

    \label{fig:isaac_supp}
\end{figure}

Figure~\ref{fig:isaac_supp} indicates that across the Isaac Lab tasks, SPiDR-PPO most consistently balances performance
and safety: it maintains high success while satisfying the cost budget on
Drawer, and leads on both return and survival under G1 action delays. These results show why high return
alone is insufficient for sim-to-real evaluation. Robust-safe methods must
be assessed for constraint satisfaction alongside task performance.

\subsection{Supplementary results}
\label{app:supplementary}

\begin{figure}[h]
    \centering
    \includegraphics[width=1\linewidth]{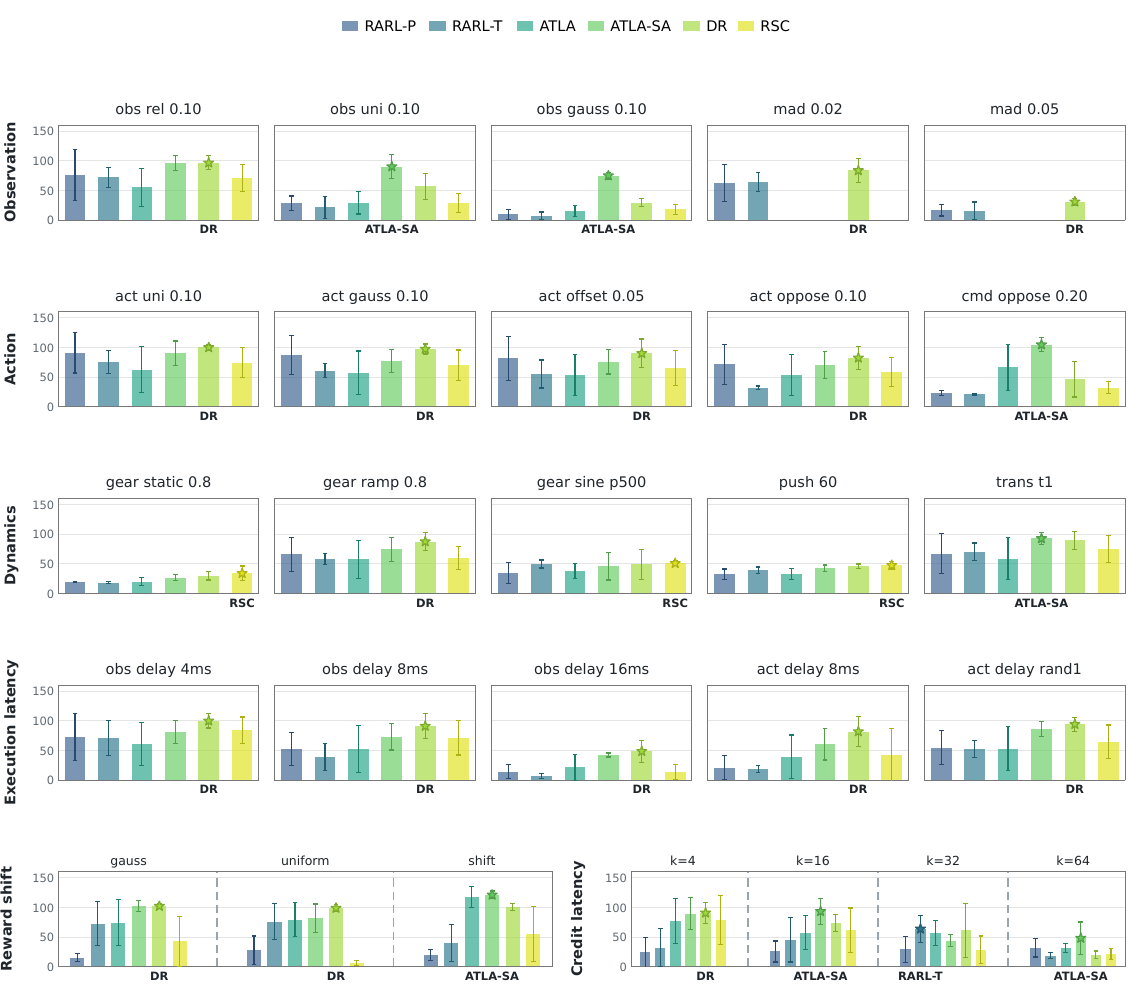}
    \caption{\textbf{Supplementary online results of Section~\ref{sec:channels}.}}
    \label{fig:online_modes}
\end{figure}

\begin{figure}[h]
    \centering
    \includegraphics[width=1\linewidth]{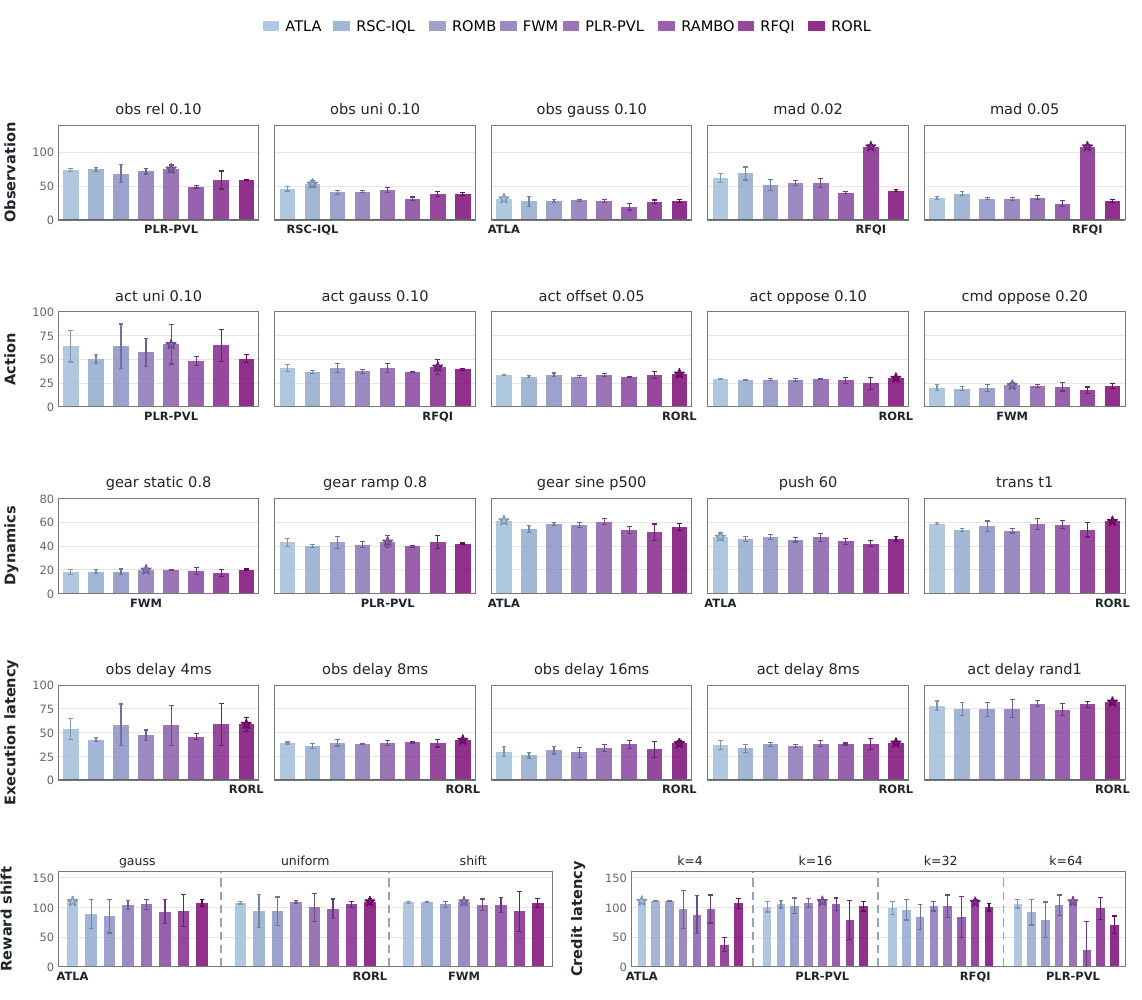}
     \caption{\textbf{Supplementary offline results of Section~\ref{sec:channels}.}}
    \label{fig:offline_modes}
\end{figure}

\begin{figure}[h]
    \centering
    \includegraphics[width=1\linewidth]{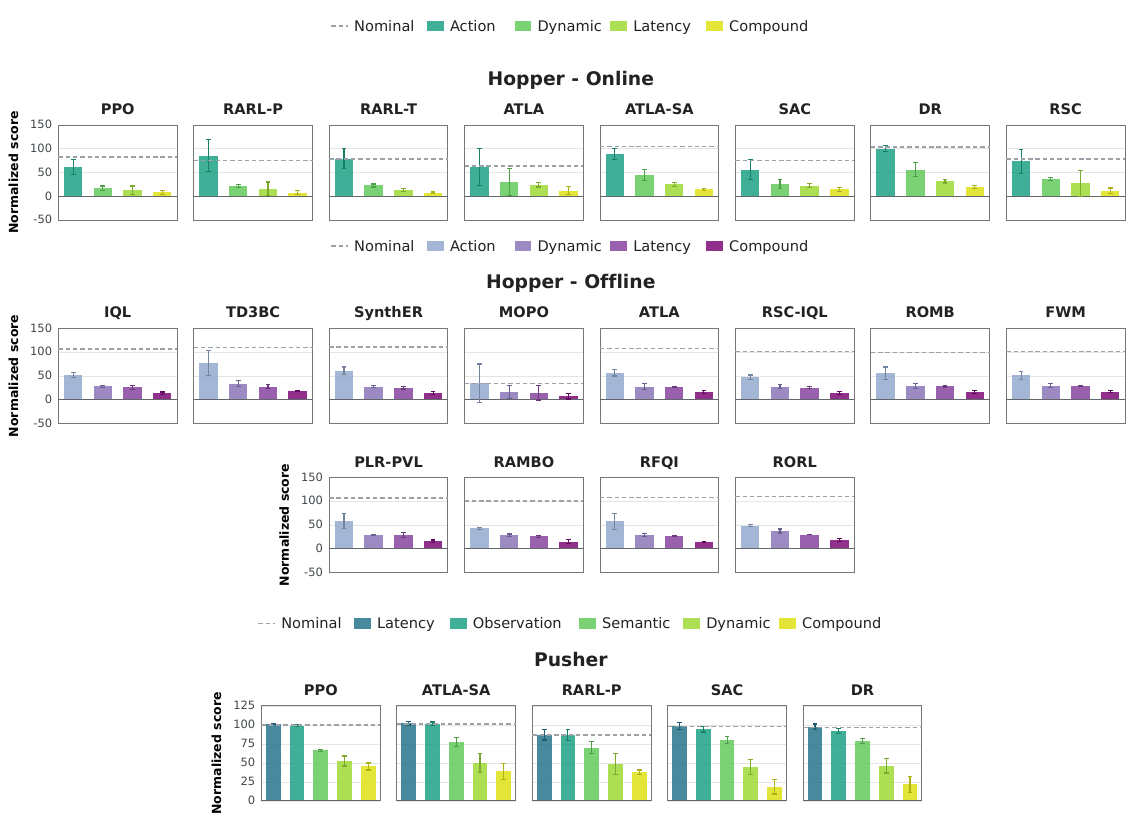}
     \caption{\textbf{Supplementary results of Section~\ref{sec:channels}.}}
    \label{fig:profile_channel_bars}
\end{figure}

Figures~\ref{fig:online_robustness_results}-\ref{fig:offline_robustness_results} show our complete comparison on 220 different configurations across 4 task families in Section~\ref{sec:exp_part1}. Figures~\ref{fig:online_modes} and~\ref{fig:offline_modes} show our full shift-isolated evaluation on 44 distinct perturbation configurations. Figure~\ref{fig:profile_channel_bars} further shows full results on 9 perturbation configurations including compound-shifts scenario.

\begin{figure}[h]
    \centering
    \includegraphics[width=\linewidth,height=0.84\textheight,keepaspectratio]{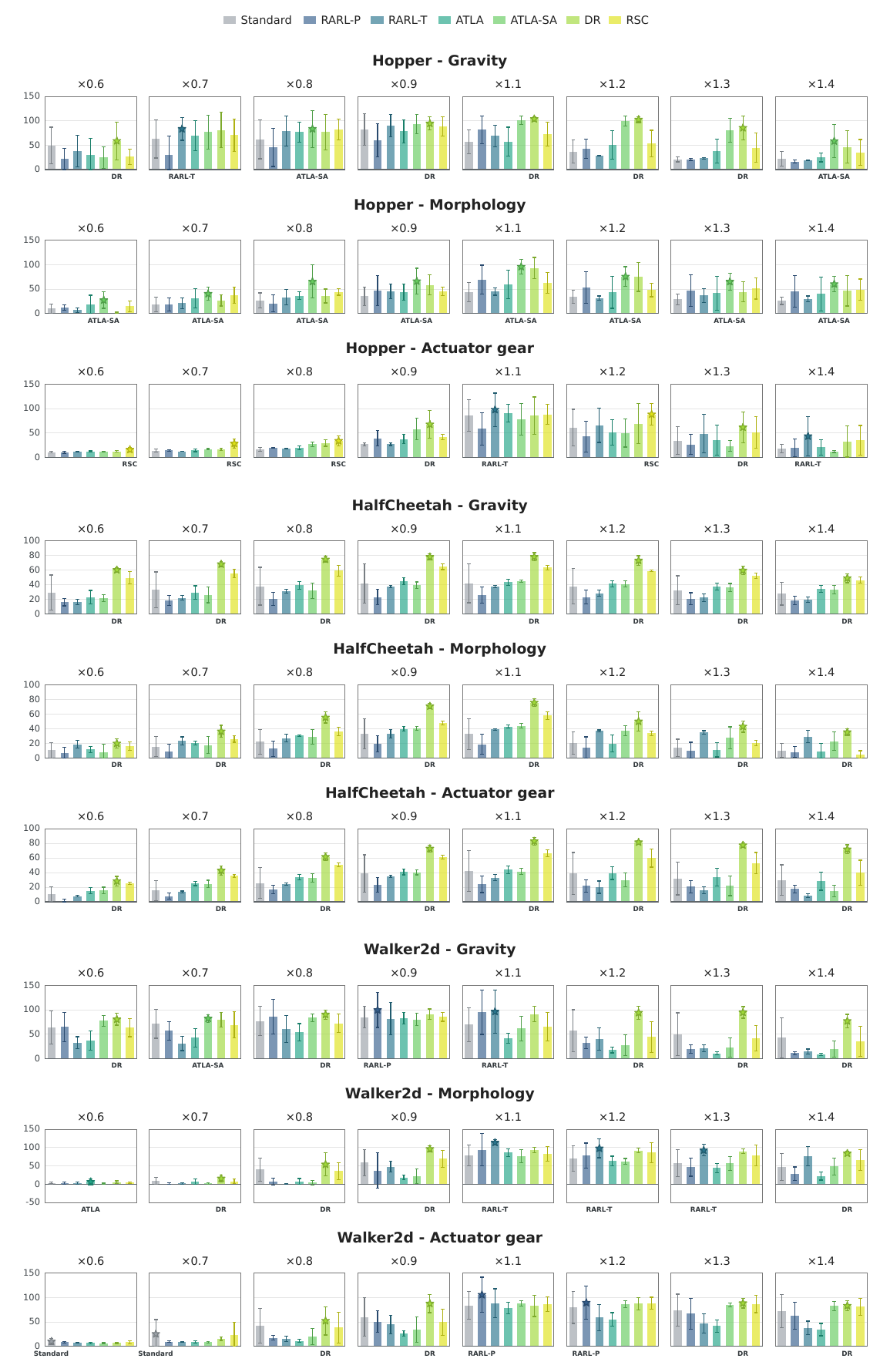}
    \caption{\textbf{Supplementary online results of Section~\ref{sec:exp_part1}.}}
    \label{fig:online_robustness_results}
\end{figure}
\begin{figure}[h]
    \ContinuedFloat
    \centering
    \includegraphics[width=\linewidth,height=0.84\textheight,keepaspectratio,page=2]{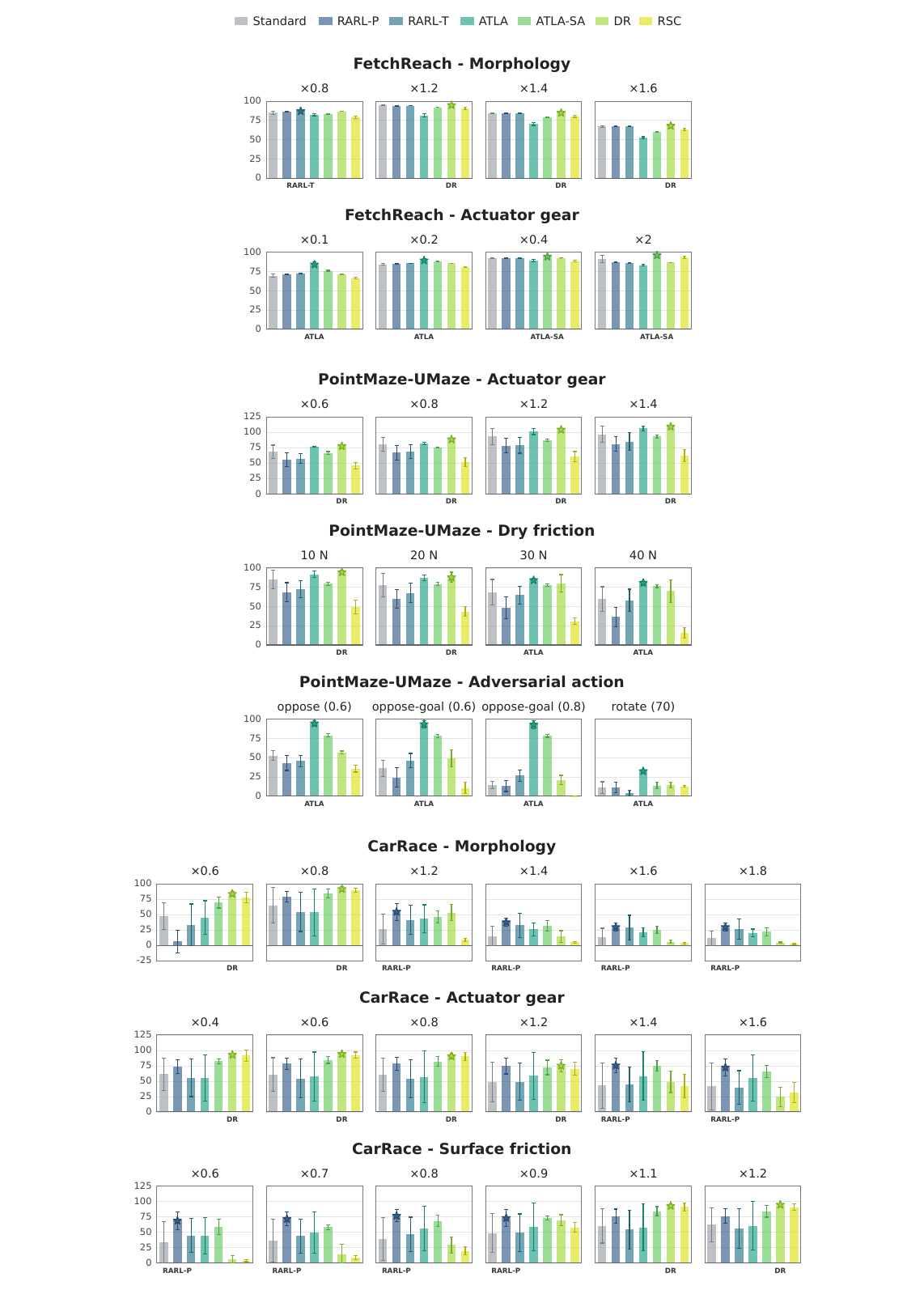}
    \caption{\textbf{Supplementary online results of Section~\ref{sec:exp_part1}.}}
    \label{fig:online_robustness_results_continued}
\end{figure}

\begin{figure}[h]
    \centering
    \includegraphics[width=\linewidth,height=0.84\textheight,keepaspectratio]{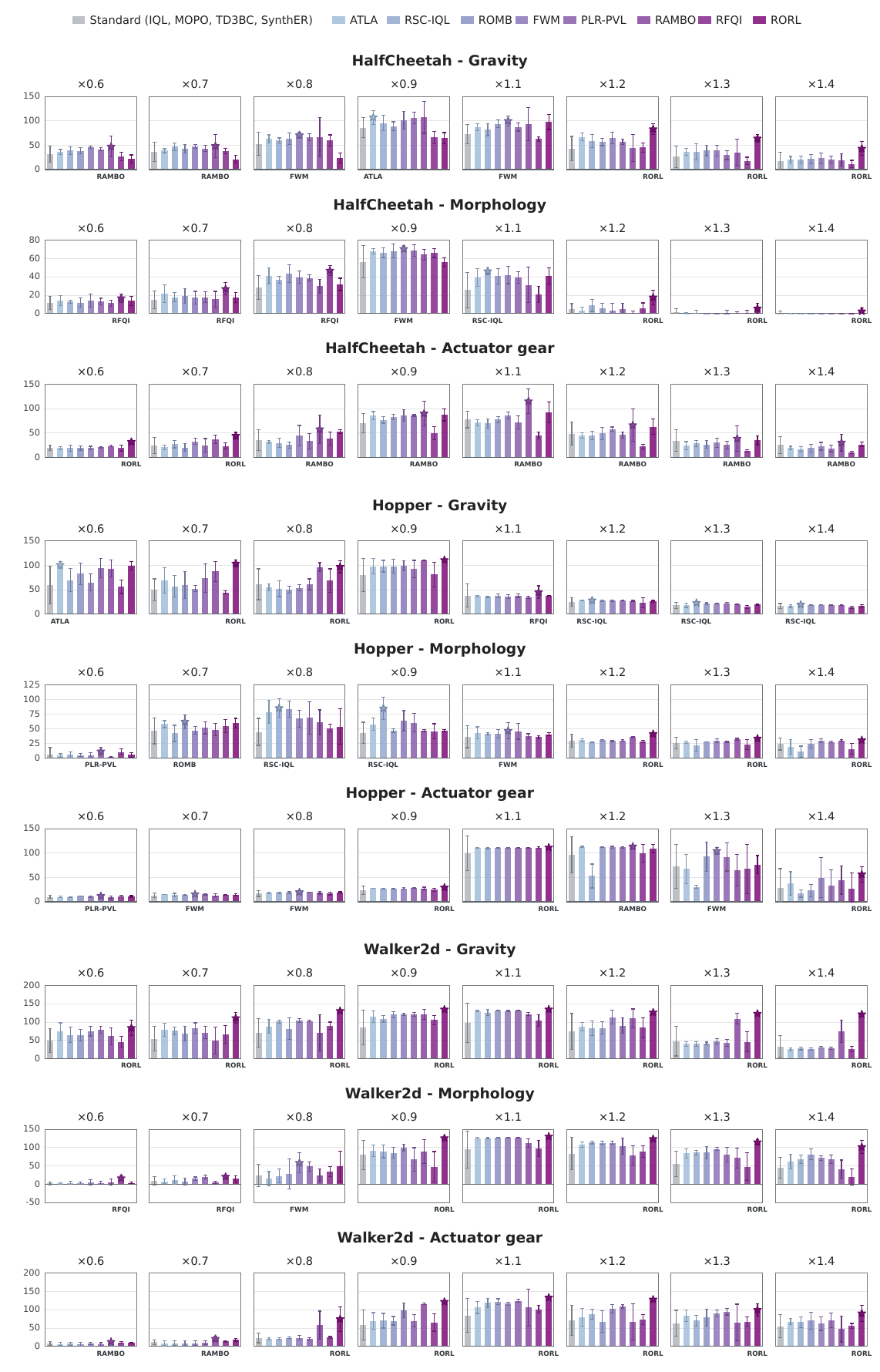}
    \caption{\textbf{Supplementary offline results of Section~\ref{sec:exp_part1}.}}
    \label{fig:offline_robustness_results}
\end{figure}

\begin{figure}[h]
    \ContinuedFloat
    \centering
    \includegraphics[width=\linewidth,height=0.84\textheight,keepaspectratio,page=2]{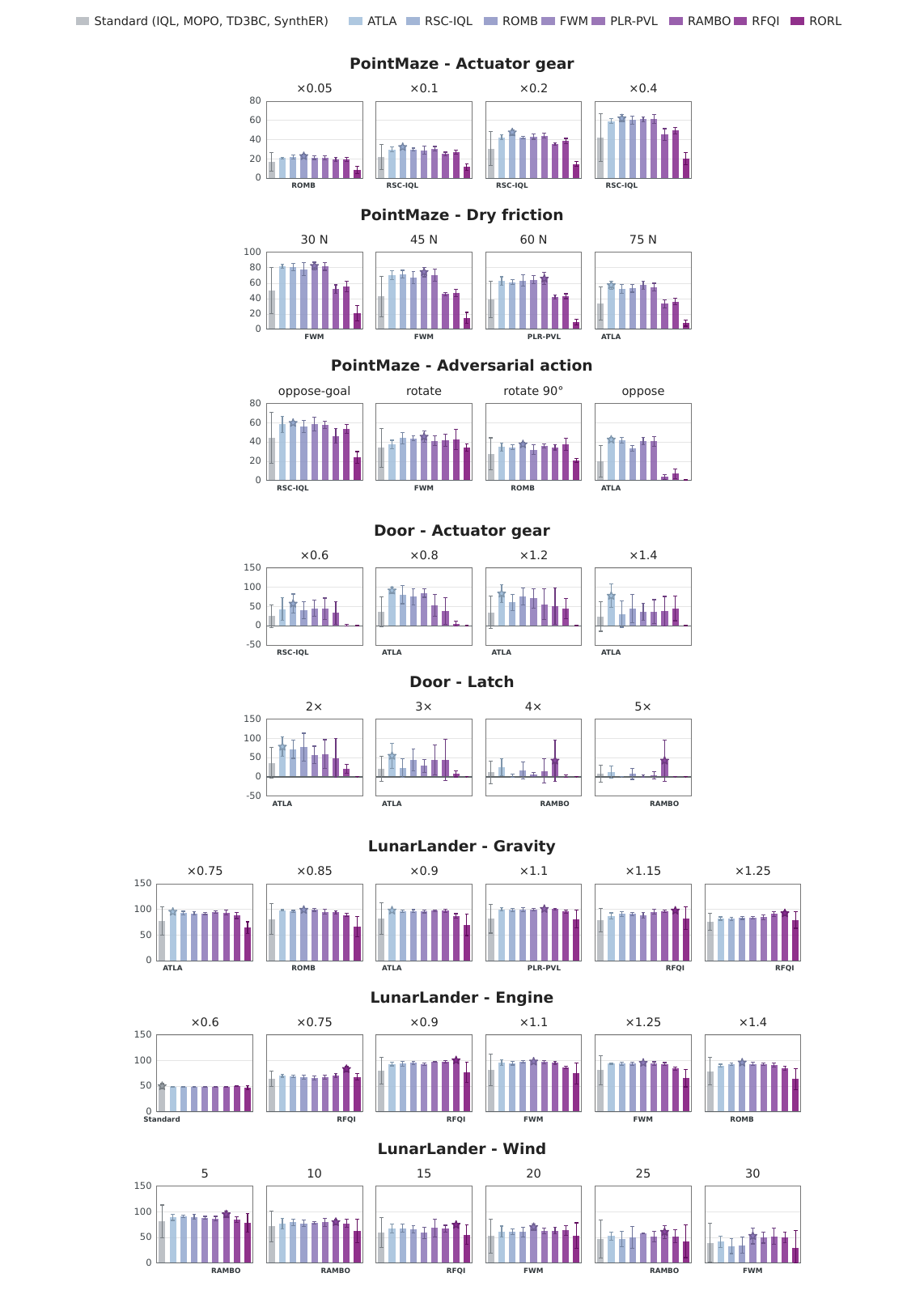}
    \caption{\textbf{Supplementary offline results of Section~\ref{sec:exp_part1}.}}
    \label{fig:offline_robustness_results_continued}
\end{figure}


\FloatBarrier

\end{document}